\documentclass[review]{elsarticle}
\usepackage[utf8]{inputenc}
\usepackage{epstopdf}
\usepackage{placeins}
\usepackage[font=small]{subcaption}
\usepackage[numbers]{natbib}
\usepackage{multirow}
\usepackage{algorithm}
\usepackage{algpseudocode}
\usepackage{amsmath}
\usepackage{graphics}
\usepackage{epsfig}
\usepackage{amsthm}
\usepackage{amssymb}
\usepackage[font=small,skip=0pt]{caption}
\bibpunct[, ]{(}{)}{;}{a}{}{,}
\renewcommand\bibfont{\fontsize{10}{12}\selectfont}

\theoremstyle{plain}
\newtheorem{theorem}{Theorem}[section]

\newtheorem{assumption}[theorem]{Assumption}
\theoremstyle{definition}

\theoremstyle{remark}

\def \E {\mathop{{\mathbb{E}}}\limits}

\def \N {\mathcal{N}}
\def \L {\mathcal{L}}

\def \E {\mathop{{\mathbb{E}}}\limits}

\def \N {\mathcal{N}}
\def \L {\mathcal{L}}

\journal{}

\begin{document}

\begin{frontmatter}

\title{Auto robust relative radiometric normalization via latent change noise modelling}

\author[myfootnote]{Shiqi Liu\corref{mycorrespondingauthor}}\author[mysecondfootnote]{Lu Wang}
\author[mysecondfootnote]{Jie Lian}\author[mysecondfootnote]{Ting chen}\author[mythirdfootnote]{Cong Liu} \author[mythirdfootnote]{Xuchen Zhan}  \author[mysecondfootnote]{Jintao Lu} \author[mysecondfootnote]{Jie Liu} \author[mysecondfootnote]{Ting Wang} \author[mysecondfootnote]{Dong Geng}
\author[mysecondfootnote]{Hongwei Duan}\author[mysecondfootnote]{Yuze Tian}
\address[myfootnote]{aaa\_330473405@qq.com}
\address[mysecondfootnote]{\{chenting,wanglu,lujintao,liujie,wangting,gengdong,liaojian, lianjie,wangfan,duanhongwei\}@diit.cn}
\address[mythirdfootnote]{\{liusongwhu,xczhan94\}@gmail.com}

%
\address{Innovation Center of Beijing Data Intelligence Information Technology Co., Ltd., Wuhan, Hubei, China}
%


\begin{abstract}
Relative radiometric normalization(RRN) of different satellite images of the same terrain is necessary for change detection, object classification/segmentation, and map-making tasks. However, traditional RRN models are not robust, disturbing by object change, and RRN models precisely considering object
change can not robustly obtain the no-change set. This paper proposes auto robust relative radiometric normalization methods via latent change noise modeling. They utilize the prior knowledge that no change points possess small-scale noise under relative radiometric normalization and that change points possess large-scale radiometric noise after radiometric normalization, combining the stochastic expectation maximization method to quickly and robustly extract the no-change set to learn the relative radiometric normalization mapping functions. This makes our model theoretically grounded regarding the probabilistic theory and mathematics deduction. Specifically, when we select histogram matching as the relative radiometric normalization learning scheme integrating with the mixture of Gaussian noise(HM-RRN-MoG), the HM-RRN-MoG model achieves the best performance. Our model
possesses the ability to robustly against clouds/fogs/changes. Our method naturally generates a robust evaluation indicator for RRN that is the no-change set root mean square error. We apply the HM-RRN-MoG model to the latter vegetation/water change detection task, which reduces the radiometric contrast and NDVI/NDWI differences on the no-change set, generates consistent and comparable results. We utilize the no-change set into the building change detection task, efficiently reducing the pseudo-change and boosting the precision.
\end{abstract}


\end{frontmatter}


\section{Introduction}

The radiance/color information are critic for semantic segmentation of different objects, classification of vegetations, change detection, mapping and retrieval in remote sensing tasks. However, due to the imaging time, light source, solar altitude angle, atmospheric attenuation, sensor type, and other factors, there are inconsistencies in an invariant object's radiances in different remote sensing images. These inconsistencies may result in different reflectance and spectral indices(such as NDVI, NDWI) even after coarse radiometric calibration and atmospheric correction, hindering many tasks, including classifying vegetation and other change detection tasks. Moreover, when mosaicking multiple remote sensing images, these inconsistencies cause radiometric/color contrast in the mosaic image, creating unsightly seam lines (\cite{liu2021automatically}). Therefore, eliminating or alleviating these inconsistencies becomes essential.

There are two main approaches to dealing with radiometric inconsistencies: radiometric calibration combining with atmospheric correction and relative radiometric correction. The first approach usually utilizes the gain and bias parameters of the satellite sensor to convert digital number to spectral radiance to achieve the radiometric calibration and utilizes atmospheric radiative transfer model,such as 6S(\cite{vermote1997second}), MORTRAN etc, with atmospheric parameters, such as atmospheric optical depth(AOD) obtained from MODIS/MERRA2 data and other satellite physical parameters to implement atmosphere correction. This method is an idealistic way to obtain the absolute surface reflectance. Nevertheless, the radiometric calibration step might be subject to the low update frequency of the gain and bias parameter measured by the ground-based calibration. The change range of the parameters can annually be up to 100 percent in some satellite sensors, making estimating these parameters difficult. Besides, the atmospheric correction step might be subject to spatial and temporal resolution of atmospheric parameters. MODIS's spatial resolution ranges from 10km, and the temporal resolution is about one day. Therefore, there might be a lot of precision error induced by the time/spatial resolution. In some extreme cases where gain, bias, or AOD parameters are unavailable, the radiometric calibration and atmospheric correction become intractable. The second approach does not depend on those physic parameters but utilizing a reference image to implement relative radiometric normalization(RRN) to calibrate the radiometric level of the target image to the radiometric level of the reference image. RRN methods combining with linear or monotone mapping assumption could alleviate the radiometric inconsistency of the digital number remote sensing images as well as the coarsely radiometric calibrated and atmospherical-corrected images. \cite{el2008relative,du2002radiometric} utilizes RRN to obtain normalized difference vegetation index(NDVI) for latter tasks, and \cite{el2008relative} shows that the NDVI obtained by the RRN approach has a high correlation and has a comparable pattern with the NDVI with the atmospheric correction method despite a relative error of 12.66 percent between values. Those results indicate that relative radiometric normalized remote sensing images might be helpful in vegetation classification and change detection tasks.

RRN methods include linear image regression model(L)(\cite{jensen1983urban},\cite{singh1989review},\cite{olsson1993regression},\cite{zhang2008automatic}), histogram matching model(HM)\cite{richards1999remote}, pseudo-invariant feature(PIF)\cite{schott1988radiometric}, radiometric control sets(RCS)(\cite{hall1991radiometric}), No-change set(NC)(\cite{elvidge1995relative}) but the main situation is that simple automatic RRN models are not robust disturbing by object change and RRN models precisely considering object change might not either robustly obtain sufficient no-change set.

\begin{figure}[ht]
\centering
\captionsetup[subfigure]{font=scriptsize,labelfont=scriptsize}
\begin{subfigure}[t]{.155\textwidth}
	\centering
	\includegraphics[width=\textwidth]{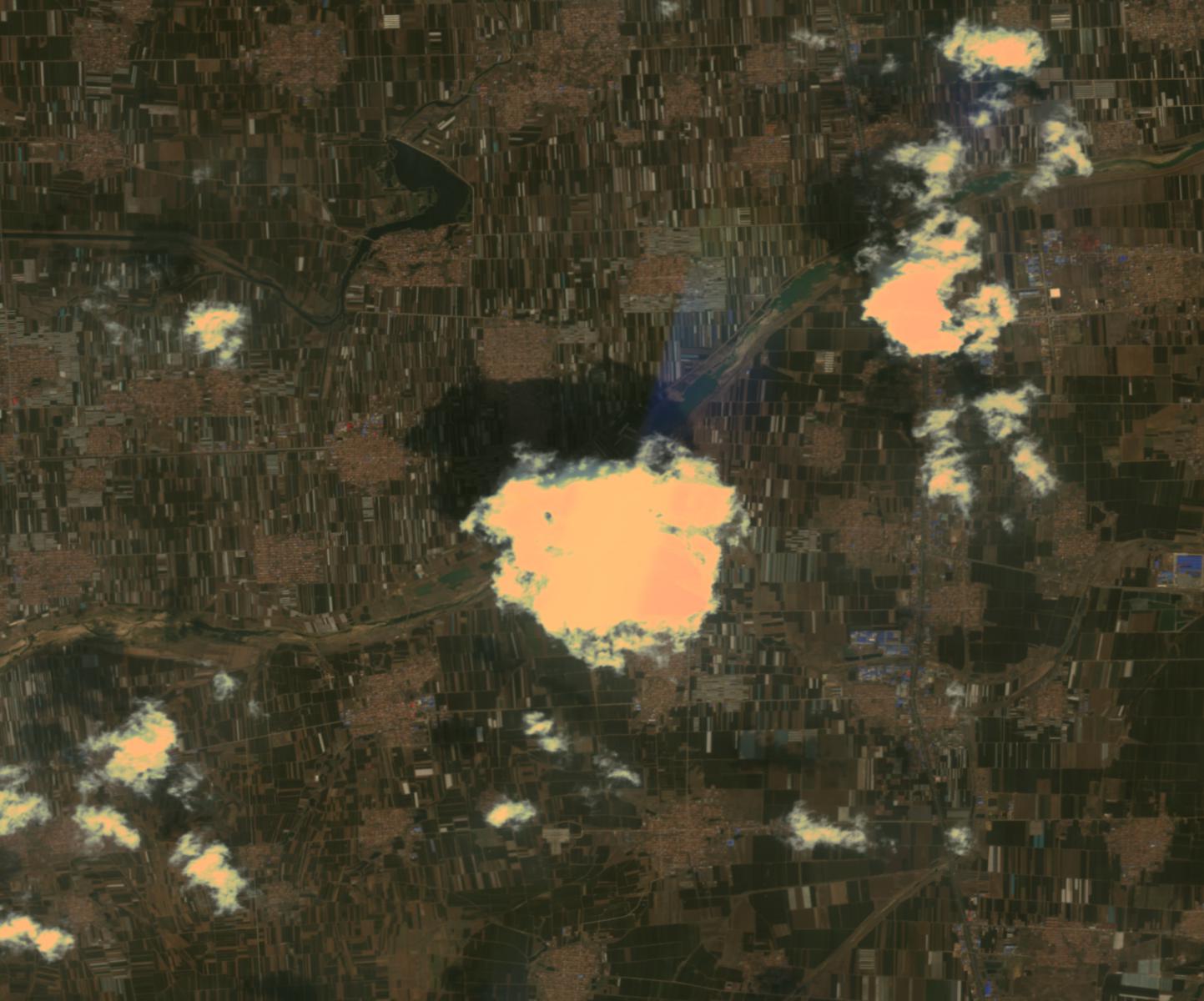}
	\caption{Target image A}
\end{subfigure}
\hfill
\begin{subfigure}[t]{.155\textwidth}
	\centering
	\includegraphics[width=\textwidth]{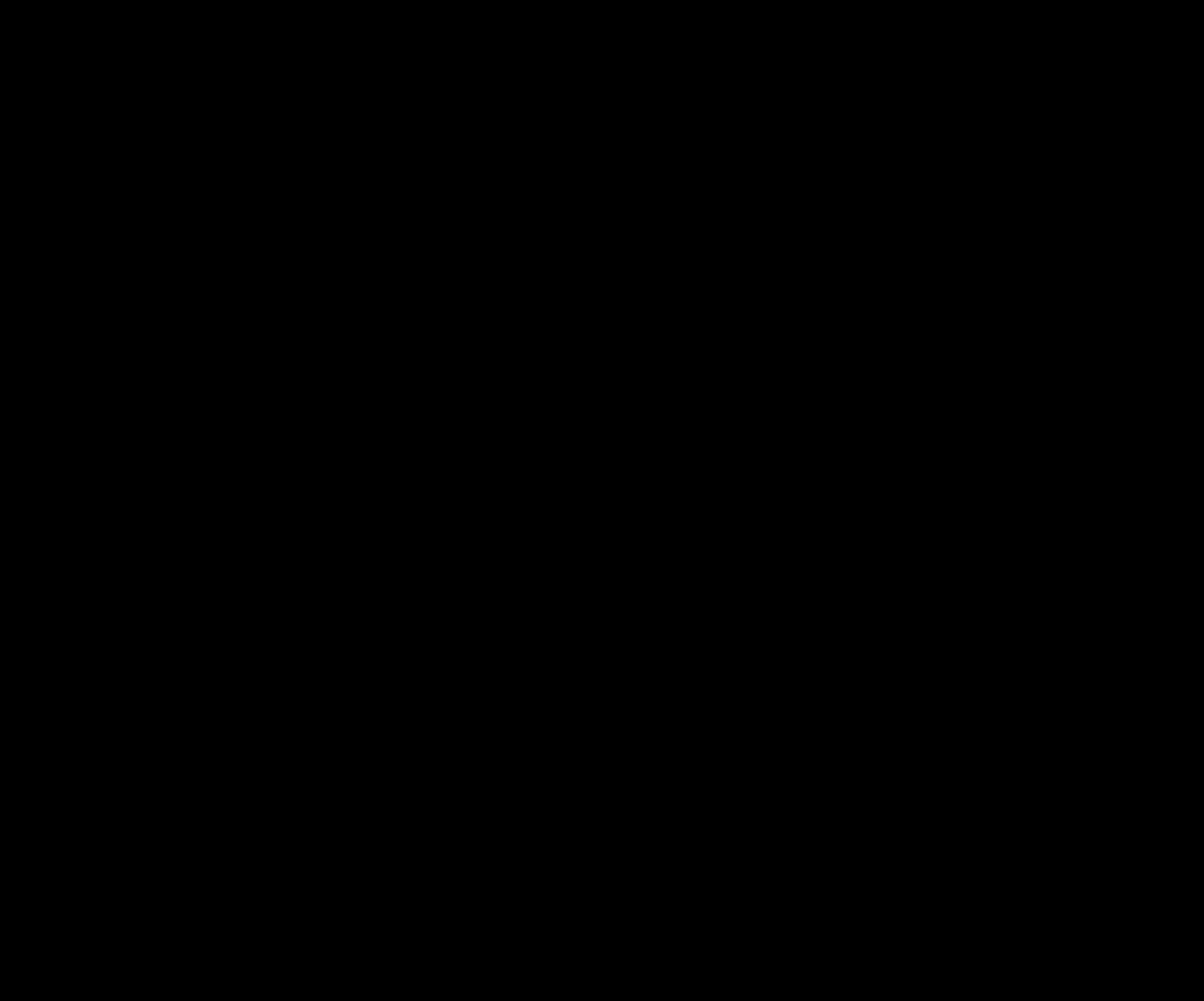}
	\caption{L-RRN NC}
\end{subfigure}
\hfill
\begin{subfigure}[t]{.155\textwidth}
	\centering
\includegraphics[width=\textwidth]{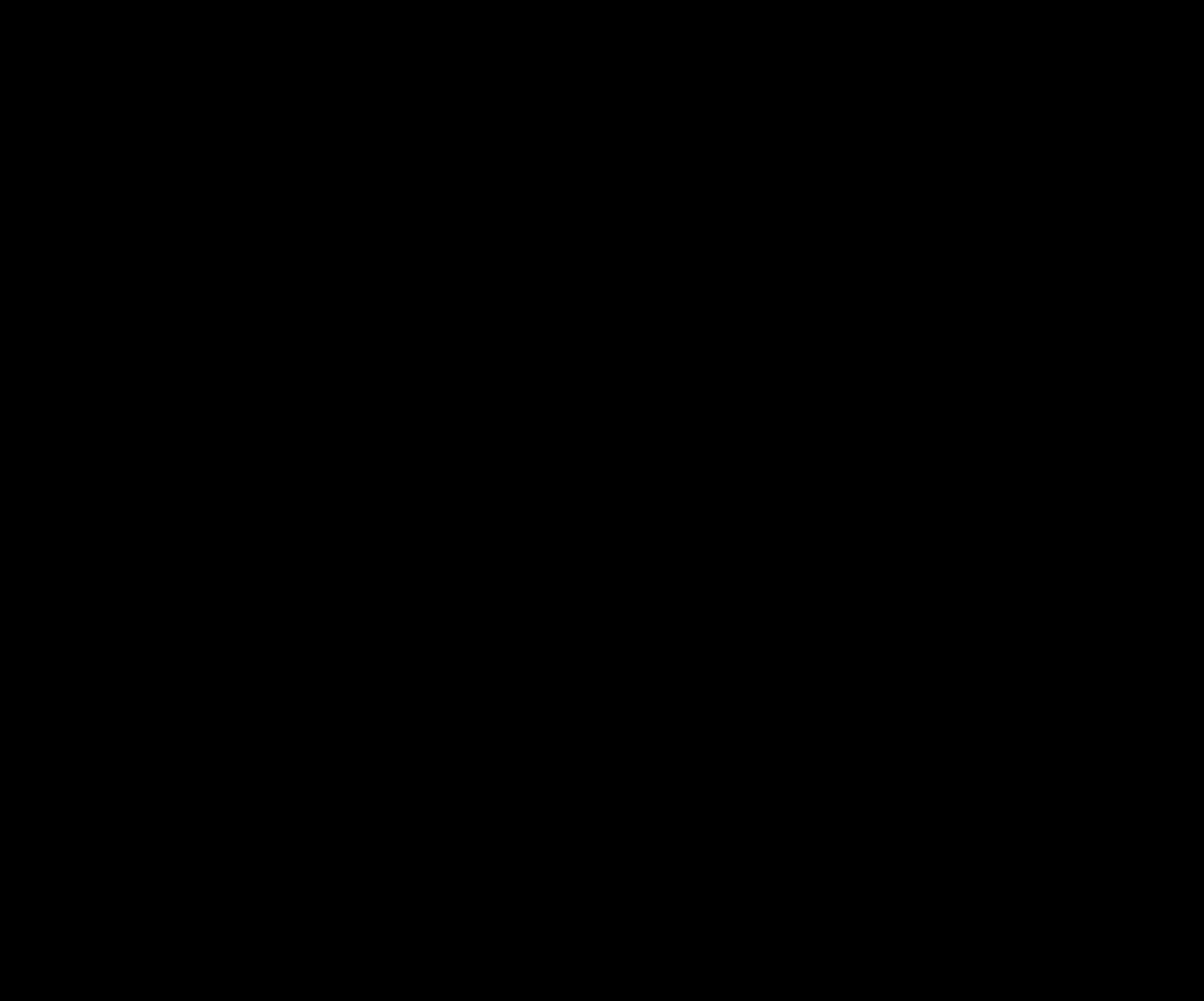}
	\caption{HM-RRN NC}
\end{subfigure}
\hfill
\begin{subfigure}[t]{.155\textwidth}
	\centering
\includegraphics[width=\textwidth]{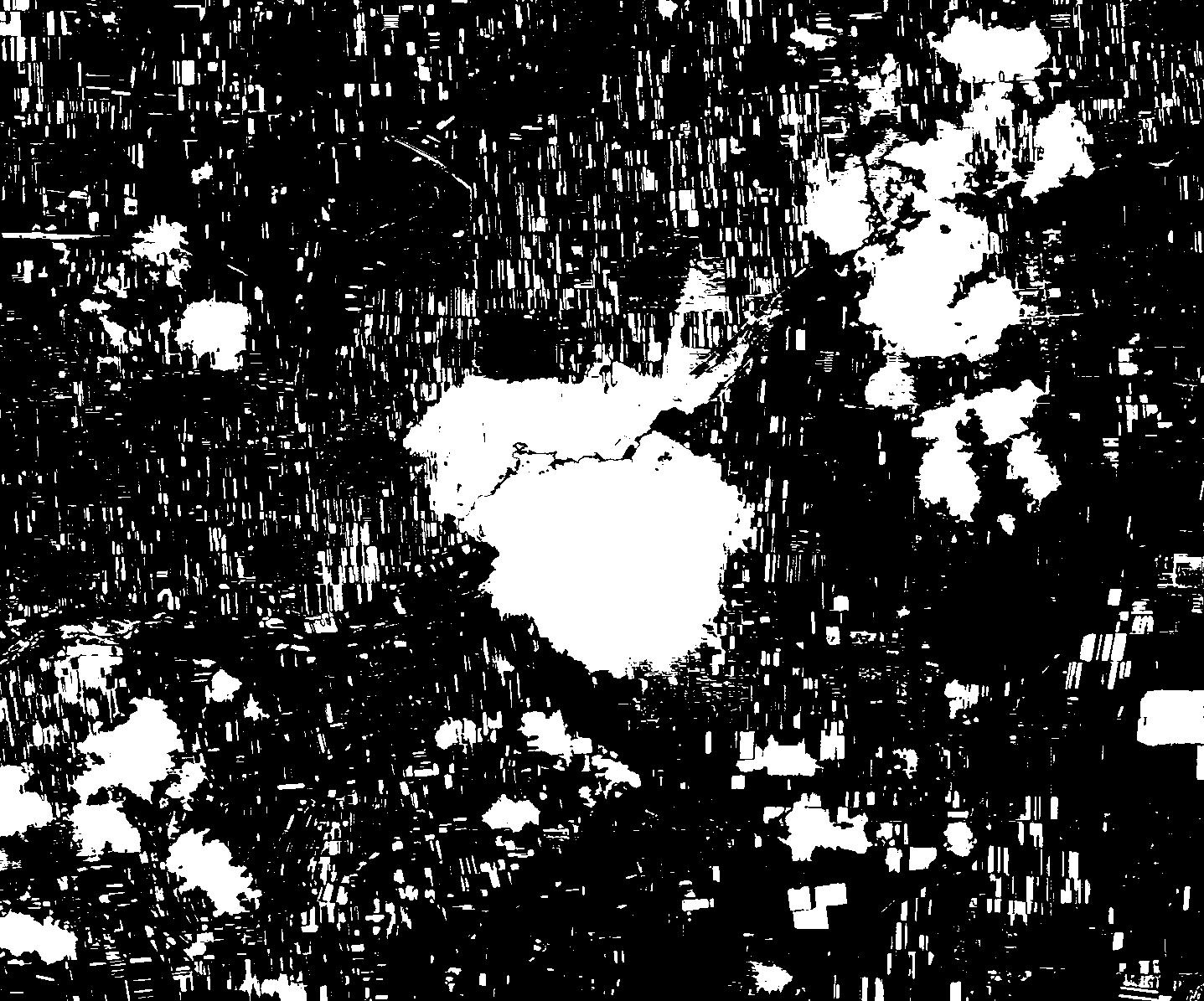}
	\caption{L-RRN-MoG NC}
\end{subfigure}
\hfill
\begin{subfigure}[t]{.155\textwidth}
	\centering
\includegraphics[width=\textwidth]{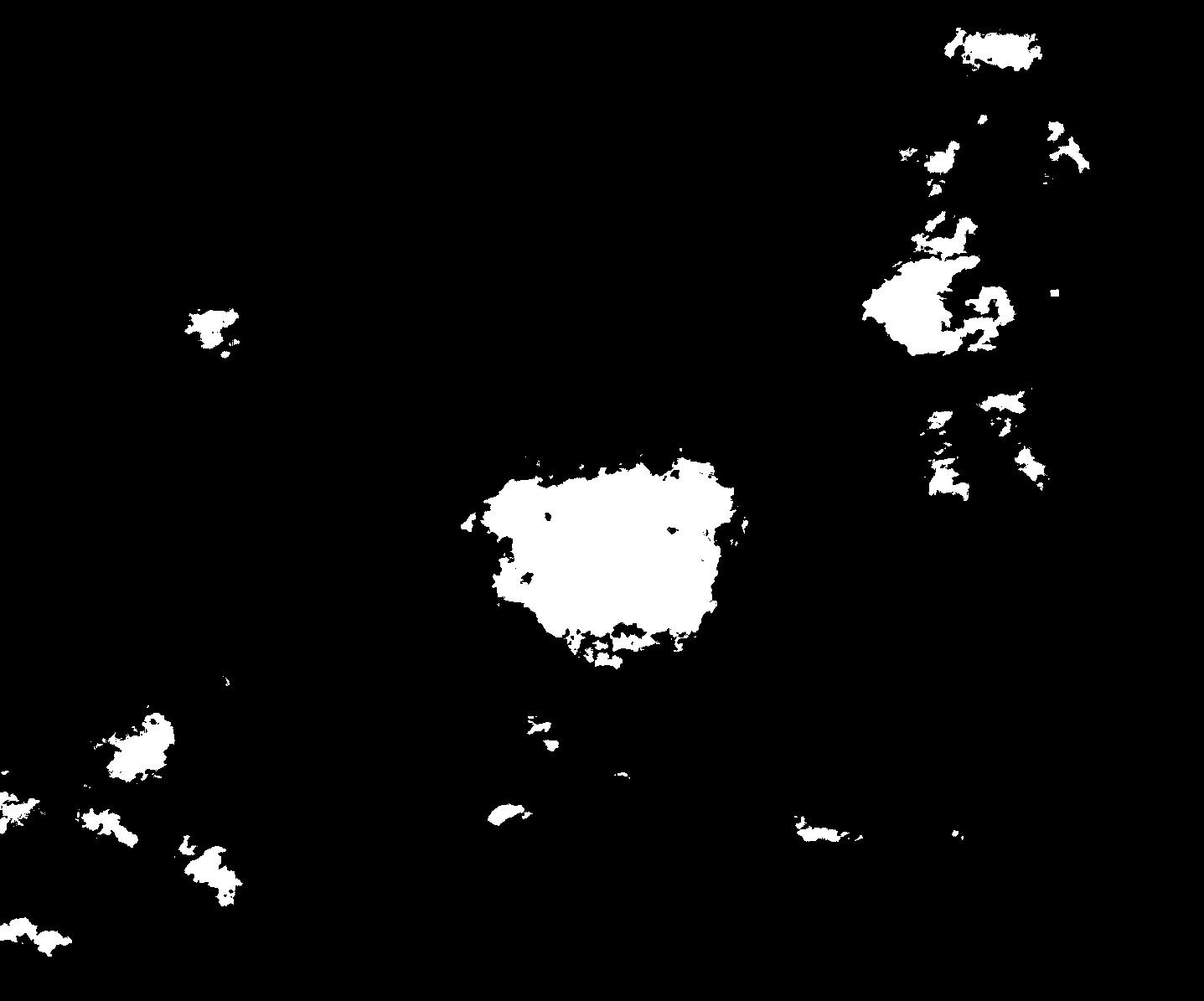}
	\caption{HM-RRN-MoL NC}
\end{subfigure}
\hfill
\begin{subfigure}[t]{.155\textwidth}
	\centering
\includegraphics[width=\textwidth]{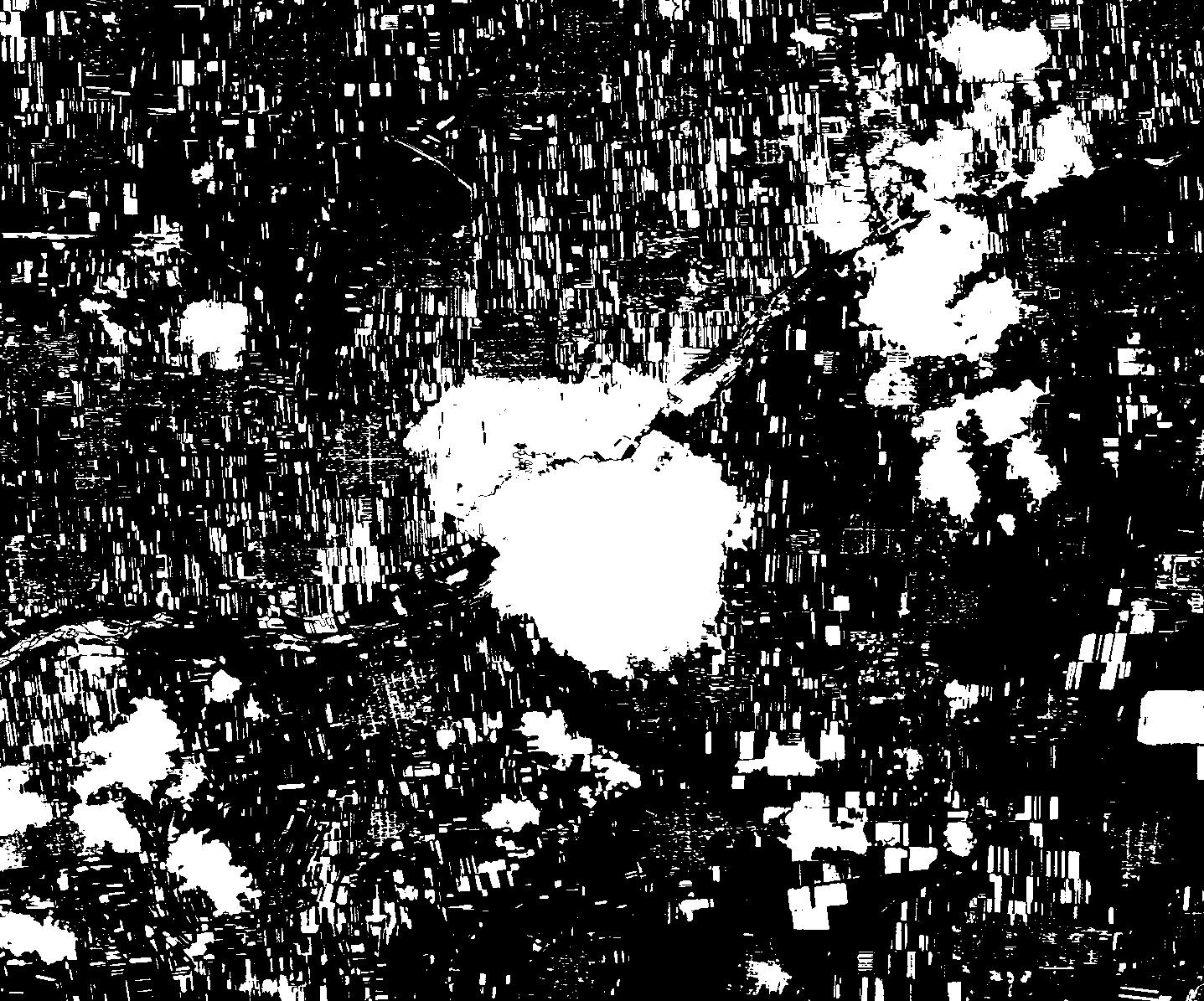}
	\caption{HM-RRN-MoG NC}
\end{subfigure}
\hfill
\begin{subfigure}[t]{.155\textwidth}
	\centering
\includegraphics[width=\textwidth]{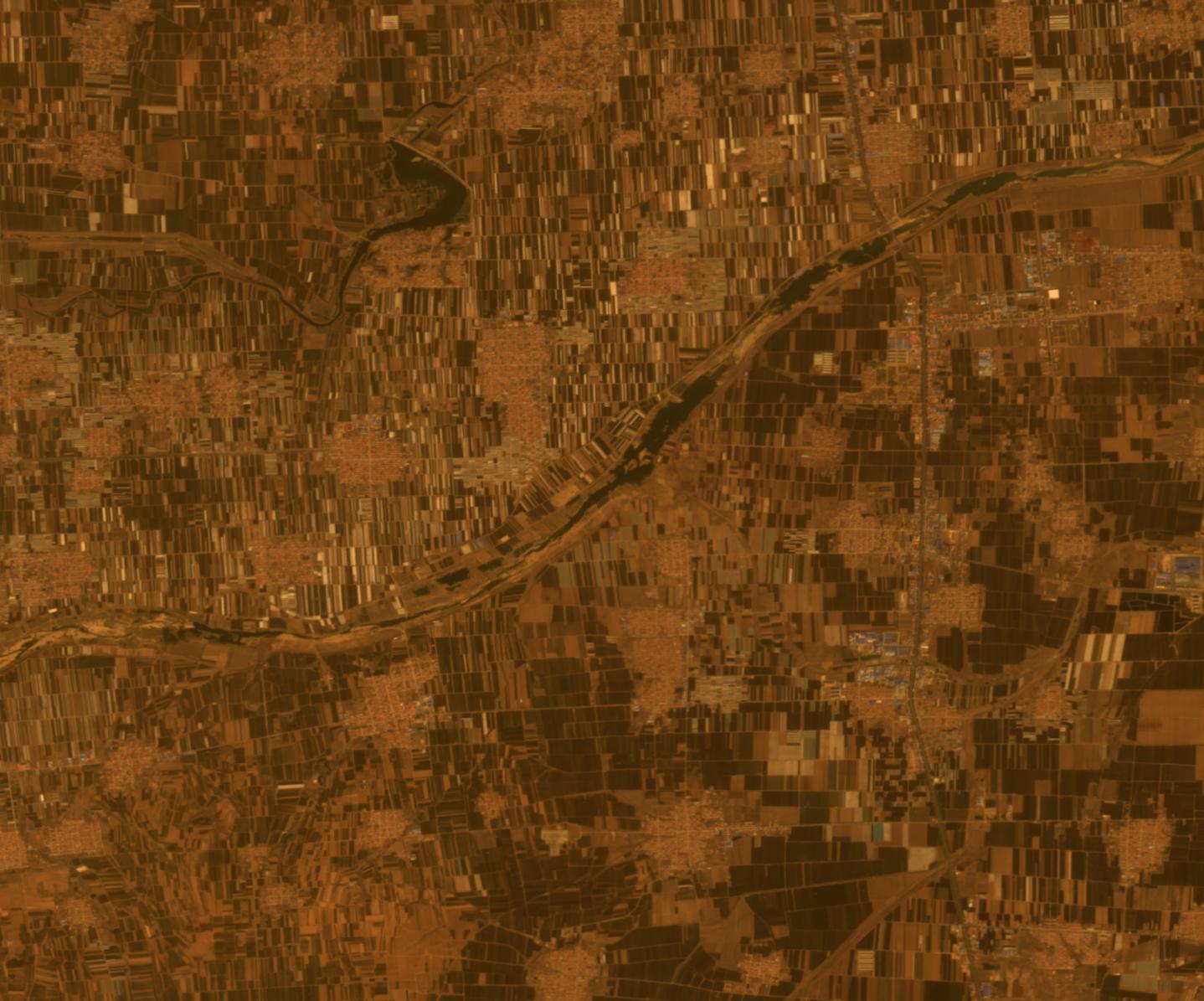}
	\caption{Source image B}
\end{subfigure}
\hfill
\begin{subfigure}[t]{.155\textwidth}
	\centering
\includegraphics[width=\textwidth]{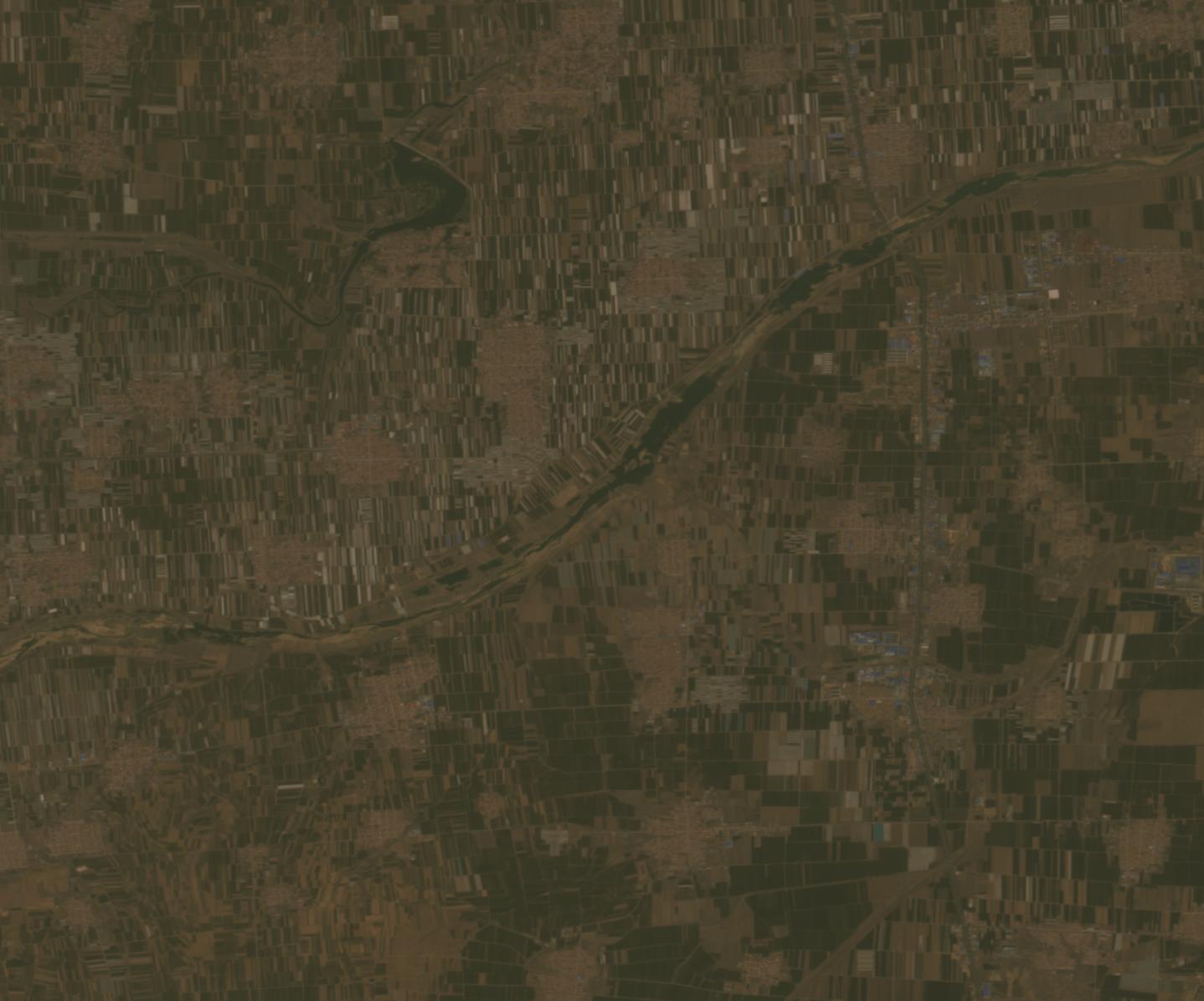}
	\caption{L-RRN B}
\end{subfigure}
\hfill
\begin{subfigure}[t]{.155\textwidth}
	\centering
\includegraphics[width=\textwidth]{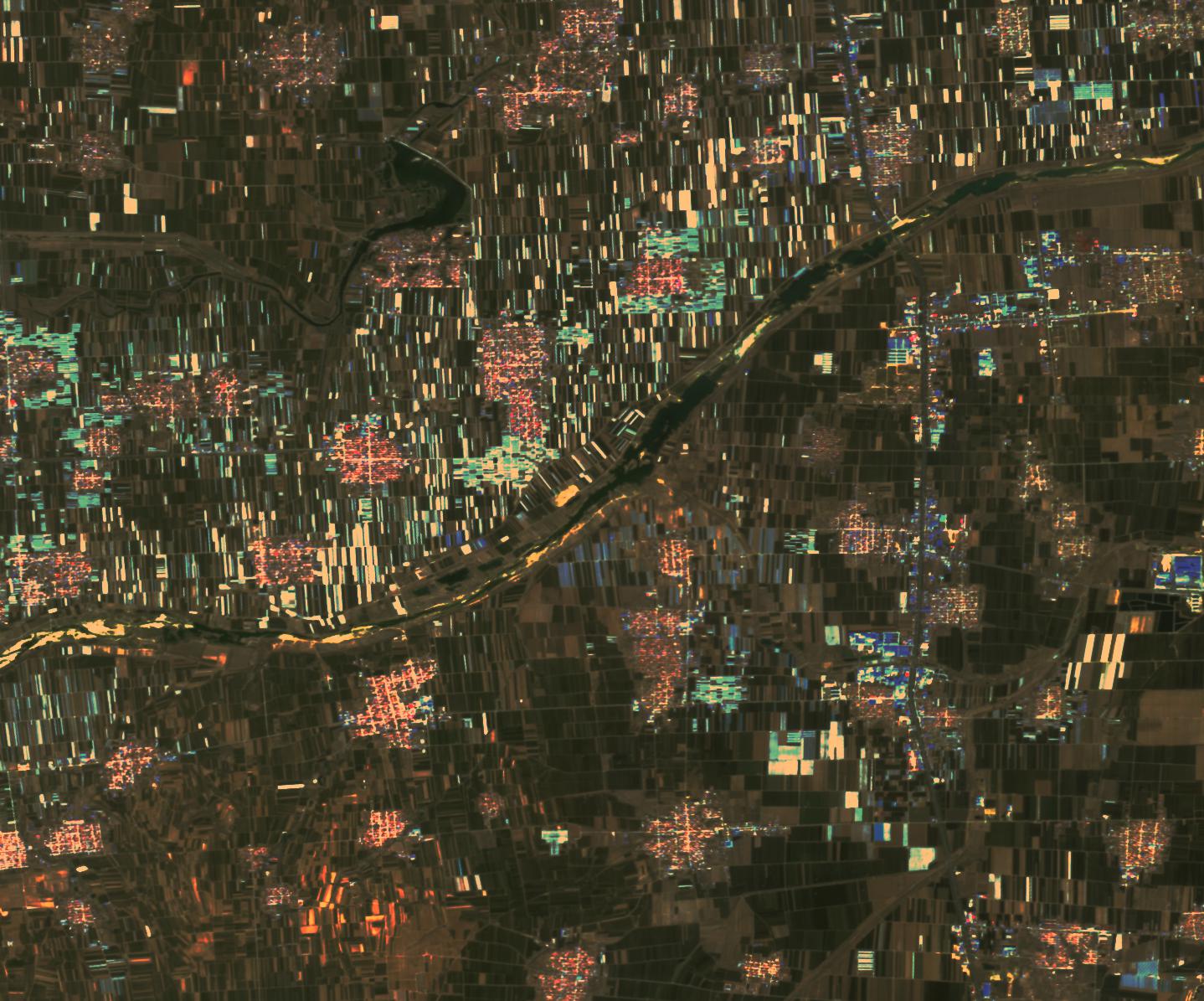}
	\caption{HM-RRN B}
\end{subfigure}
\hfill
\begin{subfigure}[t]{.155\textwidth}
	\centering
\includegraphics[width=\textwidth]{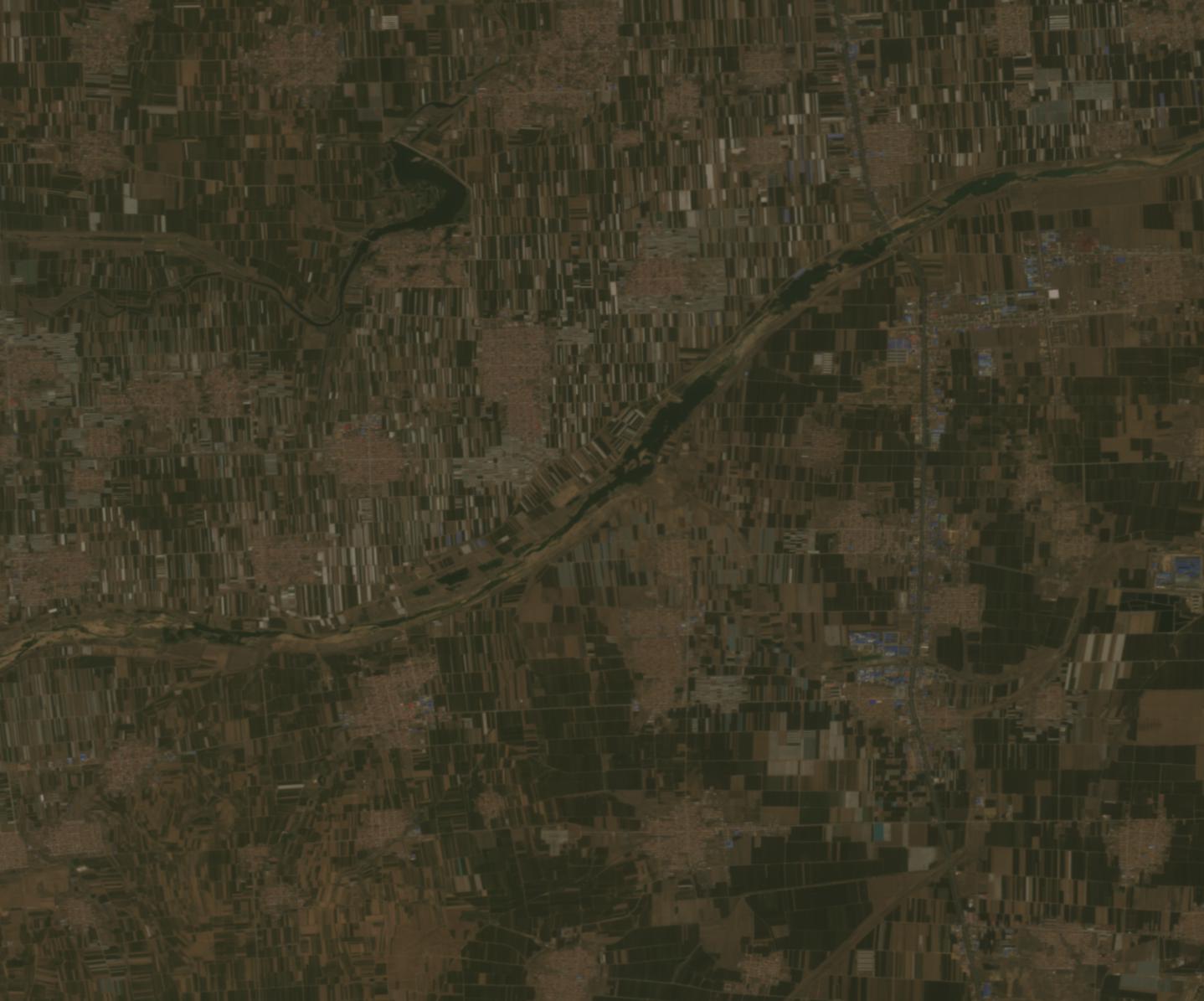}
	\caption{L-RRN-MoG B}
\end{subfigure}
\hfill
\begin{subfigure}[t]{.155\textwidth}
	\centering
\includegraphics[width=\textwidth]{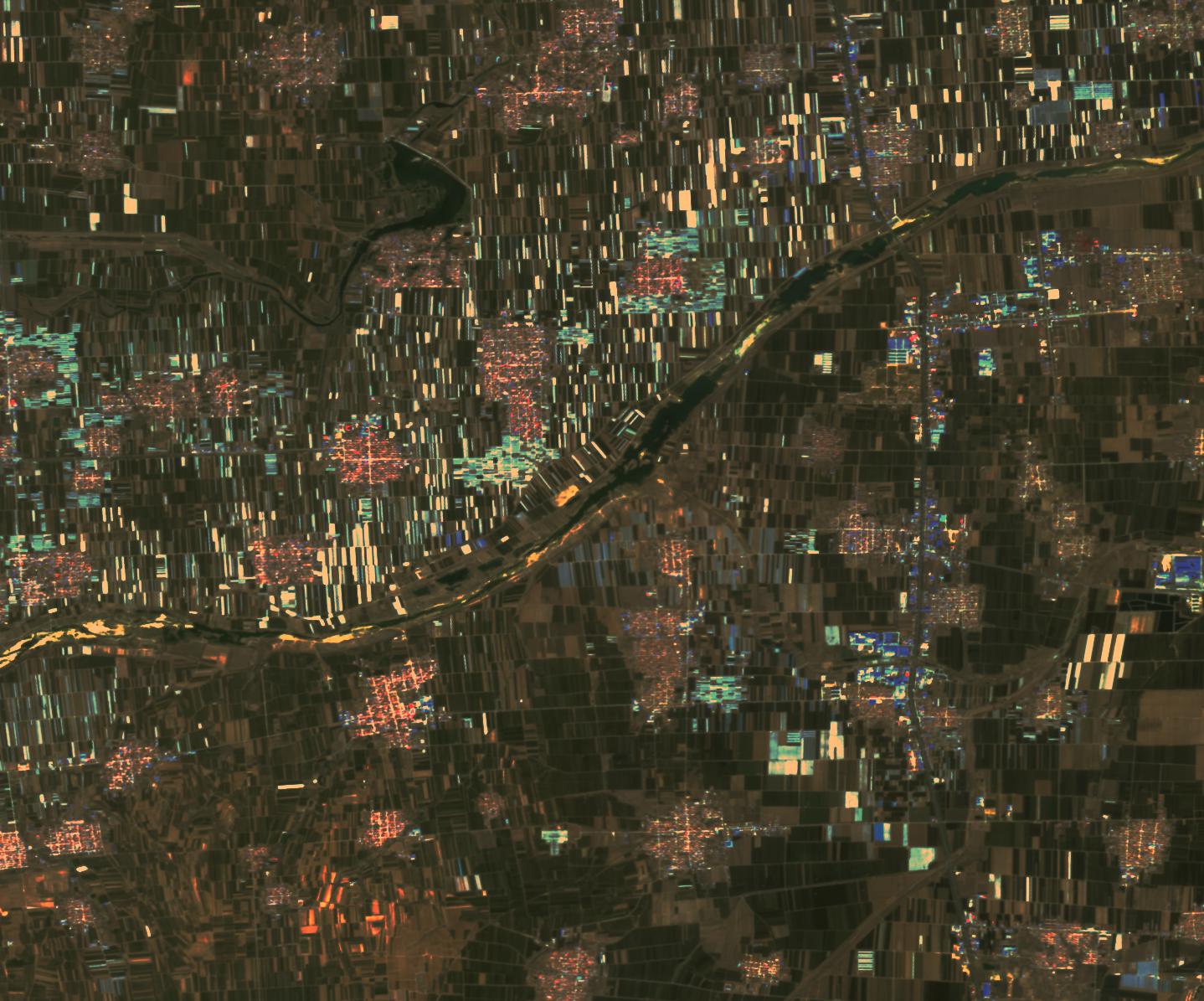}
	\caption{HM-RRN-MoL B}
\end{subfigure}
\hfill
\begin{subfigure}[t]{.155\textwidth}
	\centering
\includegraphics[width=\textwidth]{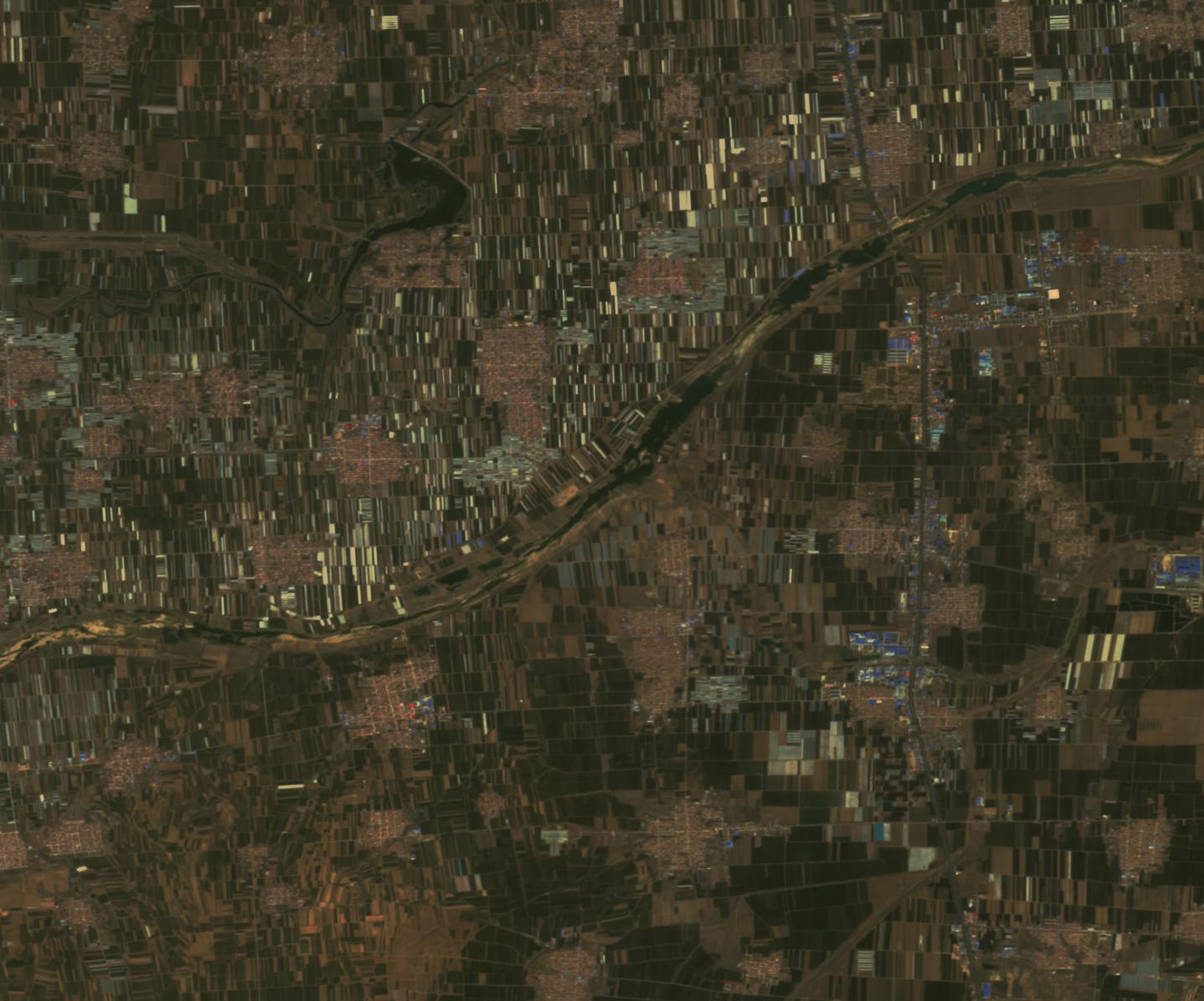}
	\caption{\textbf{HM-RRN-MoG B}}
\end{subfigure}
\hfill
\caption{Comparison figures of 5 different RRN methods part I. Several clouds are in the target image, and the radiometric conditions are different between the source and target images. The target image is greenish, while the source image is yellowish. After implementing the RRN methods, the normalized source image becomes greenish. The black pixels of subfigure(b-f) represent the no-change set(NC) assumed/derived by the different methods.   The image normalized by the HM-RRN-MoG method visually possesses the best normalization results and most accurate no-change set.}
\label{fig:Comparison figures of 5 different methods part I}
\end{figure}

\begin{itemize}
  \item \textbf{Image regression models}: The general image regression models utilize linear mapping, which may only apply to linear radiometric inconsistency scenarios. These model are further fragile when there are some changes between the reference image and target image. The iterative weighted least square method\cite{zhang2008automatic} renders a novel solution to improve the robustness of the model. However, due to the low fitting capacity of the linear model, the model may confuse by nonlinear radiometric inconsistency and wrongly promote to eliminate linear radiometric inconsistency. Meanwhile, this method only proposes the weighting mechanism of one band fitting, and the extension to the multi-band fusion weighting mechanism is not explored.
  \item \textbf{Histogram matching models}: Histogram matching models compute the histograms of reference and target images and establish a looking-up table to perform the relative radiometric normalization. These methods can handle minor nonlinear radiometric inconsistency induced by the sun angle and atmospheric effect\cite{yang2000relative}. The latter section will show that this method can be covert to the image regression model searching in the monotone function via minimizing L1-norm under moderate conditions. However, histogram matching models are either not robust when there are some changes among the reference image and target image.
  \item \textbf{Pseudo-invariant feature models}: Pseudo-invariant feature methods utilize statistical invariance of reflectance of manufactured elements(\cite{schott1988radiometric}) such as asphalt, concrete, rooftops to establish the relative radiometric normalization mapping. The urban feature can be obtained using the infrared band to red band ratio(\cite{biegel1985radiometric}) excluding water, vegetation. These methods might be more robust than the previous methods if water/vegetation changed while pseudo-invariant features did not change. However, the pseudo-invariant features may not always exist, and pseudo-invariant features may also change. For example, people may construct a house on a concrete surface. Those may all influence the robustness of the PIF methods.
  \item \textbf{Radiometric control sets models}: Radiometric control sets methods utilize the sets of scene elements with a mean reflectance remaining nearly unchanged over time(\cite{hall1991radiometric}). Typically, they derive a bright control set using the approach similarly with PIF methods, and a dark control set, which can be the dark water, from the scattergram to linear modeling the relative radiometric normalization mapping. However, the bright radiometric control sets face the same problems described in the PIF methods. Also, nearly unchanged the mean reflectance of the control sets assumption may help linear normalization but maybe not robust in nonlinear cases, particularly when the sub-object reflectance in the control set changed.
  \item \textbf{No-change set model}: No-change set methods (\cite{elvidge1995relative}) utilize the infrared band and red band scattergrams to determine the water and land maximum radiance value among the source and target images to establish a coarse relative radiometric normalization mapping of the infrared band and red band. They use the threshold of normalizing error of coarse RRN mapping to determine the no-change set and use the no-change set to determine the fine RRN mapping. These methods are pretty novel since they utilize the normalization error to determined the no-change set. However, water may not always exist in the image, which is fatal for these methods' robustness. Some minor issues of these methods could be the nonlinearity of the radiometric inconsistency and the disuse of the image's red and green band. Sometimes, this method might need human intervention to obtain stability. One promising direction is to use scale-invariant feature transform(SIFT) to extract a subset of whole no-change set(\cite{sun2012automatic}) and implement the radiometric normalization. However, this direction temporarily is limited by its relatively small number of points in no-change set and may be hard to sustain the complex nonlinear cases while maintaining a sufficient radiometric resolution.
\end{itemize}
Therefore, an auto robust nonlinear RRN method by considering all bands' information and auto extracting sufficient/complete no-change set becomes valuable and desirable.

The core problem becomes how to determine the no-changes in the images robustly. Portraying the change or no-changes with noise modeling methods possesses some practical experience in computer vision and machine learning fields(\cite{yong2017robust,meng2013robust,chen2017denoising,cao2016robust,cao2015low,zhao2014robust,yue2021unsupervised,rui2021learning}). Background subtraction methods via noise modeling (\cite{yong2017robust}), utilizing the low-rank prior knowledge to portray the no-changes and utilizing the mixture of Gaussian noise to model the change of the foreground, achieved incredible success. These methods can utilize different noise assumptions to separate the slight change of leaf shaking and the massive change of a driving car. Similarly, the human face factorization method via noise modeling, utilizing the low dimension prior of the human face to portray the no-changes, utilizing a mixture of Gaussian noise to model change of light and high-frequency face feature change, obtains excellent results. The studies above inspire us to model the no-changes of the reference image and target image under relative normalization and model the noise induced by the change.

This paper proposes auto robust relative radiometric normalization methods via latent change noise modeling. They utilize the prior knowledge that no change points possess small-scale noise under relative radiometric normalization and that change points possess large-scale radiometric noise after radiometric normalization, combining the stochastic expectation maximization method to quickly and robustly extract the no-change set to learn the relative radiometric normalization mapping. Specifically, when we select histogram matching as the relative radiometric normalization learning scheme integrating with the mixture of Gaussian noise(HM-RRN-MoG), the HM-RRN-MoG model achieves the best performance. Our method naturally generates a robust evaluation indicator for RRN that is the no-change set root mean square error. We apply the HM-RRN-MoG model to the latter vegetation/water change detection task, which reduces the radiometric contrast and NDVI/NDWI(\cite{mcfeeters1996use}) differences, generates consistent and comparable results. We utilize the no-change set into the building change detection task, efficiently eliminating the pseudo-change and boosting the precision. In conclusion, our method has advantages as follows:
\begin{itemize}
  \item \textbf{automatic and robust}: it automatically robustly extracts no-change set with all band information(robust to clouds/fogs/changes).
  \item \textbf{evaluable and accurate}: it is an accurate relative radiometric normalization using histogram matching and can be evaluated through robust no-change set mean square error and log-likelihood.
  \item  \textbf{probability and mathematics theoretically grounded}: it is under probability model theoretical support and bridges of the histogram matching model with the L1-norm image regression RRN model with monotone mapping.
  \item \textbf{useful for downstream tasks} including vegetation/water cover change detection, object change detection task(eliminating pseudo-change).
\end{itemize}

\section{Methodology}
Our method is based on discovering the no-change set and simultaneously learning the radiometric normalization mapping on the no-change set of two overlapped images.

Unlike the pseudo-invariant feature-based methods that may fail in the scenarios that lack that kind of feature,  we aim to discover the robust no-change set based on the scenario itself.
\subsection{Latent change noise modeling}
In order to model the no-change set in a theoretically grounded manner, we introduce the probability framework. We first consider the values of a shared point on the same geographic location in the two images. $X_c$ denotes the random variable of the value of that source image point in $c$ band, while $Y_c$ denotes the random variable of the value of that target image point in $c$ band. Suppose the normalization mapping in $c$ band is $f_c$, since except for the $c$ band noise $\varepsilon_c$, the value of $Y_c$ can be determined by $f_c(X_c)$ , then we model the relation of $X_c$ and $Y_c$ using the noise modeling paradigm as follow,
\begin{equation}\label{Noise Modelling}
  Y_c = f_c(X_c) +\varepsilon_c.
  \end{equation}
The noise is influenced by the factor that whether the radiometric/color of the  point  changed or not. Let $Z=(Z_1,Z_2)$ be the 2 dimension binary random variable where $Z_1,Z_2\in \{0,1\}$ and $Z_1+Z_2=1$. $Z_1=1$  implies the value of point didn't change between the two images while $Z_2 = 1$ implies the value of point changed between the two images. Suppose $p(Z_1=1|\Pi) = \pi_1$, $p(Z_2=1|\Pi)=\pi_2$ where $\Pi =(\pi_1,\pi_2)$.

We provide three examples for $\varepsilon_c$ and $f_c$.
\begin{itemize}
\item Case 1: $\varepsilon_c$ follows mixture of 0 mean Gaussian distribution and $f_c(X_c)= w_c X_c + b_c$.
\item Case 2: $\varepsilon_c$ follows mixture of 0 mean Laplace distribution and $f_c$ is monotone (i.e. $f_c(a)\ge f_c(b)$ if $a>=b$) discrete mapping.
\item Case 3: $\varepsilon_c$ follows mixture of 0 mean Gaussian distribution and $f_c$ is monotone discrete mapping.
\end{itemize}
We will later show that first and second cases yield the weighted least square relative radiometric normalization regime and weighted histogram matching relative radiometric normalization regime under moderate condition. The third case yield the robust weighted histogram matching relative radiometric normalization. The case 3 shares the same equation with case 1 except for the $f_c$ part.
\subsubsection{Condition probability distribution assumption}
Let \begin{itemize}
      \item $\Sigma_c=({\sigma_c}_1^2,{\sigma_c}_2^2)$ for case 1 and 3,
      \item $\Sigma_c=({\beta_c}_1,{\beta_c}_2)$ for case 2,
    \end{itemize}
and $\Sigma = [\Sigma_1,\cdots,\Sigma_C]$ where ${\sigma_c}_k^2$ is the variance of $k$th Gaussian distribution of band $c$, ${\beta_c}_k$ is the scale parameter of $k$th Laplace distribution of band $c$ and $C$ is the total band numbers. we have

\begin{itemize}
\item  Case 1 or 3: \begin{equation}\label{conditional noise distribution 13} p(\varepsilon_c|Z_k = 1,\Sigma){=}\N(\varepsilon_c|0,{\sigma_c}_k^2)  \quad k\in\{1,2\}
    \end{equation},
\item Case 2: \begin{equation}\label{conditional noise distribution 2}  p(\varepsilon_c|Z_k = 1,\Sigma) {=}\L(\varepsilon_c|0,{\beta_c}_k)  \quad k\in\{1,2\}\end{equation},
\end{itemize}
where $\N(\varepsilon_c|0,{\sigma_c}_k^2)=\frac{1}{\sqrt{2\pi{\sigma_c}_k^2}}e^{-\frac{(\varepsilon_c-0)^2}{2{\sigma_c}_k^2}}$ and $\L(\varepsilon_c|0,{\beta_c}_k)=\frac{1}{2{\beta_c}_k}e^{-\frac{\vert\varepsilon_c-0\vert}{{\beta_c}_k}}$.
The above equation implies that the change points and no-change points will have different noise levels in the band $c$. Intuitively, the no-change points set will have a low noise level, and change points will have a high noise level and sometimes may be with a heavy tail. The marginal distribution of the $\varepsilon_c$ as follows,
\begin{eqnarray}
  p(\varepsilon_c|\Pi,\Sigma) &=& \sum_{k=1}^{K}p(\varepsilon_c|Z_k =1,\Sigma)p(Z_k = 1|\Pi)  \\
   &\overset{case\ 1(3)}{=}& \sum_{k=1}^{K}\pi_k\N(\varepsilon_c|0,{\sigma_c}_k^2) \\
   &\overset{case\ 2}{=}& \sum_{k=1}^{K}\pi_k \L(\varepsilon_c|0,{\beta_c}_k).
\end{eqnarray}\label{mixture of Gaussian}

Let the noises be independent amount bands. Given the radiometric normalization $f_c$ and $F=[f_1,\cdots,f_C]$, the conditional distribution of $Y_1,\cdots,Y_C|X_1,\cdots,X_C,Z$ is the following,
\begin{eqnarray}\label{Y|X,Zconditional distribution}
   && p(Y_1,\cdots,Y_C|X_1,\cdots,X_C,Z,\Sigma,F) \\
   &=&\prod_{c=1}^{C}p(Y_c|X_c,Z_k =1,F,\Sigma) \\
   &\overset{case\ 1(3)}{=}& \prod_{c=1}^{C}\prod_{k=1}^{K}\N(Y_c|f_c(X_c),{\sigma_c}_k^2)^{Z_k}\\
   &\overset{case\ 2}{=}& \prod_{c=1}^{C}\prod_{k=1}^{K}\L(Y_c|f_c(X_c),{\beta_c}_k)^{Z_k}.
\end{eqnarray}
\begin{figure}[ht]
\centering
\includegraphics[width=5cm]{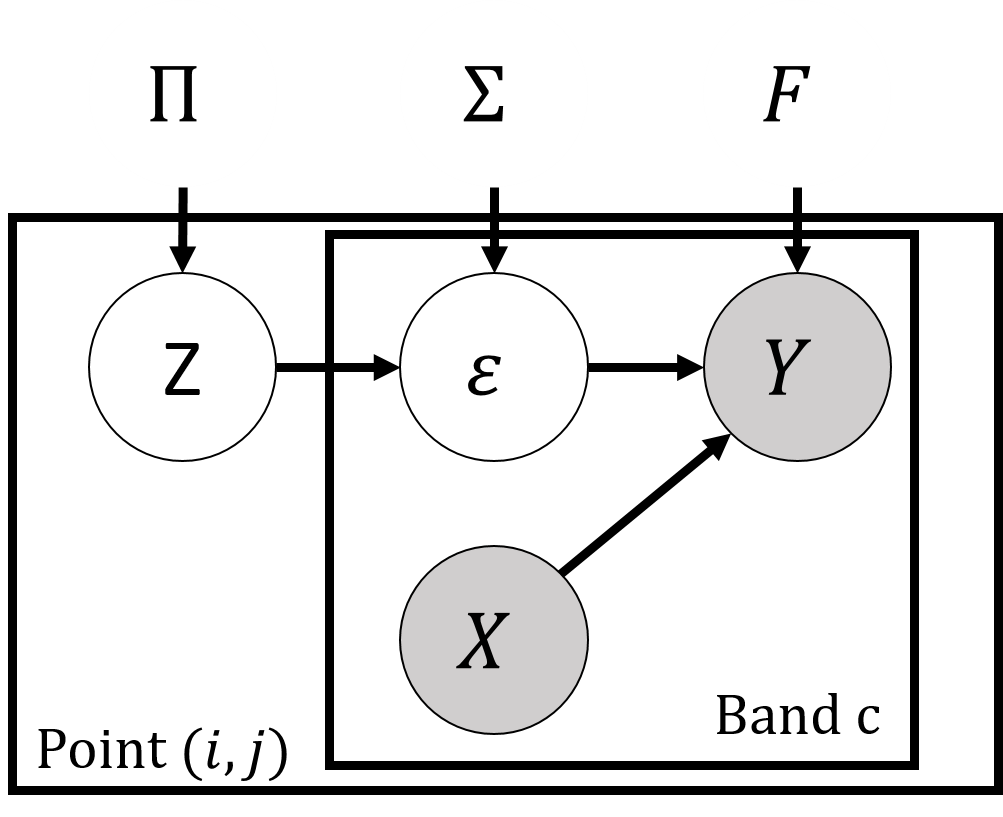}
\caption{Probabilistic graph model of the random variables }
\label{fig:Probabilistic graph model of the random variables}
\end{figure}
\subsubsection{Marginal log-likelihood}
 It yields the conditional distribution of $Y_1,\cdots,Y_C|X_1,\cdots,X_C$ as follows,
\begin{eqnarray}\label{likelihood model}
 & & p(Y_1,\cdots,Y_C|X_1,\cdots,X_C,\Pi,\Sigma,F) \\
  &=& \sum_{k=1}^{K}(\prod_{c=1}^{C}p(Y_c|X_c,Z_k =1,F,\Sigma))p(Z_k = 1|\Pi) \\
   &\overset{case\ 1(3)}{=}&\sum_{k=1}^{K}\pi_k \prod_{c=1}^{C}\N(Y_c|f_c(X_c),{\sigma_c}_k^2)\\
   &\overset{case\ 2}{=}&\sum_{k=1}^{K}\pi_k \prod_{c=1}^{C}\L(Y_c|f_c(X_c),{\beta_c}_k).
\end{eqnarray}

Consider the situation for all the points in the overlap area with index $(i,j)\in \Omega$ and each point is with $C$ bands values, the complete probabilistic graph model(\cite{koller2009probabilistic}) illustrating the dependence relations of the random variables are shown in figure~\ref{fig:Probabilistic graph model of the random variables}.

Then we can formulate a maximum log-likelihood model to obtain the RRN mapping $F$, parameters $\Sigma$ and $\Pi$  as the following,
\begin{eqnarray}
  F^*,\Pi^*,\Sigma^* &=& \arg\max\limits_{F,\Pi,\Sigma} \log p(Y^{total}|X^{total},\Pi,\Sigma,F) \\
   &=&  \arg\max\limits_{F,\Pi,\Sigma} \log \prod_{(i,j)\in\Omega} p(y_{ij1}\cdots y_{ijC}|x_{ij1}\cdots x_{ijC},\Pi,\Sigma,F) \\
   &\overset{case\ 1(3)}{=}& \arg\max\limits_{F,\Pi,\Sigma}\log \prod_{(i,j)\in\Omega} \sum_{k=1}^{K}\pi_k \prod_{c=1}^{C}\N(y_{ijc}|f_c(x_{ijc}),{\sigma_c}_k^2)\\
   &\overset{case\ 2}{=}& \arg\max\limits_{F,\Pi,\Sigma}\log \prod_{(i,j)\in\Omega} \sum_{k=1}^{K}\pi_k \prod_{c=1}^{C}\L(y_{ijc}|f_c(x_{ijc}),{\beta_c}_k),
\end{eqnarray}\label{likelihood model}
where $Y^{total}=\{y_{ijc}|(i,j)\in\Omega \ c\in \{1,\cdots,C\}\}$, $X^{total}=\{x_{ijc}|(i,j)\in\Omega \ c\in \{1,\cdots,C\}\}$, $y_{ijc}$ and $x_{ijc}$ are the $c$ band value of point $(i,j)$ of target and source image respectively.

However, it's hard to find a closed form solution of the maximum likelihood objective. The typical method to solve this kind of objective is the Expectation Maximization(EM) algorithm(\cite{dempster1977maximum}).
\subsection{EM algorithm}
EM algorithm maximizes the likelihood objective by uplifting a surrogate objective induced by the latent random variables, $Z^{total}=\{({Z_1}_{i,j},{Z_2}_{i,j})| (i,j)\in\Omega\}$:
\begin{eqnarray}\label{EM inequality}
  & &\log p(Y^{total}|X^{total},\Pi,\Sigma,F) \\&=&  \log \int p(Y^{total},Z^{total}|X^{total},\Pi,\Sigma,F)dZ \\
    &=&  \log \int q(Z^{total}) \frac{p(Y^{total},Z^{total}|X^{total},\Pi,\Sigma,F)}{q(Z^{total})}dZ \\
    &\overset{Jensen \ Inequality}{\ge}& \int q(Z^{total}) \log \frac{p(Y^{total},Z^{total}|X^{total},\Pi,\Sigma,F)}{q(Z^{total})}dZ\\
    &=& \E_{Z^{total}\sim q(Z^{total})}\log  \frac{p(Y^{total},Z^{total}|X^{total},\Pi,\Sigma,F)}{q(Z^{total})}\\
    &=& E(q,\Pi,\Sigma,F)
\end{eqnarray}
EM algorithm uses alternative search of $q$ and $\Pi,\Sigma,F$ to uplift $E(q,\Pi,\Sigma,F)$ to maximize the log likelihood.

The expectation step corresponds to the search of $q$ and to obtain new objective:
\begin{eqnarray}\label{E step  q obtain}
   &&  p(Z^{total}|X^{total},Y^{total},\Pi^{old},\Sigma^{old},f^{old}) \nonumber\\
   &=& \arg\max\limits_{q}\E_{Z^{total}\sim q(Z^{total})}\log  \frac{p(Y^{total},Z^{total}|X^{total},\Pi^{old},\Sigma^{old},f^{old})}{q(Z^{total})}.
\end{eqnarray}

The new objective $E(p(Z^{total}|X^{total},Y^{total},\Pi^{old},\Sigma^{old},f^{old}),\Pi,\Sigma,F)$ yields
\begin{eqnarray}\label{E step expectation}
  && E(p(Z^{total}|X^{total},Y^{total},\Pi^{old},\Sigma^{old},F^{old}),\Pi,\Sigma,F) \\
   &=& \E_{Z\sim p(Z^{total}|X^{total},Y^{total},\Pi^{old},\Sigma^{old},F^{old})}\log  p(Y^{total},Z^{total}|X^{total},\Pi,\Sigma,F).
\end{eqnarray}

The maximization step corresponds to the search of $\Pi,\Sigma,F$:
\begin{eqnarray}
    && \Pi^{new},\Sigma^{new},F^{new} \\
  &=& \arg\max\limits_{\Pi,\Sigma,F}\E_{Z^{total}\sim p(Z^{total}|X^{total},Y^{total},\Pi^{old},\Sigma^{old},f^{old})}\log  p(Y^{total},Z^{total}|X^{total},\Pi,\Sigma,F).
\end{eqnarray}\label{M step search}
By assigning $\Pi^{old},\Sigma^{old},F^{old}=\Pi^{new},\Sigma^{new},F^{new}$ and repeating the EM steps, the parameters will converge to a critic point of the original log likelihood objective under moderate conditions(\cite{wu1983convergence}).

In the expectation step, we first compute the following:
\begin{eqnarray}\label{expectation}
  &&{\gamma_k}_{ij}\\
   &=& \E_{Z^{total}\sim p(Z^{total}|Y^{total},X^{total},\Pi^{old},\Sigma^{old},F^{old})}{Z_k}_{ij} \\
    &=& \sum_{v=0}^{1}v p({Z_k}_{ij}=v|Y^{total},X^{total},\Pi^{old},\Sigma^{old},F^{old}) \\
    &=& p({Z_k}_{ij}=1|Y^{total},Y^{total},\Pi^{old},\Sigma^{old},F^{old}) \\
    &\overset{case\ 1(3)}{=}& \frac{\pi_k^{old} \prod_{c=1}^{C}\N(y_{ijc}|f_c^{old}(x_{ijc}),{\sigma_c^{old}}_k^2)}{\sum_{k=1}^{K}\pi_k^{old} \prod_{c=1}^{C}\N(y_{ijc}|f_c^{old}(x_{ijc}),{\sigma_c^{old}}_k^2)} \quad k\in\{1,2\}\\
    &\overset{case\ 2}{=}& \frac{\pi_k^{old} \prod_{c=1}^{C}\L(y_{ijc}|f_c^{old}(x_{ijc}),{\beta_c^{old}}_k)}{\sum_{k=1}^{K}\pi_k^{old} \prod_{c=1}^{C}\L(y_{ijc}|f_c^{old}(x_{ijc}),{\beta_c^{old}}_k)} \quad k\in\{1,2\}.
\end{eqnarray}
The ${\gamma_k}_{ij}$ is the posterior distribution of latent random variable ${Z_k}_{ij}$. ${\gamma_1}_{ij}$ indicates the posterior probability of the point $(i,j)$ being a no change point. Next we have,
\begin{eqnarray}\label{expectation objective}
  &&\E_{Z^{total}\sim p(Z^{total}|Y^{total},X^{total},\Pi^{old},\Sigma^{old},F^{old})}\log  p(Y^{total},Z^{total}|X^{total},\Pi,\Sigma,F)\\
   &\overset{case\ 1}{=}& \E_{Z^{total}\sim p(Z^{total}|Y^{total},X^{total},\Pi^{old},\Sigma^{old},F^{old})}\sum_{(i,j)\in \Omega}\log   \prod_{k=1}^{K} (\pi_k\prod_{c=1}^{C}\N(y_{ijc}|f_c(x_{ijc}),{\sigma_c}_k^2))^{{Z_k}_{ij}} \\
    &\overset{case\ 1}{=}& \E_{Z^{total}\sim p(Z^{total}|Y^{total},X^{total},\Pi^{old},\Sigma^{old},f^{old})}\sum_{(i,j)\in \Omega}\sum_{k=1}^{K}{{Z_k}_{ij}}(\log\pi_k+\sum_{c=1}^{C}\log (\N(y_{ijc}|f_c(x_{ijc}),{\sigma_c}_k^2))) \\
    &\overset{case\ 1}{=}& \sum_{(i,j)\in \Omega}\sum_{k=1}^{K}{\gamma_k}_{ij}(\log\pi_k+\sum_{c=1}^{C}\log (\N(y_{ijc}|f_c(x_{ijc}),{\sigma_c}_k^2))) \\
    &\overset{case\ 1(3)}{=}& \sum_{(i,j)\in \Omega}\sum_{k=1}^{K}{\gamma_k}_{ij}(\log\pi_k+\sum_{c=1}^{C}(-\frac{\log 2\pi +\log{\sigma_c}_k^2}{2}-\frac{(y_{ijc}-f_c(x_{ijc}))^2}{2{\sigma_c}_k^2}))\label{eqn:case 1 objective}\\
    &\overset{case\ 2}{=}& \sum_{(i,j)\in \Omega}\sum_{k=1}^{K}{\gamma_k}_{ij}(\log\pi_k+\sum_{c=1}^{C}(-\log 2{\beta_c}_k-\frac{\vert y_{ijc}-f_c(x_{ijc})\vert}{{\beta_c}_k}))\label{eqn:case 2 objective}.
\end{eqnarray}

Then with the help of alternative search for $\Pi,\Sigma$ and $F$, we can obtain the closed form solution for $\Pi,\Sigma$:
 \begin{eqnarray}
                          N_k^{new} &=&\sum_{(i,j)\in \Omega}{\gamma_k}_{ij}  \\
                          \pi_k^{new} &=& \frac{N_k^{new}}{\vert\Omega\vert} \\
                       case\ 1(3) \ :   {{\sigma_c}_k^2}^{new} &=& \frac{\sum_{(i,j)\in \Omega}{\gamma_k}_{ij}(y_{ijc}-f_c^{old}(x_{ijc}))^2}{N_k^{new}}\label{eqn:gaussian updator} \quad k\in\{1,2\}\\
                       case\ 2 \ :   {{\beta_c}_k}^{new} &=& \frac{\sum_{(i,j)\in \Omega}{\gamma_k}_{ij} \vert y_{ijc}-f_c^{old}(x_{ijc})\vert}{N_k^{new}}\label{eqn:laplace updator} \quad k\in\{1,2\}.
                        \end{eqnarray}

As for solving $F$, case 1 yields the weighted least square regime while case 2 yields the weighted histogram matching regime under moderate conditions.
\subsubsection{Case 1: weighted least square relative radiometric normalization regime}
The component in equation~\ref{eqn:case 1     objective} related to $F$ can be written as follows:
\begin{eqnarray}
  &&\sum_{(i,j)\in \Omega}\sum_{k=1}^{K}{\gamma_k}_{ij}(\sum_{c=1}^{C}(-\frac{(y_{ijc}-f_c(x_{ijc}))^2}{2{\sigma_c}_k^2})) \\ &=& -\sum_{c=1}^{C}\sum_{(i,j)\in \Omega}\sum_{k=1}^{K}(\frac{{\gamma_k}_{ij}}{2{\sigma_c}_k^2})(y_{ijc}-f_c(x_{ijc}))^2 \\
   &=&  -\sum_{c=1}^{C}\sum_{(i,j)\in \Omega}\sum_{k=1}^{K}(\frac{{\gamma_k}_{ij}}{2{\sigma_c}_k^2})(y_{ijc}-w_cx_{ijc}-b_c)^2.\label{eq:weighted least square}
\end{eqnarray}

Let $A_c = \begin{bmatrix} x_{i_1j_1c} & 1 \\ \cdots & \cdots \\ x_{i_{\vert \Omega\vert}j_{\vert \Omega \vert}c} & 1 \end{bmatrix}$, $D_c = \begin{bmatrix} y_{i_1j_1c} \\ \cdots \\ y_{i_{\vert \Omega\vert}j_{\vert \Omega \vert}c} \end{bmatrix}$,$G_c = \begin{bmatrix} \sum_{k=1}^{K}(\frac{{\gamma_k}_{i_1j_1}}{2{\sigma_c}_k^2}) \\ \cdots \\ \sum_{k=1}^{K}(\frac{{\gamma_k}_{i_{\vert \Omega \vert}j_{\vert \Omega \vert}}}{2{\sigma_c}_k^2}) \end{bmatrix}$ and $diag(G_c)$ be the square matrix with diagonal elements equaling $G_c$.
Since the objective in equation~\ref{eq:weighted    least square} can be divided into $C$ independent parts in regard to $c$, if $(A_c^T diag(G_c) A_c)$ is invertible, it yields that \begin{eqnarray}
               \begin{bmatrix} w_c^{new} \\  b_c^{new} \end{bmatrix}  &=& \arg\min\limits_{w_c,b_c} \sum_{(i,j)\in \Omega}\sum_{k=1}^{K}(\frac{{\gamma_k}_{ij}}{2{\sigma_c}_k^2})(y_{ijc}-w_cx_{ijc}-b_c)^2\\
                  &=& (A_c^T diag(G_c) A_c)^{-1}A_c^T diag(G_c)D_c, \quad c \in \{1,\cdots,C\},
               \end{eqnarray}
where $A_c^T$ is the transpose of $A_c$ and $^{-1}$ represents the inverse (or pseudo inverse). Here we obtain the mapping functions $F$.

Theoretically, the alternative search optimization step in the maximization step needs to be repeated several times until $\Pi,\Sigma$ and $F$ converge. However, empirically, it works fine if we compute $\Pi,\Sigma$ and $F$ once and directly move to the expectation part in the EM procedure. In order to speed up the optimization for $\Pi,\Sigma,F$, we use the stochastic EM procedure (i.e., only use a random subset $\Omega_{sub}\subseteq \Omega$ to implement each EM procedure) ,and we use the whole points to implement EM algorithm in the final turn.

\begin{algorithm}[htb]
\caption{ No-change set linear relative radiometric normalization via mixture of Gaussian noise modeling. (L-RRN-MoG)}
\label{alg::conjugateGradient}
\begin{algorithmic}[1]
\Require
$(X^{total},Y^{total})$: paired band values of junction point in source and target images;
\Ensure
$F^{new}$, $\Pi^{new}$, $\Sigma^{new}$
\State initial $F^{new}$, $\Pi^{new}$ and $\Sigma^{new}$, $\delta$;
\Repeat
\State $F^{old},\Pi^{old},\Sigma^{old} = F^{new}$, $\Pi^{new}$, $\Sigma^{new}$
\State \textbf{E step:} compute $\gamma_{kij}=\frac{\pi_k^{old} \prod_{c=1}^{C}\N(y_{ijc}|f_c^{old}(x_{ijc}),{\sigma_c^{old}}_k^2)}{\sum_{k=1}^{K}\pi_k^{old} \prod_{c=1}^{C}\N(y_{ijc}|f_c^{old}(x_{ijc}),{\sigma_c^{old}}_k^2)} $ ;
\State \textbf{M step:} compute $\Pi^{new}$,$\Sigma^{new}$:

 $N_k^{new} =\sum_{(i,j)\in \Omega}{\gamma_k}_{ij}$

 $\pi_k^{new} = \frac{N_k^{new}}{\vert\Omega\vert}$

 ${{\sigma_c}_k^2}^{new} = \frac{\sum_{(i,j)\in \Omega}{\gamma_k}_{ij}(y_{ijc}-f_c^{old}(x_{ijc}))^2}{N_k^{new}}$

\State \textbf{\quad \quad \quad \ } compute $F^{new}$:

$\begin{bmatrix} w_c^{new} \\  b_c^{new} \end{bmatrix}  =  (A_c^T diag(G_c) A_c)^{-1}A_c^T diag(G_c)D_c$

\Until{($\vert \log p(Y^{total}|X^{total},\Pi^{new},\Sigma^{new},F^{new})-\log p(Y^{total}|X^{total},\Pi^{old},\Sigma^{old},F^{old})\vert<\delta$)}
\end{algorithmic}
\end{algorithm}

\subsubsection{Case 2: weighted histogram matching relative radiometric normalization regime}

The component in equation~\ref{eqn:case 2 objective} related to $F$ can be written as follows:

\begin{eqnarray}
  &&\sum_{(i,j)\in \Omega}\sum_{k=1}^{K}{\gamma_k}_{ij}(\sum_{c=1}^{C}(-\frac{\vert y_{ijc}-f_c(x_{ijc})\vert}{{\beta_c}_k})) \\
  &=& -\sum_{c=1}^{C}\sum_{(i,j)\in \Omega}(\sum_{k=1}^{K}\frac{{\gamma_k}_{ij}}{{\beta_c}_k})\vert y_{ijc}-f_c(x_{ijc})\vert .\label{eq:weighted histogram matching}
\end{eqnarray}
The objective in equation~\ref{eq:weighted histogram matching} can be divided into C independent parts in regard to c. We only analyze one part. In order to give an approximation to the solution of $f_c$ in this part, we first suppose the whole data satisfy the following assumption, and later we will discuss the situation that disobeys the assumption.
\paragraph{Equivalence of histogram matching and L1-norm image regression}{
\begin{assumption}\label{assumption}\textbf{Color Sort Assumption}.
$x_{i_sj_sc}\le x_{i_oj_oc}$  implies $y_{i_sj_sc}\le y_{i_oj_oc}$ for $c\in\{1,\cdots,C\}, \quad s,o\in \{1,\cdots,\vert\Omega\vert\}$.
\end{assumption}

This assumption claims that the order of the band value should keep in the shared points in both source and target images.

If assumption~\ref{assumption} holds, since $f_c$ is monotone mapping, $x_{i_sj_sc}\le x_{i_tj_tc}$ also implies $f_c(x_{i_sj_sc})\le f_c(x_{i_tj_tc})$. Then we could rearrange the index of points such that $x_{i_1j_1c}\le \cdots \le x_{i_{\vert \Omega \vert}j_{\vert \Omega \vert}c}$, $f_c(x_{i_1j_1c})\le \cdots \le f_c(x_{i_{\vert \Omega \vert}j_{\vert \Omega \vert}c})$, and $y_{i_1j_1c}\le \cdots \le y_{i_{\vert \Omega \vert}j_{\vert \Omega \vert}c}$.

 Suppose $f_c(X_c),Y_c,X_c$ take value on $\{0,\cdots,T\}$, $g_{ijc} = \sum_{k=1}^{K}\frac{{\gamma_k}_{ij}}{{\beta_c}_k}$, ${\Omega_t}_c^f = \{(i,j)|f_c(x_{ijc})=t\}$, ${\Omega_t}_c^y = \{(i,j)|y_{ijc}=t\}$, the weighted value histogram for $f_c(X_c)$ is  $histogram_c^f=({n^f_0}_c,\cdots,{n^f_T}_c)$ where ${n^f_t}_c= \sum_{(i,j) \in {\Omega_t}_c^f}g_{ijc}$. Similarly, we have $histogram_c^y=({n^y_0}_c,\cdots,{n^y_T}_c)$ where ${n^y_t}_c= \sum_{(i,j) \in {\Omega_t}_c^y}g_{ijc}$.

Then the equivalence of match distance(\cite{werman1985distance})\footnote{The match distance is also the objective of the histogram matching task.} of histograms and the objective is demonstrated as  following:
\begin{eqnarray}
 && d(histogram_c^f,histogram_c^y) \\
 &\triangleq& \sum_{t=0}^{T}\vert \sum_{u=0}^{t}{n^f_u}_c -\sum_{u=0}^{t}{n^y_u}_c\vert  \\
      &=& \sum_{t=0}^{T}\vert \sum_{u=0}^{t}(\sum_{(i,j)|f_c(x_{ijc})=u}g_{ijc}-\sum_{(i,j)|y_{ijc}=u}g_{ijc})\vert  \\
          &=&  \sum_{t=0}^{T}\vert \sum_{(i,j)|f_c(x_{ijc})=\{0,\cdots,t\}}g_{ijc}-\sum_{(i,j)|y_{ijc}=\{0,\cdots,t\}}g_{ijc}\vert\label{eq:sum1}\\
    &=& \sum_{s=1}^{\vert \Omega\vert}g_{i_sj_sc}\vert y_{i_sj_sc}-f_c(x_{i_sj_sc})\vert \label{eq:sum2}
\end{eqnarray}
The equivalence of equation~\ref{eq:sum1} and equation~\ref{eq:sum2} is shown in the figure~\ref{fig:sum1} and figure~\ref{fig:sum2}. They correspond to integration among different axes.

\begin{figure}[ht]
\centering
\includegraphics[width=13cm]{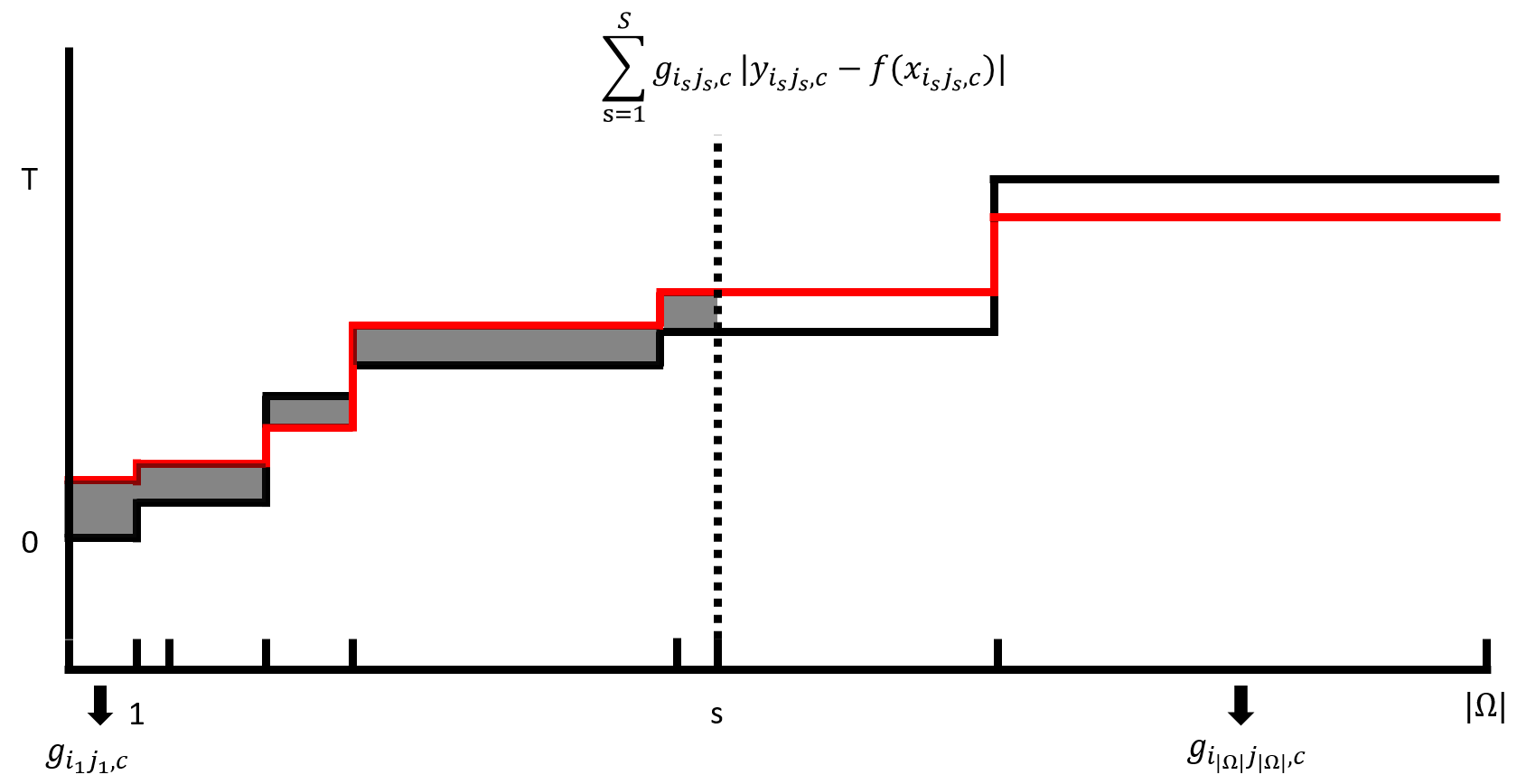}
\caption{Equation~\ref{eq:sum1} area demonstration. The black line corresponds to the order sequence $f_c(x_{i_1j_1c})\le \cdots \le f_c(x_{i_{\vert \Omega \vert}j_{\vert \Omega \vert}c})$ and the red line corresponds to the order sequence $y_{i_1j_1c}\le \cdots \le y_{i_{\vert \Omega \vert}j_{\vert \Omega \vert}c}$. The gray area is the sum of $\sum_{s=1}^{S}g_{i_sj_sc}\vert y_{i_sj_sc}-f_c(x_{i_sj_sc})\vert$. The equation~\ref{eq:sum1} is summation according to point index axis.}
\label{fig:sum1}
\end{figure}
\begin{figure}[ht]
\centering
\includegraphics[width=13cm]{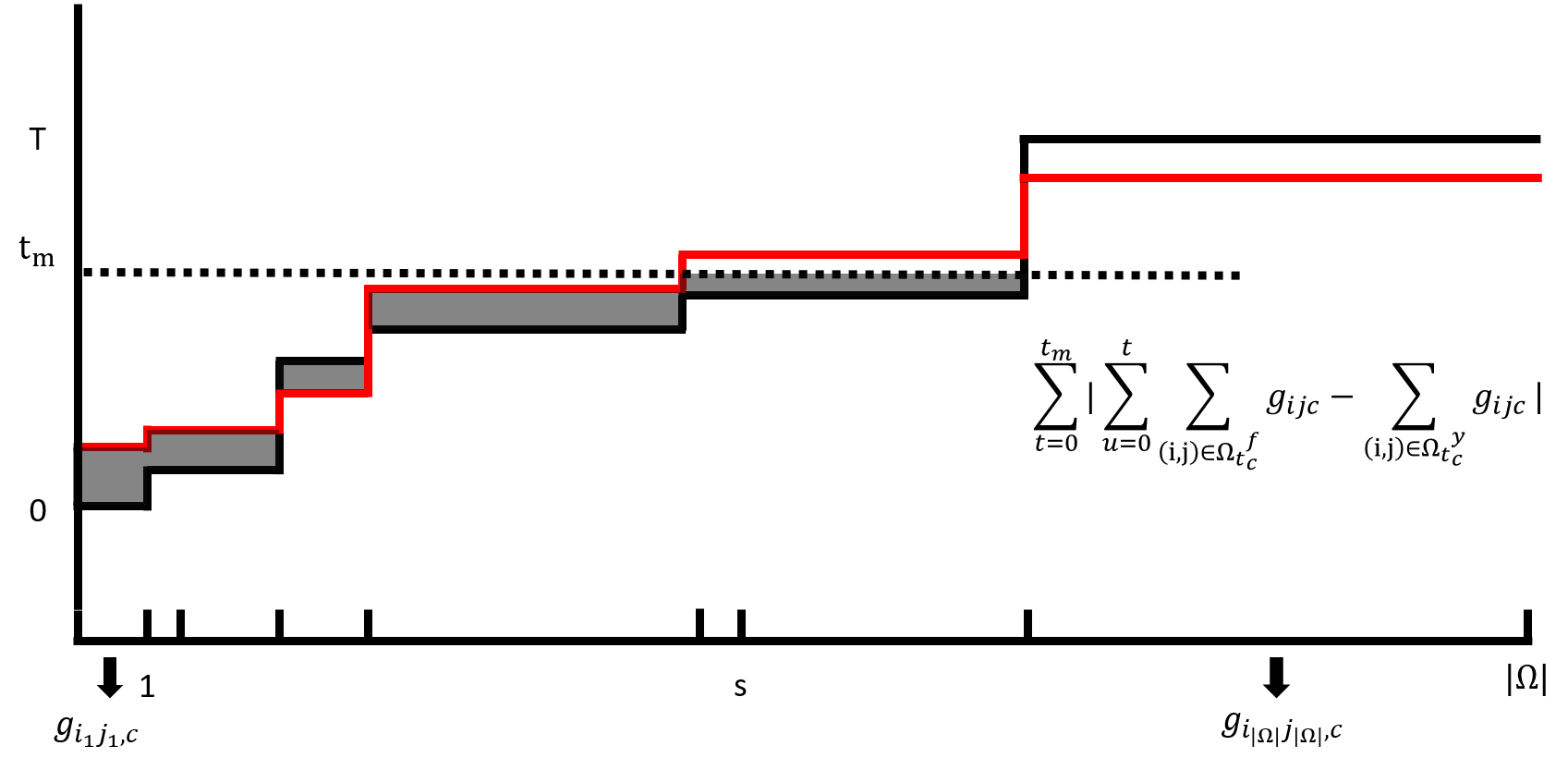}
\caption{Equation~\ref{eq:sum2} area demonstration. The black line corresponds to the order sequence $f_c(x_{i_1j_1c})\le \cdots \le f_c(x_{i_{\vert \Omega \vert}j_{\vert \Omega \vert}c})$ and the red line corresponds to the order sequence $y_{i_1j_1c}\le \cdots \le y_{i_{\vert \Omega \vert}j_{\vert \Omega \vert}c}$. The gray area is the sum of $\sum_{t=0}^{t_m}\vert \sum_{(i,j)|x_{ijc}=\{0,\cdots,t\}}g_{ijc}-\sum_{(i,j)|y_{ijc}=\{0,\cdots,t\}}g_{ijc}\vert$.  The equation~\ref{eq:sum2} is summation according to value axis.}
\label{fig:sum2}
\end{figure}

Suppose that $CP$ is the cumulative probability function of the histogram. $CP_c^u=\frac{1}{\sum_{l=0}^{T}{n^u_T}_c}(\sum_{l=0}^{0}{n^u_0}_c,\cdots,\sum_{l=0}^{T}{n^u_T}_c)$ $u\in\{x,y,f\}$.
Notice that
\begin{eqnarray}
               && \frac{d(histogram_c^f,histogram_c^y)}{\sum_{s=1}^{\vert\Omega\vert}g_{i_sj_sc}}\\
               &=&  \sum_{t=0}^{T}\vert \sum_{u=0}^{t}\frac{{n^f_u}_c }{\sum_{s=1}^{\vert\Omega\vert}g_{i_sj_sc}} -\sum_{u=0}^{t}\frac{{n^y_u}_c}{\sum_{s=1}^{\vert\Omega\vert}g_{i_sj_sc}}\vert\\
               &=& \sum_{t=0}^{T}\vert CP^f_c(t) - CP^y_c(t)\vert.
            \end{eqnarray}
This is the objective of histogram matching problem, the solution for $f_c$ is the following: \begin{equation}\label{eq:histogram matching solution}
  f_c^{new}: t\rightarrow \arg\min_{i}\vert CP^y_c(i)- CP^x_c(t)\vert.
\end{equation}.

Since discrete monotone $f_c$ that minimizes $\frac{d(histogram_c^f,histogram_c^y)}{\sum_{s=1}^{\vert\Omega\vert}g_{i_sj_sc}}$ also minimizes $\sum_{s=1}^{\vert \Omega\vert}g_{i_sj_sc}\vert y_{i_sj_sc}-f_c(x_{i_sj_sc})\vert$, therefore
\begin{equation}\label{eq:F solution}
  F^{new} = [f_1^{new},\cdots,f_C^{new}] = \arg\max\limits_{f_1,\cdots,f_C}-\sum_{c=1}^{C}\sum_{(i,j)\in \Omega}(\sum_{k=1}^{K}\frac{{\gamma_k}_{ij}}{{\beta_c}_k})\vert y_{ijc}-f_c(x_{ijc})\vert.
\end{equation}
}
Although not all points might satisfy the color sort assumption, the no-change set intuitively should satisfy this assumption\footnote{Despite strong atmospheric effect, the radiometric induced by different sensors should satisfy this assumption in high probability.}. As for changed points, empirically they will be assigned higher probability to have higher noise level that the $\gamma_0<<\gamma_1$ and $\beta_0<<\beta_1$. Their weights $\frac{\gamma_0}{\beta_0}+\frac{\gamma_1}{\beta_1}<<\frac{0.5}{\beta_0}+\frac{0.5}{\beta_1}\le\frac{1}{\beta_0}$ is small and would not significantly influence the objective such that the objective is still similar to the objective of histogram matching. Therefore it's still reasonable to use histogram matching to solve $F$. Theoretically, the alternative search optimization step in maximization step needs to be repeated several time until $\Pi,\Sigma$ and $F$ converge. However, empirically, it works fine if we just compute $\Pi,\Sigma$ and $F$ once and directly move to expectation part in the EM procedure. In order to speed up the optimization for $\Pi,\Sigma,F$, we use the stochastic EM procedure (i.e. only use a random subset $\Omega_{sub}\subseteq \Omega$ to implement each EM procedure) and in the final turn we use the whole points to implement EM algorithm.

\begin{algorithm}[htb]
\caption{ No-change set histogram matching relative radiometric normalization via mixture of Laplace noise modeling. (HM-RRN-MoL)}
\label{alg::HM-RRN-MoL}
\begin{algorithmic}[1]
\Require
$(X^{total},Y^{total})$: paired band values of junction point in source and target images;
\Ensure
$F^{new}$, $\Pi^{new}$, $\Sigma^{new}$
\State initial $F^{new}$, $\Pi^{new}$ and $\Sigma^{new}$, $\delta$;
\Repeat
\State $F^{old},\Pi^{old},\Sigma^{old} = F^{new}$, $\Pi^{new}$, $\Sigma^{new}$
\State \textbf{E step:} compute $\frac{\pi_k^{old} \prod_{c=1}^{C}\L(y_{ijc}|f_c^{old}(x_{ijc}),{\beta_c^{old}}_k)}{\sum_{k=1}^{K}\pi_k^{old} \prod_{c=1}^{C}\L(y_{ijc}|f_c^{old}(x_{ijc}),{\beta_c^{old}}_k)} \quad k\in\{1,2\} $ ;
\State \textbf{M step:} compute $\Pi^{new}$,$\Sigma^{new}$:

 $N_k^{new} =\sum_{(i,j)\in \Omega}{\gamma_k}_{ij}$

 $\pi_k^{new} = \frac{N_k^{new}}{\vert\Omega\vert}$

 ${{\beta_c}_k}^{new} = \frac{\sum_{(i,j)\in \Omega}{\gamma_k}_{ij} \vert y_{ijc}-f_c^{old}(x_{ijc})\vert}{N_k^{new}}$

\State \textbf{\quad \quad \quad \ } compute $F^{new}$:

$g_{ijc} = \sum_{k=1}^{K}\frac{{\gamma_k}_{ij}}{{\beta_c}_k}$

${\Omega_t}_c^u = \{(i,j)|u_{ijc}=t\}$  ${n^u_t}_c= \sum_{(i,j) \in {\Omega_t}_c^u}g_{ijc}$ $u\in\{x,y\}$

$histogram_c^u=({n^u_0}_c,\cdots,{n^u_T}_c)$ $u\in\{x,y\}$

$CP_c^u=\frac{1}{\sum_{l=0}^{T}{n^u_T}_c}(\sum_{l=0}^{0}{n^u_0}_c,\cdots,\sum_{l=0}^{T}{n^u_T}_c)$ $u\in\{x,y\}$

$f_c^{new}: t\rightarrow \arg\min_{i}\vert CP^y_c(i)- CP^x_c(t)\vert$

\Until{($\vert \log p(Y^{total}|X^{total},\Pi^{new},\Sigma^{new},F^{new})-\log p(Y^{total}|X^{total},\Pi^{old},\Sigma^{old},F^{old})\vert<\delta$)}
\end{algorithmic}
\end{algorithm}

\subsubsection{Case 3: weighted least square histogram matching relative radiometric normalization regime}
The component in equation~\ref{eqn:case 1     objective} related to $F$ can be written as follows:
\begin{eqnarray}
  &&\sum_{(i,j)\in \Omega}\sum_{k=1}^{K}{\gamma_k}_{ij}(\sum_{c=1}^{C}(-\frac{(y_{ijc}-f_c(x_{ijc}))^2}{2{\sigma_c}_k^2})) \\ &=& -\sum_{c=1}^{C}\sum_{(i,j)\in \Omega}\sum_{k=1}^{K}(\frac{{\gamma_k}_{ij}}{2{\sigma_c}_k^2})(y_{ijc}-f_c(x_{ijc}))^2.\label{eq:weighted least square histogram matching}
\end{eqnarray}
where $f_c$ is the monotone discrete mapping. The above objective is not easy to solve.
Let $g_{ijc} = \frac{{\gamma_1}_{ij}}{{\sigma_c}_1^2}$ and ${\Omega_t}_c^x = \{(i,j)|x_{ijc})=t\}$. The weighted value histogram for $X_c$ is  $histogram_c^x=({n^x_0}_c,\cdots,{n^x_T}_c)$ where ${n^x_t}_c= \sum_{(i,j) \in {\Omega_t}_c^x}g_{ijc}$. Similarly, we have $histogram_c^y=({n^y_0}_c,\cdots,{n^y_T}_c)$ where ${n^y_t}_c= \sum_{(i,j) \in {\Omega_t}_c^y}g_{ijc}$. Suppose that $CP$ is the cumulative probability function of the histogram. $CP_c^u=\frac{1}{\sum_{l=0}^{T}{n^u_T}_c}(\sum_{l=0}^{0}{n^u_0}_c,\cdots,\sum_{l=0}^{T}{n^u_T}_c)$ $u\in\{x,y,f\}$.

We treat the weighted histogram matching
\begin{equation}\label{eq:histogram matching solution}
  f_c^{new}: t\rightarrow \arg\min_{i}\vert CP^y_c(i)- CP^x_c(t)\vert,
\end{equation} as the "surrogate" solution for $f_c$.

Similar to the above two cases, we give the algorithm as the following:

\begin{algorithm}[htb]
\caption{ No-change set histogram matching relative radiometric normalization via mixture of Gaussian noise modeling. (HM-RRN-MoG)}
\label{alg::conjugateGradient}
\begin{algorithmic}[1]
\Require
$(X^{total},Y^{total})$: paired band values of junction point in source and target images;
\Ensure
$F^{new}$, $\Pi^{new}$, $\Sigma^{new}$
\State initial $F^{new}$, $\Pi^{new}$ and $\Sigma^{new}$, $\delta$;
\Repeat
\State $F^{old},\Pi^{old},\Sigma^{old} = F^{new}$, $\Pi^{new}$, $\Sigma^{new}$
\State \textbf{E step:} compute $\gamma_{kij}=\frac{\pi_k^{old} \prod_{c=1}^{C}\N(y_{ijc}|f_c^{old}(x_{ijc}),{\sigma_c^{old}}_k^2)}{\sum_{k=1}^{K}\pi_k^{old} \prod_{c=1}^{C}\N(y_{ijc}|f_c^{old}(x_{ijc}),{\sigma_c^{old}}_k^2)} $ ;
\State \textbf{M step:} compute $\Pi^{new}$,$\Sigma^{new}$:

 $N_k^{new} =\sum_{(i,j)\in \Omega}{\gamma_k}_{ij}$

 $\pi_k^{new} = \frac{N_k^{new}}{\vert\Omega\vert}$

 ${{\sigma_c}_k^2}^{new} = \frac{\sum_{(i,j)\in \Omega}{\gamma_k}_{ij}(y_{ijc}-f_c^{old}(x_{ijc}))^2}{N_k^{new}}$

\State \textbf{\quad \quad \quad \ } compute $F^{new}$:

$g_{ijc} = \frac{{\gamma_1}_{ij}}{{\sigma_c}_1^2}$

${\Omega_t}_c^u = \{(i,j)|u_{ijc}=t\}$  ${n^u_t}_c= \sum_{(i,j) \in {\Omega_t}_c^u}g_{ijc}$ $u\in\{x,y\}$

$histogram_c^u=({n^u_0}_c,\cdots,{n^u_T}_c)$ $u\in\{x,y\}$

$CP_c^u=\frac{1}{\sum_{l=0}^{T}{n^u_T}_c}(\sum_{l=0}^{0}{n^u_0}_c,\cdots,\sum_{l=0}^{T}{n^u_T}_c)$ $u\in\{x,y\}$

$f_c^{new}: t\rightarrow \arg\min_{i}\vert CP^y_c(i)- CP^x_c(t)\vert$
\Until{($\vert \log p(Y^{total}|X^{total},\Pi^{new},\Sigma^{new},F^{new})-\log p(Y^{total}|X^{total},\Pi^{old},\Sigma^{old},F^{old})\vert<\delta$) or $(iteration>10)$ }
\end{algorithmic}
\end{algorithm}


\section{Experiments and results}
We designed three kinds of experiments.

The first one is the relative radiometric normalization on complex cases such as cloudy and foggy images. Cloudy and foggy images are the most disturbing images when implementing relative radiometric normalization methods. This experiment demonstrates the auto robust characteristic of our method. Moreover, our method is evaluable regarding the log-likelihood or the variance/scale factor of the noise of the no-change set pixels.

The second experiment demonstrates that our method's relative radiometric normalization mapping can support vegetation change detection and water change detection.

The third experiment illustrates that the no-change set derived by our method can support the building change detection task by reducing the pseudo-change obtained by segmentation symmetry difference. Change detection is an important monitoring task. The radiometric no-change set produced by our algorithm is good to indicate the invariance of the object. Classification-based change detection methods only access one image per time and calculate the change of the classes, losing the joint information from the two images. The integration of no-change set into the classification-based change detection methods may alleviate the misclassification due to the light condition, atmospheric condition.
%
\subsection{Materials}

We implement our methods on one cloudy image and one non-cloudy image. The images are obtained by $GF1C$ satellite, located in the Shandong province of China. We performed the preprocessing steps of RPC orthorectification and geometric correction of these images. These images are multi-spectral, and they are shown in red, green, and blue. They are clipped to the range shown as the source and target images in the figure~\ref{fig:Comparison figures of 5 different methods part I}. We exchange the order of these two images to be the target and source image once each. We compare five different methods with its relative radiometric image and its derived no-change set.

We implement our methods on one foggy image and one non-foggy image.
The source image is obtained by $GF6$ satellite, located in Henan province of China. We performed the preprocessing steps of RPC orthorectification and geometric correction of this image and down-sample to around 32-meter resolution. The source image is multi-spectral, and we only select the normalized(ranged from 0-255) red, green and blue bands to implement RRN. It is shown in the figure~\ref{fig:Comparison figure 5 different methods part III}. The target image is an artificial template image of normalized red, green, and blue color. We implement the relative radiometric normalization on the foggy image. We compare five different methods with its relative radiometric image and its derived no-change set.

We implement the HM-RRN-MoG method on additional reflectance images.
The source image F and target image E are obtained by $GF1D$ satellite, located in China's Shandong province. F was acquired in April 2020, and E was acquired in March 2020. We performed the preprocessing steps of RPC orthorectification, geometric correction, radiometric calibration, and atmospheric correction of these images and extracted the overlapped area. These images are multi-spectral. As for change detection, we compared reflectance/digital number,  normalized difference vegetation index(NDVI), and normalized difference water index NDWI of image pairs (E, F), (E, HM-RRN-MoG F), (A, B) and (HM-RRN-MoG A, B) on the no-change set derived by HM-RRN-MoG method. The comparison results are shown in table~\ref{table:HM-RRN-MoG reducing reflectance/(radiance), NDVI, NDWI inconsistence on no-change set.}.

We use the LEVIR-CD (\cite{chen2020spatial}) dataset and building area segmentation model to show that no-change set can boost the classification-based change detection methods. There are 128 test image pairs in LEVIR-CD. The building area segmentation model is trained on the building area data of Xinjiang province, China, with the improved D-linknet structure(\cite{zhou2018d}).

In order to guarantee the HM-RRN-MoG method's stability we implement the band value normalization to 0-255 and then normalize it back to the scale of target image after the algorithm. The followings are the results.
\FloatBarrier
\subsection{Relative radiometric normalization on complex cases}

\subsubsection{Cloudy cases}

\begin{figure}[ht]
\centering
\captionsetup[subfigure]{font=scriptsize,labelfont=scriptsize}
\begin{subfigure}[t]{.15\textwidth}
	\centering
	\includegraphics[width=\textwidth]{figure/GF1C_PMS_552.jpg}
	\caption{Target image B}
\end{subfigure}
\begin{subfigure}[t]{.15\textwidth}
	\centering
	\includegraphics[width=\textwidth]{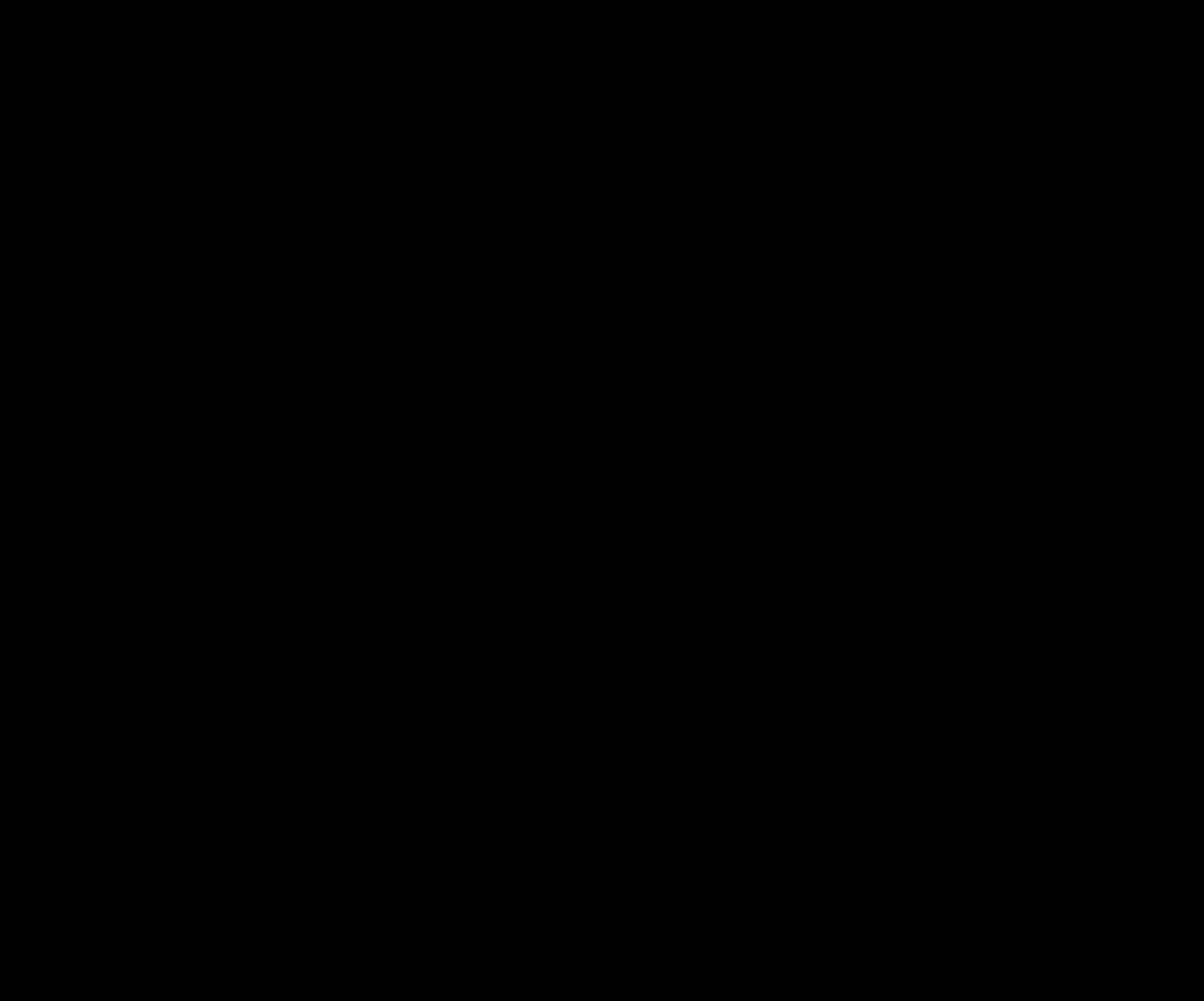}
	\caption{L-RRN NC}
\end{subfigure}
\begin{subfigure}[t]{.15\textwidth}
	\centering
\includegraphics[width=\textwidth]{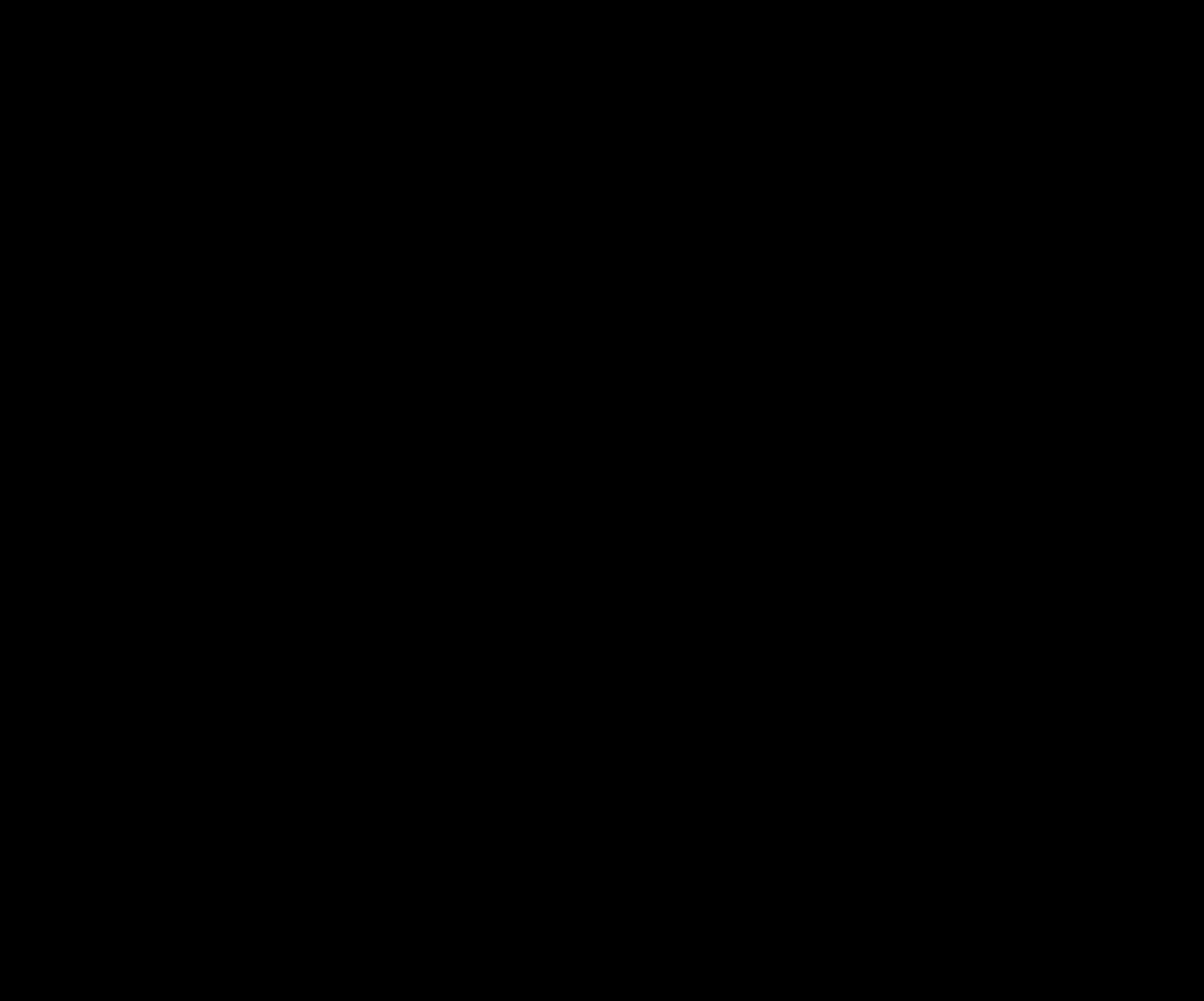}
	\caption{HM-RRN NC}
\end{subfigure}
\begin{subfigure}[t]{.15\textwidth}
	\centering
\includegraphics[width=\textwidth]{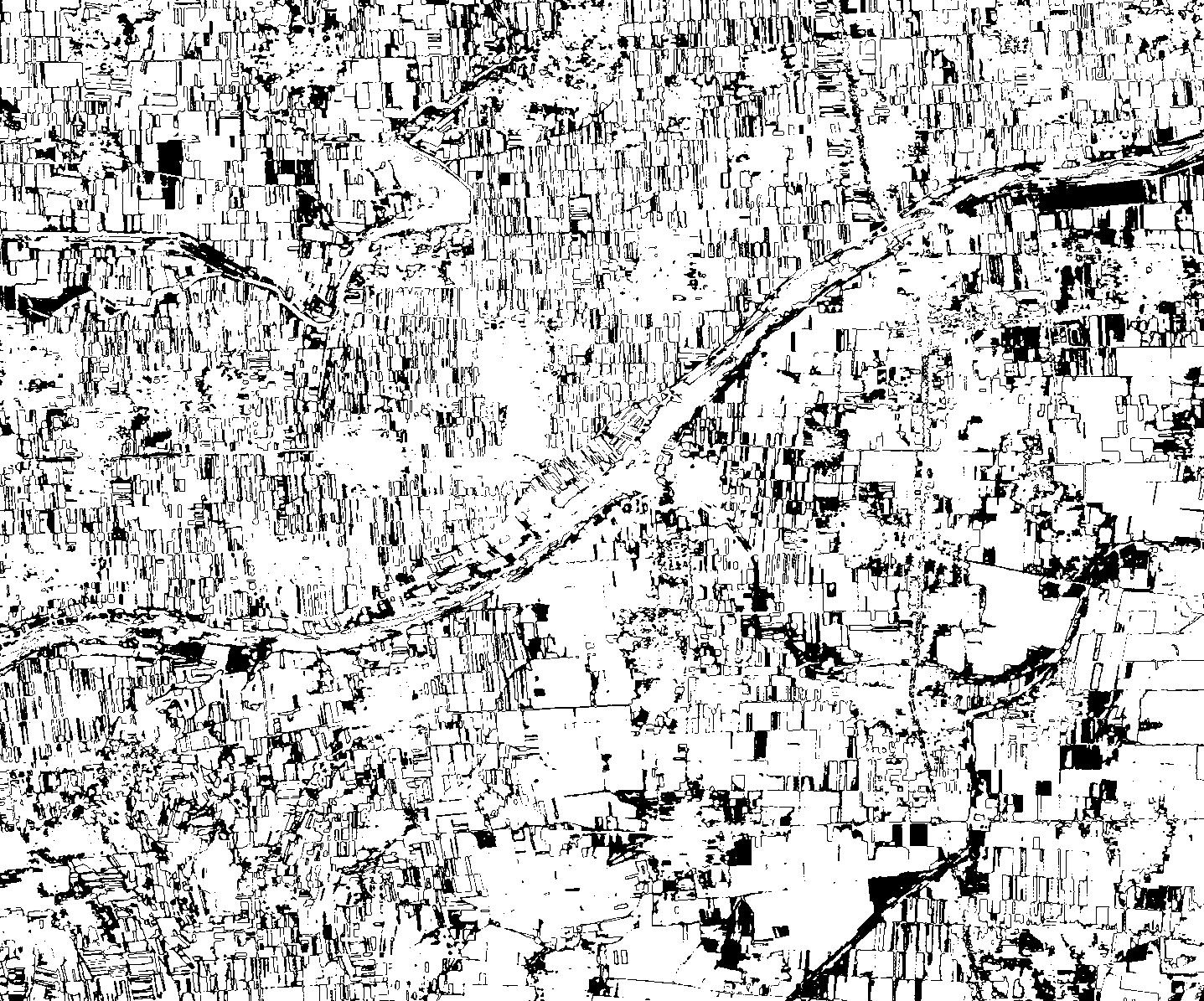}
	\caption{L-RRN-MoG NC}
\end{subfigure}
\begin{subfigure}[t]{.15\textwidth}
	\centering
\includegraphics[width=\textwidth]{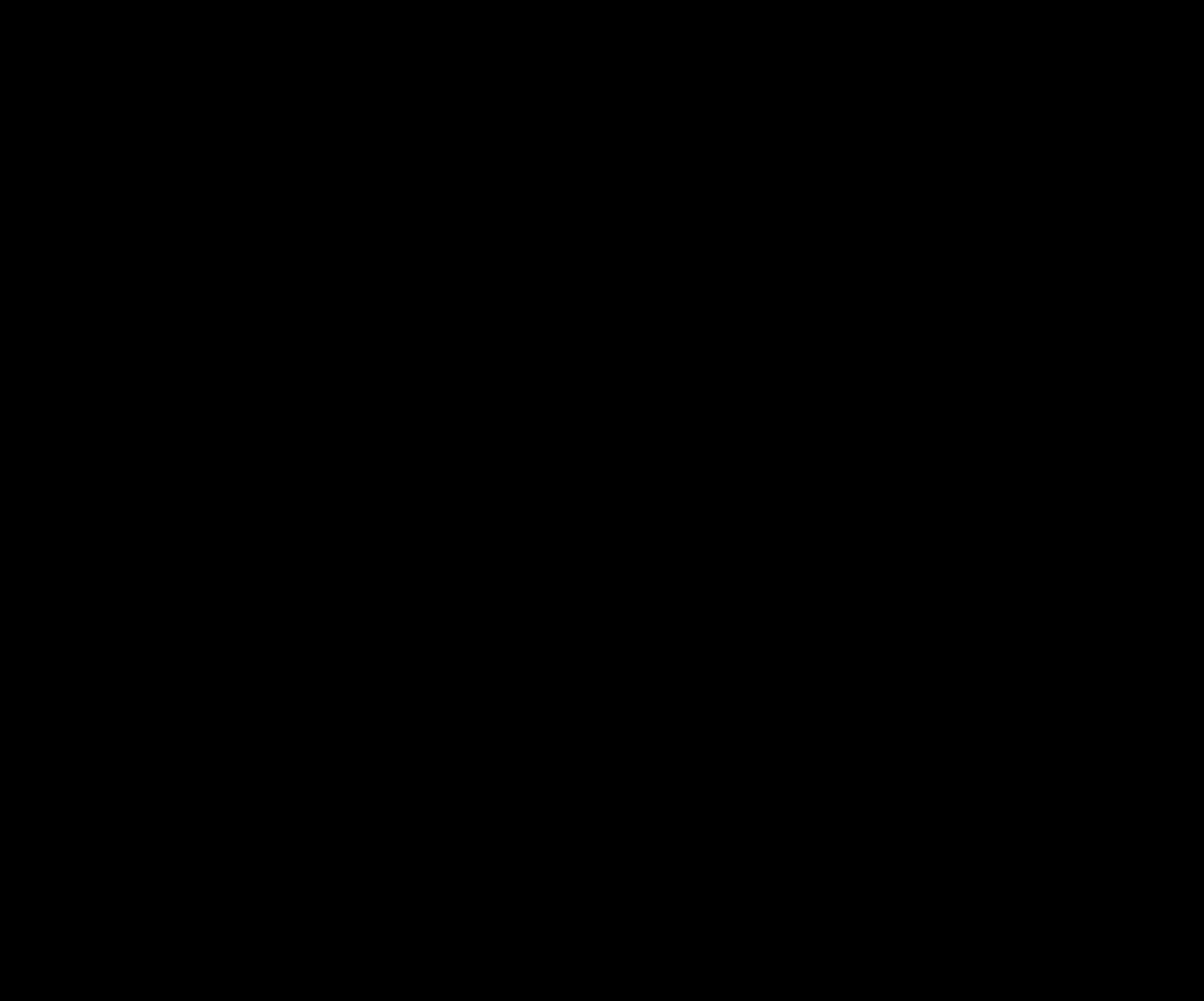}
	\caption{HM-RRN-MoL NC}
\end{subfigure}
\begin{subfigure}[t]{.15\textwidth}
	\centering
\includegraphics[width=\textwidth]{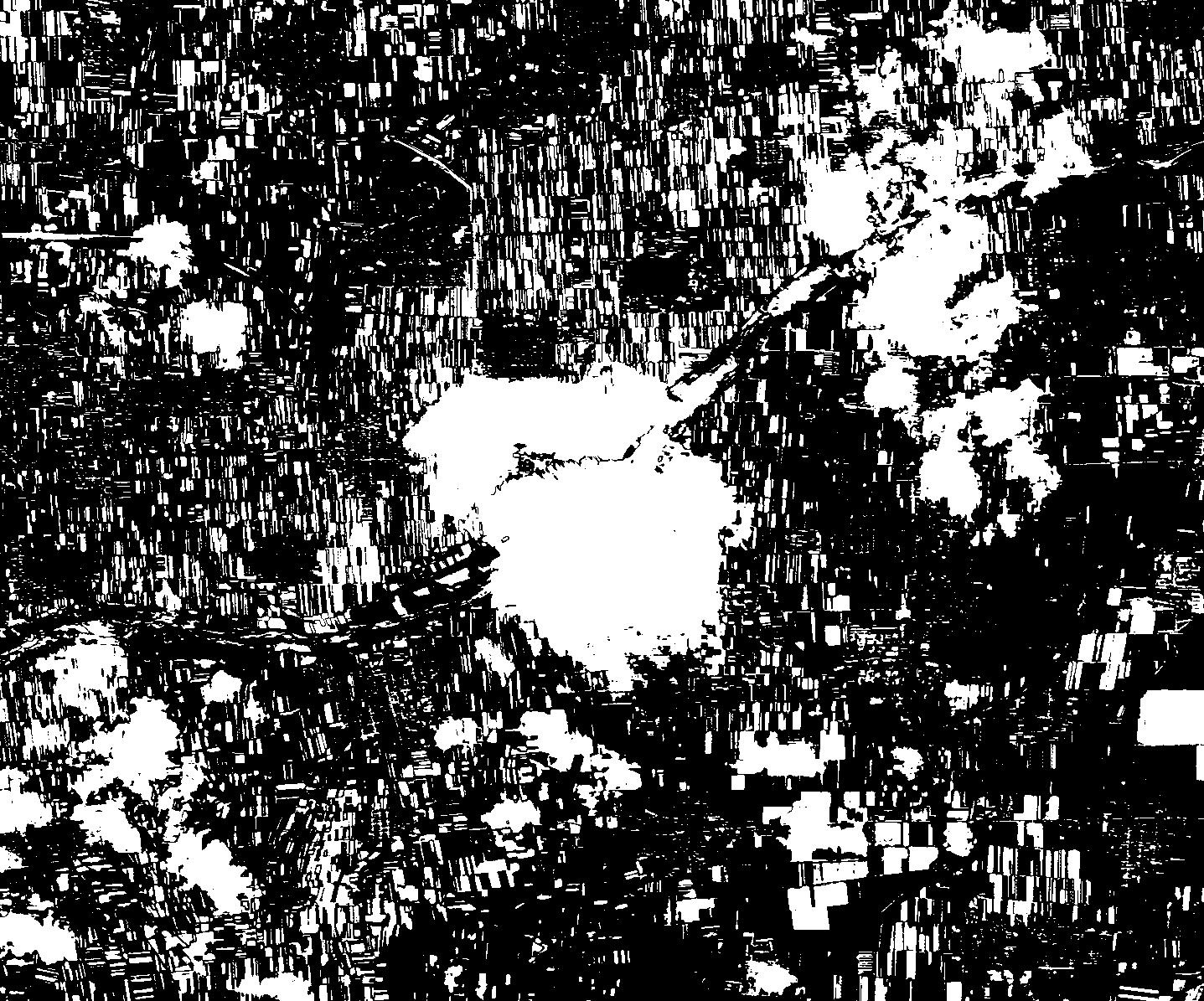}
	\caption{HM-RRN-MoG NC}
\end{subfigure}
\begin{subfigure}[t]{.15\textwidth}
	\centering
\includegraphics[width=\textwidth]{figure/GF1C_PMS_127.jpg}
	\caption{Source image A}
\end{subfigure}
\begin{subfigure}[t]{.15\textwidth}
	\centering
\includegraphics[width=\textwidth]{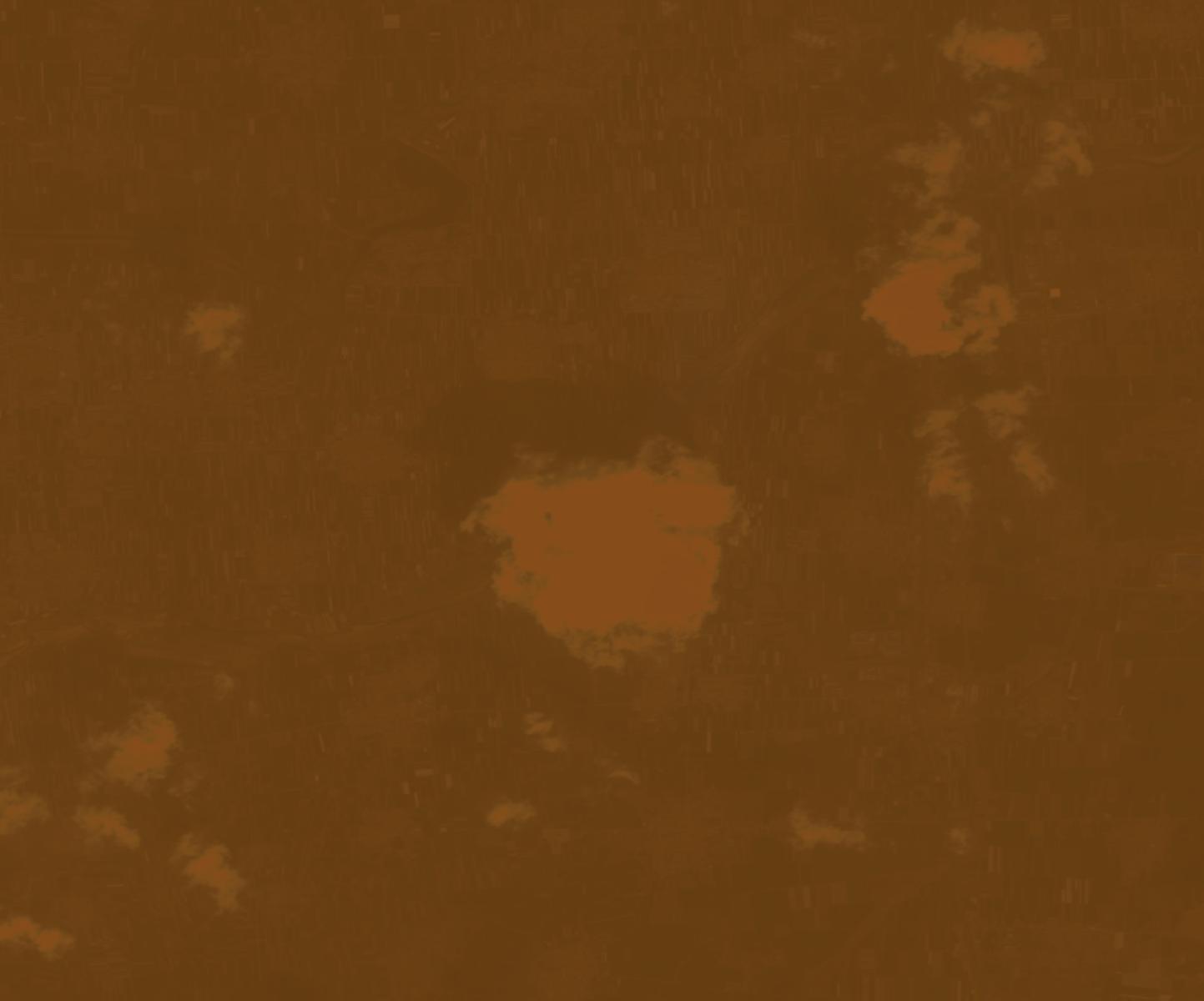}
	\caption{L-RRN A}
\end{subfigure}
\begin{subfigure}[t]{.15\textwidth}
	\centering
\includegraphics[width=\textwidth]{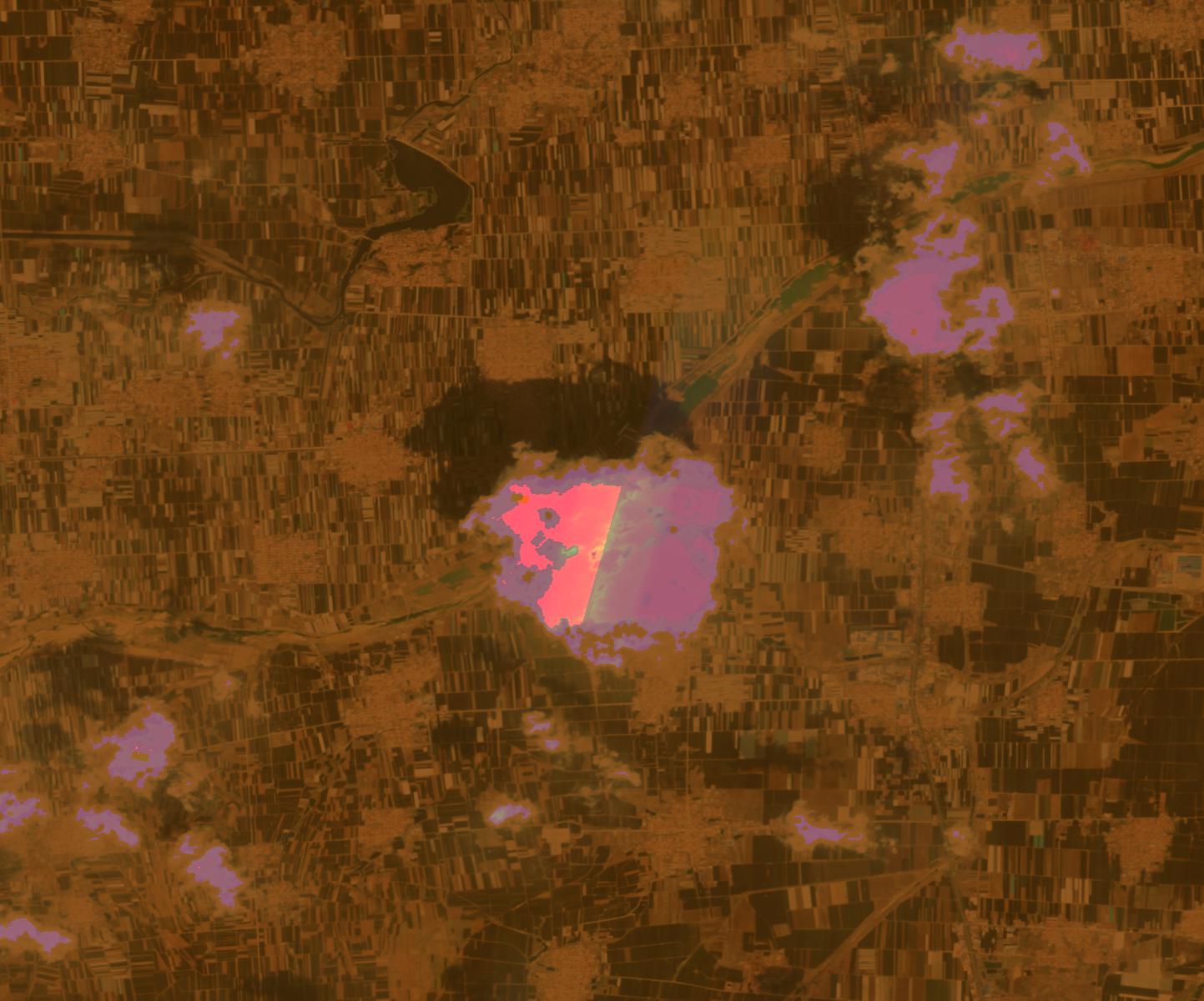}
	\caption{HM-RRN A}
\end{subfigure}
\begin{subfigure}[t]{.15\textwidth}
	\centering
\includegraphics[width=\textwidth]{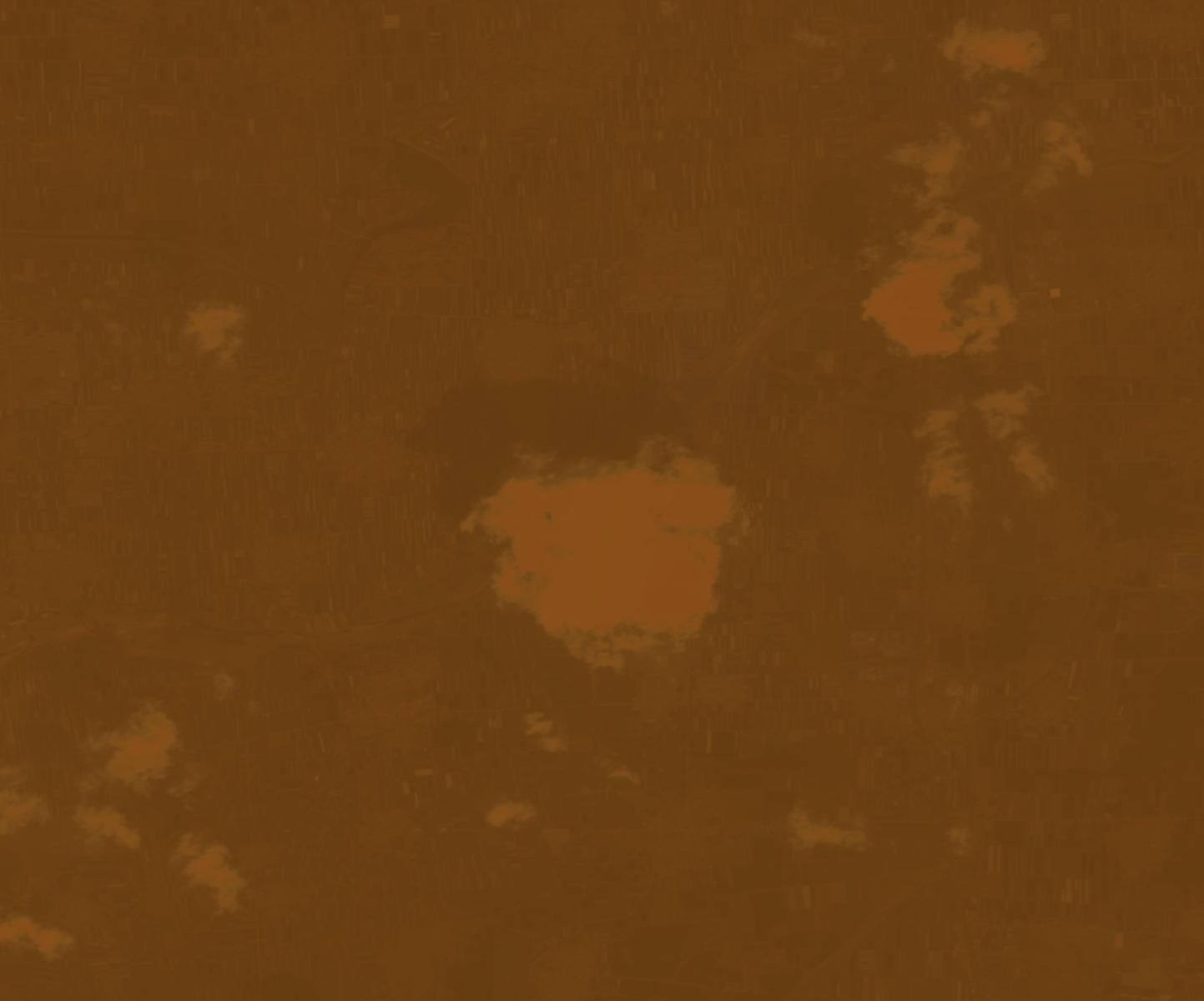}
	\caption{L-RRN-MoG A}
\end{subfigure}
\begin{subfigure}[t]{.15\textwidth}
	\centering
\includegraphics[width=\textwidth]{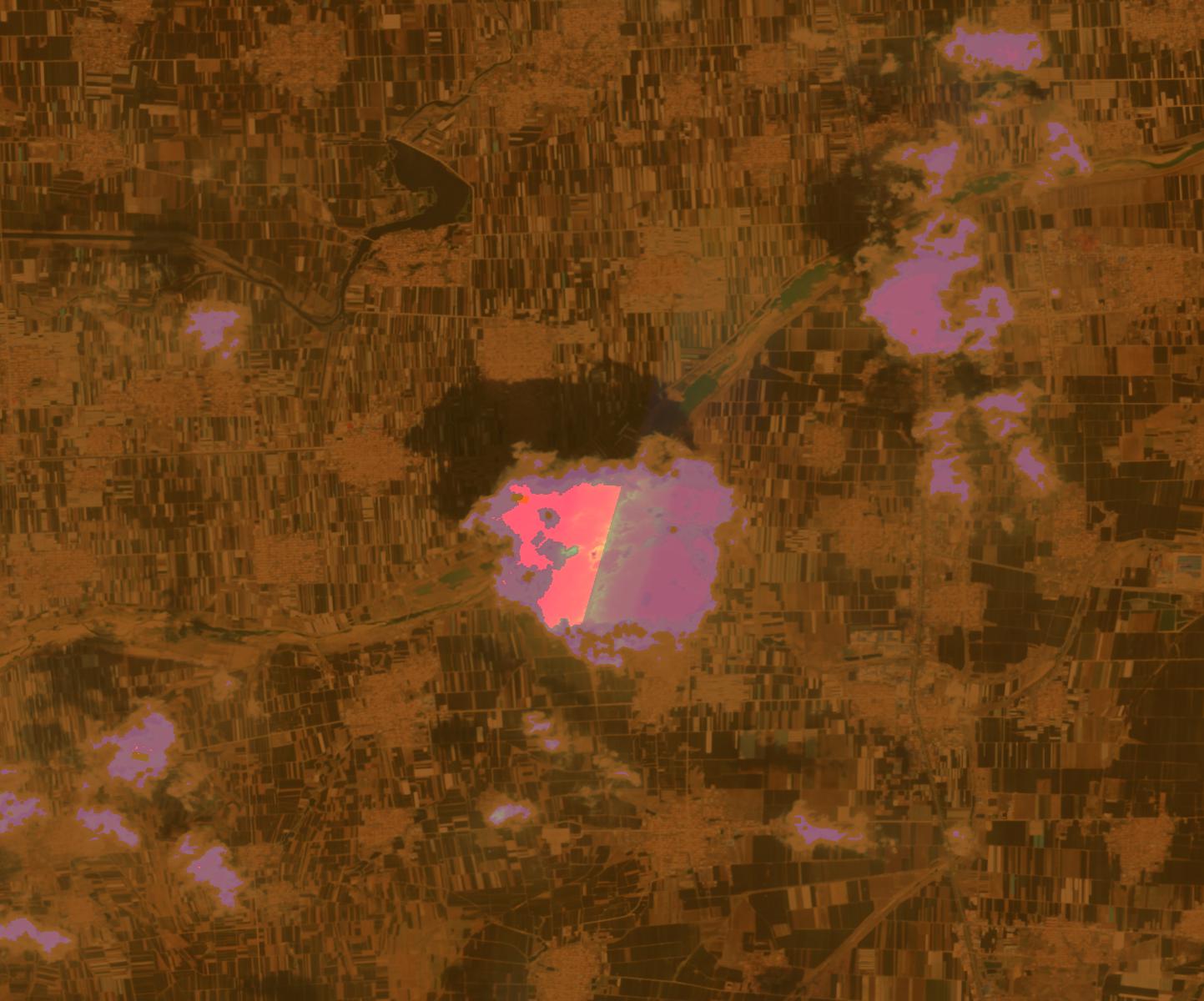}
	\caption{HM-RRN-MoL A}
\end{subfigure}
\begin{subfigure}[t]{.15\textwidth}
	\centering
\includegraphics[width=\textwidth]{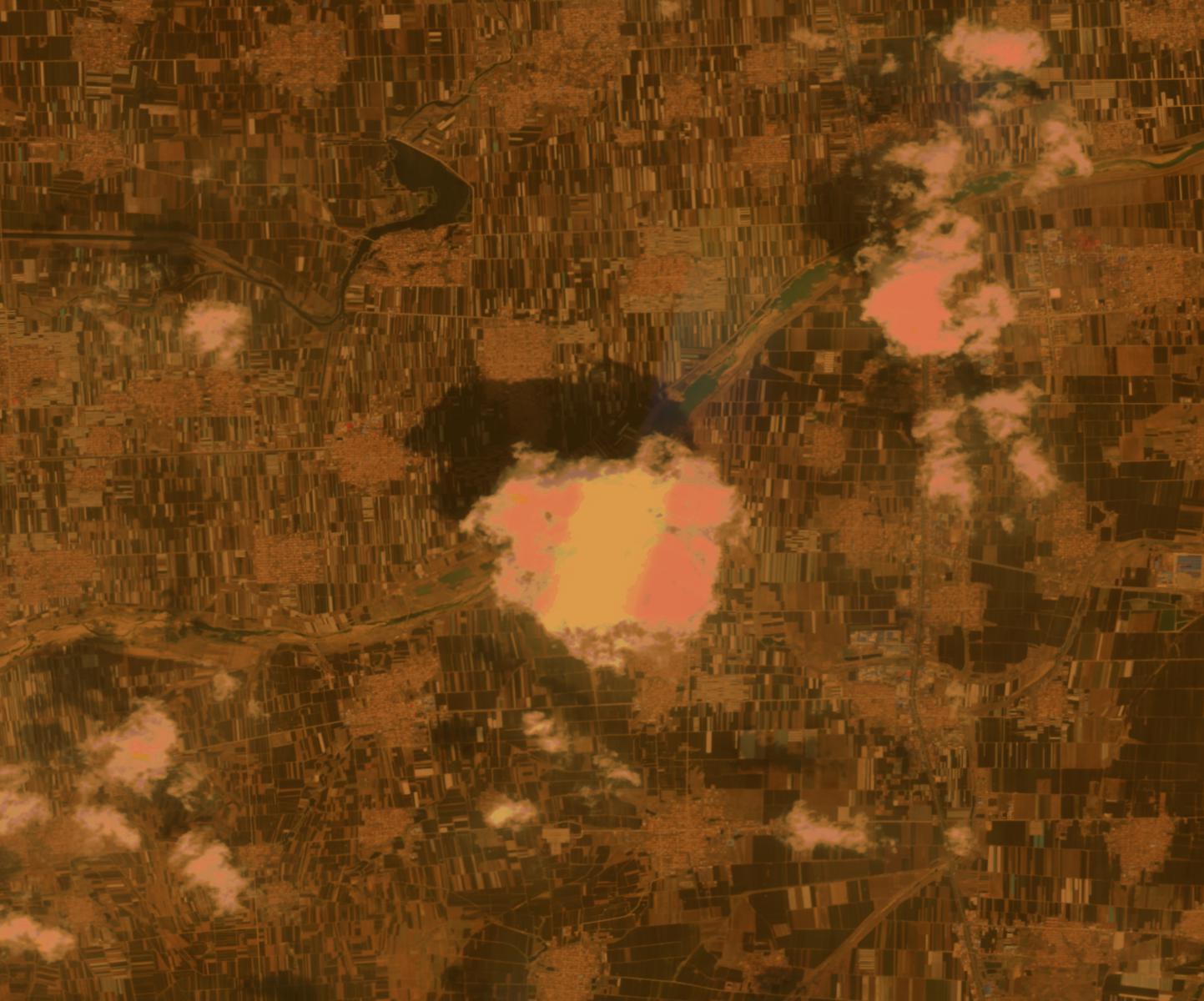}
	\caption{\textbf{HM-RRN-MoG A}}
\end{subfigure}
\caption{Comparison figure 5 different methods part II. Several clouds are in the source image, and the radiometric conditions are different between the source and target images. The target image is yellowish, while the source image is greenish. After implementing the RRN methods, the normalized source image becomes yellowish. The black pixels of subfigure(b-f) represent the no-change set assumed/derived by the different methods.   The image normalized by HM-RRN-MoG visually possesses the best normalization results and most accurate no-change set.}
\label{fig:Comparison figure 5 different methods part II}
\end{figure}

The qualitative results are shown in figure~\ref{fig:Comparison figures of 5 different methods part I} and ~\ref{fig:Comparison figure 5 different methods part II}. The quantitative results are shown in table~\ref{No-change set relative radiometric normalization methods comparison II.} and ~\ref{No-change set relative radiometric normalization methods comparison I.}.

\begin{table}[]
\caption{No-change set relative radiometric normalization methods comparison I.}
\label{No-change set relative radiometric normalization methods comparison I.}
\small
\begin{tabular}{llllll}
\hline
Method          & MLL & NCR & Time& $\Sigma$                                                                                                                   \\ \hline
Linear-RRN & -39.91              & 1                   & 0.021     & {[}149986.33,272626.67,512409,627340.12{]}                                                                                                                            \\ \hline
HM-RRN     & -34.76              & 1                   & 0.032                                                                                        & {[}600.29, 923.13,1031.75, 2499.41{]}                                       \\ \hline
\multirow{2}*{HM-RRN-MoL}      & \multirow{2}*{-34.73 }             & \multirow{2}*{0.94}                & \multirow{2}*{0.35}      &                                                                                             {[}540.32, 832.12, 902.49, 2389.90{]}\\
&&&&{[}1594.38,2382.67,3106.26,4248.49{]} \\ \hline
\multirow{2}*{Linear-RRN-MoG}  & \multirow{2}*{-28.53}              & \multirow{2}*{0.77}                & \multirow{2}*{0.51}      & {[}12419.05,13981.96,25485,08,158048.78{]}\\
& & & &{[}651706.31,1200102.39,2224256.03,2275694.71{]}                                                                              \\ \hline
\multirow{2}*{\textbf{HM-RRN-MoG}}      & \multirow{2}*{\textbf{-18.77}}              & \multirow{2}*{0.76}                & \multirow{2}*{0.58}      & {[}320.11,142.52,109.57,487.87{]}\\ &&&&{[}14566.24,10884.89,8660.29,6569.69{]}                                                                                              \\ \hline
\end{tabular}
\small{where MLL is mean log likelihood and NCR is no-change set ratio, time is counted in minute,
$\Sigma=\begin{bmatrix} \sigma_{1}^2,\sigma_{2}^2,\sigma_{3}^2  \end{bmatrix}$ for Linear-RRN,
$\Sigma=\begin{bmatrix} \beta_{1},\beta_{2},\beta_{3}  \end{bmatrix}$ for HM-RRN
$\Sigma=\begin{bmatrix} \sigma_{11}^2,\sigma_{21}^2,\sigma_{31}^2  \\ \sigma_{12}^2,\sigma_{22}^2,\sigma_{32}^2 \end{bmatrix}$ for linear-RRN-MoG and HM-RRN-MoG,
$\Sigma= \begin{bmatrix} \beta_{11},\beta_{21},\beta_{31}  \\ \beta_{12},\beta_{22},\beta_{32} \end{bmatrix}$ for HM-RRN-MoL.}\small{where MLL is Mean Log Likelihood and NCR is no-change set Ratio.}
\end{table}

\begin{table}[]
\caption{No-change set relative radiometric normalization methods comparison II.}
\label{No-change set relative radiometric normalization methods comparison II.}
\small
\begin{tabular}{llllll}
\hline
Method          & MLL & NCR & Time& $\Sigma$                                                                                                                   \\ \hline
Linear-RRN & -64.36             & 1                   & 0.014     & {[}19350.06,61976.83,160015.21,321052.40{]}                                                                                                                            \\ \hline
HM-RRN     & -34.31              & 1                   & 0.034                                                                                        & {[}198.02, 901.53,1571.68, 3233.84{]}                                       \\ \hline
\multirow{2}*{HM-RRN-MoL}      & \multirow{2}*{-34.31 }             & \multirow{2}*{0.99}                & \multirow{2}*{0.31}      &                                                                                             {[}197.67, 900.97, 1570.85, 3234.26{]}\\
&&&&{[}228.90,950.92,1644.67,3196.34{]} \\ \hline
\multirow{2}*{Linear-RRN-MoG}  & \multirow{2}*{-28.12}              & \multirow{2}*{0.21}                & \multirow{2}*{0.46}      & {[}691.84,2282.03,10133.96,204785.08{]}\\
& & & &{[}24572.08,78599.59,202866.12,354469.48{]}                                                                              \\ \hline
\multirow{2}*{\textbf{HM-RRN-MoG}}      & \multirow{2}*{\textbf{-18.79}}              & \multirow{2}*{0.76}                & \multirow{2}*{0.57}      & {[}105.90,150.41,185.66,528.78{]}\\ &&&&{[}3105.84,2857.70,3799.21,5408.17{]}                                                                                              \\ \hline
\end{tabular}
\small{where MLL is mean log likelihood and NCR is no-change set ratio, time is counted in minute,
$\Sigma=\begin{bmatrix} \sigma_{1}^2,\sigma_{2}^2,\sigma_{3}^2  \end{bmatrix}$ for Linear-RRN,
$\Sigma=\begin{bmatrix} \beta_{1},\beta_{2},\beta_{3}  \end{bmatrix}$ for HM-RRN
$\Sigma=\begin{bmatrix} \sigma_{11}^2,\sigma_{21}^2,\sigma_{31}^2  \\ \sigma_{12}^2,\sigma_{22}^2,\sigma_{32}^2 \end{bmatrix}$ for linear-RRN-MoG and HM-RRN-MoG,
$\Sigma= \begin{bmatrix} \beta_{11},\beta_{21},\beta_{31}  \\ \beta_{12},\beta_{22},\beta_{32} \end{bmatrix}$ for HM-RRN-MoL.}\small{where MLL is Mean Log Likelihood and NCR is no-change set Ratio.}
\end{table}

We can find that the traditional linear least square and histogram matching relative radiometric normalization method assume all the points are the no change points. Affected by the cloud, these two methods either blur the image or make the image overexposed. The linear least square RRN method blurs the source image in subfigure~h in figure~\ref{fig:Comparison figures of 5 different methods part I} and ~\ref{fig:Comparison figure 5 different methods part II}. In subfigure~i in figure~\ref{fig:Comparison figures of 5 different methods part I} we can find that some farmlands and some buildings are wrong overexposed, while in subfigure~i in figure~\ref{fig:Comparison figure 5 different methods part II} the cloud are overexposed, and the farmlands and buildings are darkened. Regarding the mean log-likelihood, HM-RRN with $-34.76/-34.31$ is slightly better than Linear-RRN with $-39.91/-64.36$

Due to the heave tail characteristic of the Laplacian distribution, which assigns a higher probability than the Gaussian distribution's probability of large loss sample, the histogram matching RRN with a mixture of Laplacian noise modeling method(HM-RRN-MoL) inclines to underestimate the no-change set. In the subfigure~k in figure~\ref{fig:Comparison figures of 5 different methods part I} the clouds were detected, but the changes of farmlands and cloud shadows are not detected. As a result, some farmlands and buildings are still overexposed. In the subfigure~k in figure~\ref{fig:Comparison figures of 5 different methods part I}, this underestimation of the no-change set situation occurs again. The clouds are still overexposed. The normalized source image is almost like the normalized source image in subfigure~i.

We can find that the linear RRN with a mixture of Gaussian noise modeling (linear-RRN-MoG) method alleviates the blurring in subfigure~j in figure~\ref{fig:Comparison figures of 5 different methods part I} but some farmlands and buildings are not as bright as that in the target image. In subfigure~j in figure~\ref{fig:Comparison figure 5 different methods part II}, due to the lack of fitting capacity of the linear model, linear-RRN-MoG overestimates the no-change set. As a result, the blurring is not alleviated. Regarding the mean log-likelihood, HM-RRN with $-28.53/-28.12$ is slightly better than Linear-RRN with $-34.73/-34.31$.

Compared to 4 methods above, the histogram matching method with a mixture of Gaussian noise modeling method(HM-RRN-MoG) properly estimates the no-change set. The normalized source images in subfigure~l in figure~\ref{fig:Comparison figures of 5 different methods part I} and ~\ref{fig:Comparison figure 5 different methods part II} have the best visual appearances, which possess no overexposed buildings, farmlands, and clouds and are most similar to the original one. HM-RRN-MoG has the most significant mean log-likelihood $-18.77/-18.79$.

\FloatBarrier
\subsubsection{Foggy cases}

\begin{figure}[ht]
\centering
\captionsetup[subfigure]{font=scriptsize,labelfont=scriptsize}
\begin{subfigure}[t]{.15\textwidth}
	\centering
	\includegraphics[width=\textwidth]{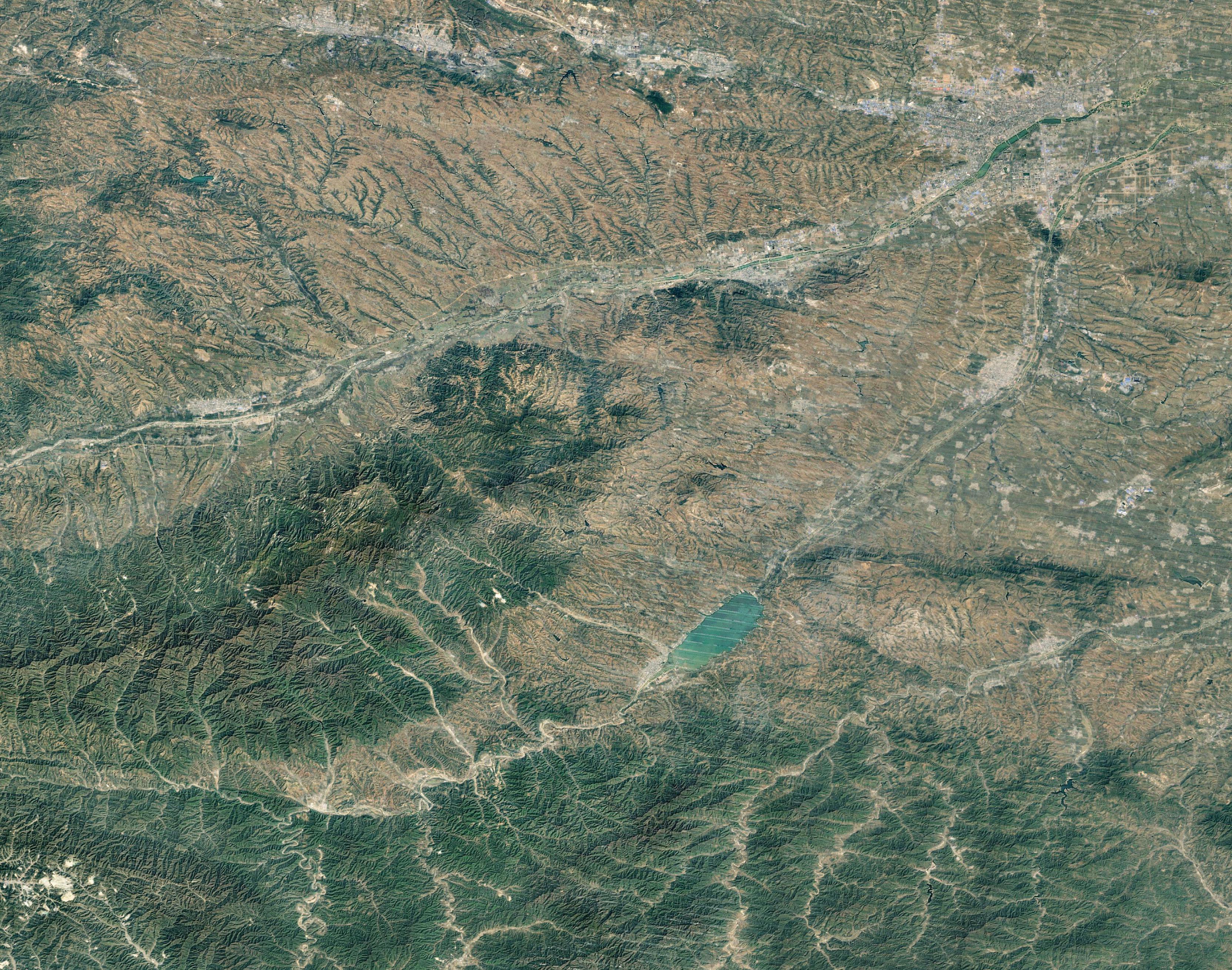}
	\caption{Target image C}
\end{subfigure}
\begin{subfigure}[t]{.15\textwidth}
	\centering \includegraphics[width=\textwidth]{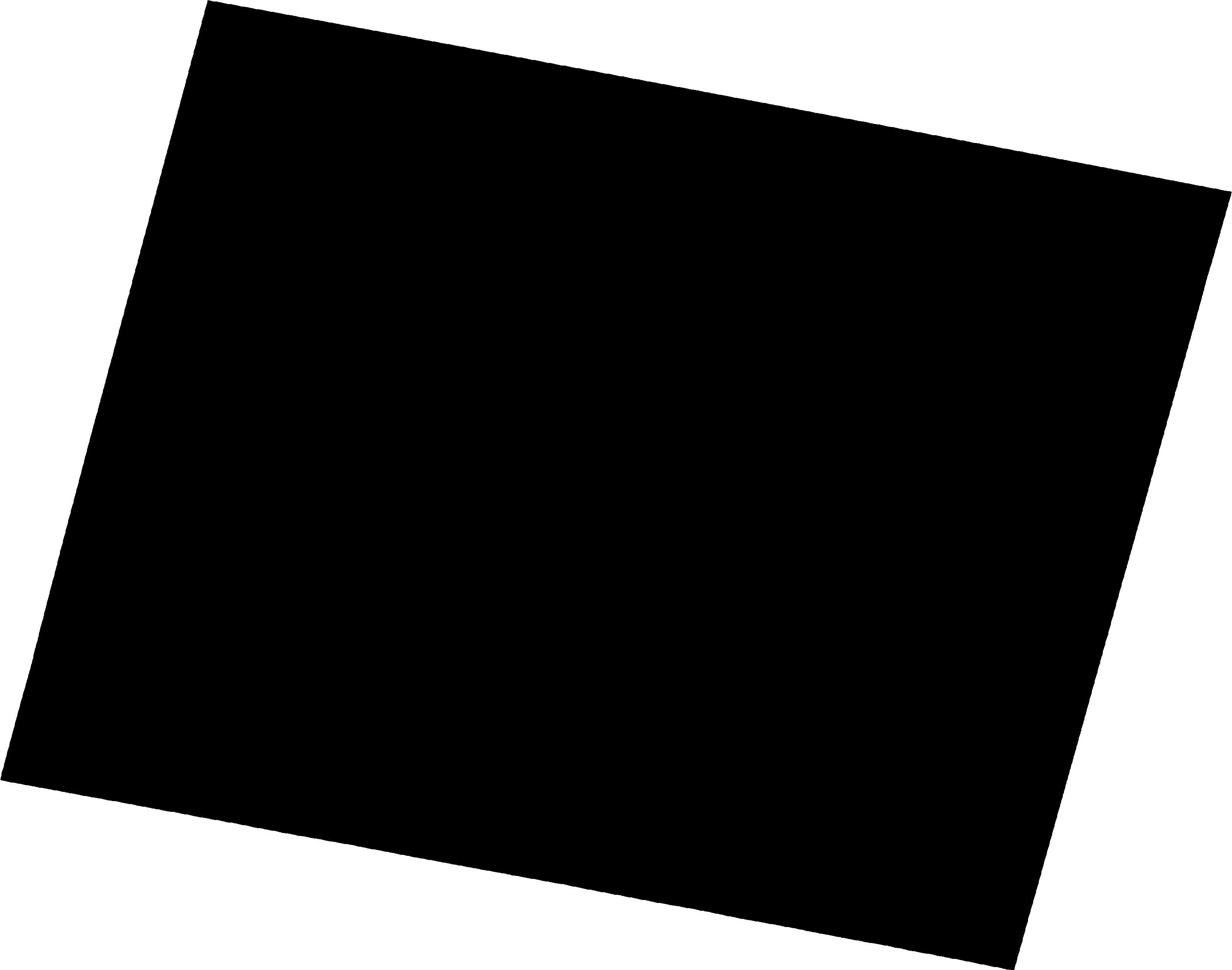}
	\caption{L-RRN NC}
\end{subfigure}
\begin{subfigure}[t]{.15\textwidth}
	\centering
\includegraphics[width=\textwidth]{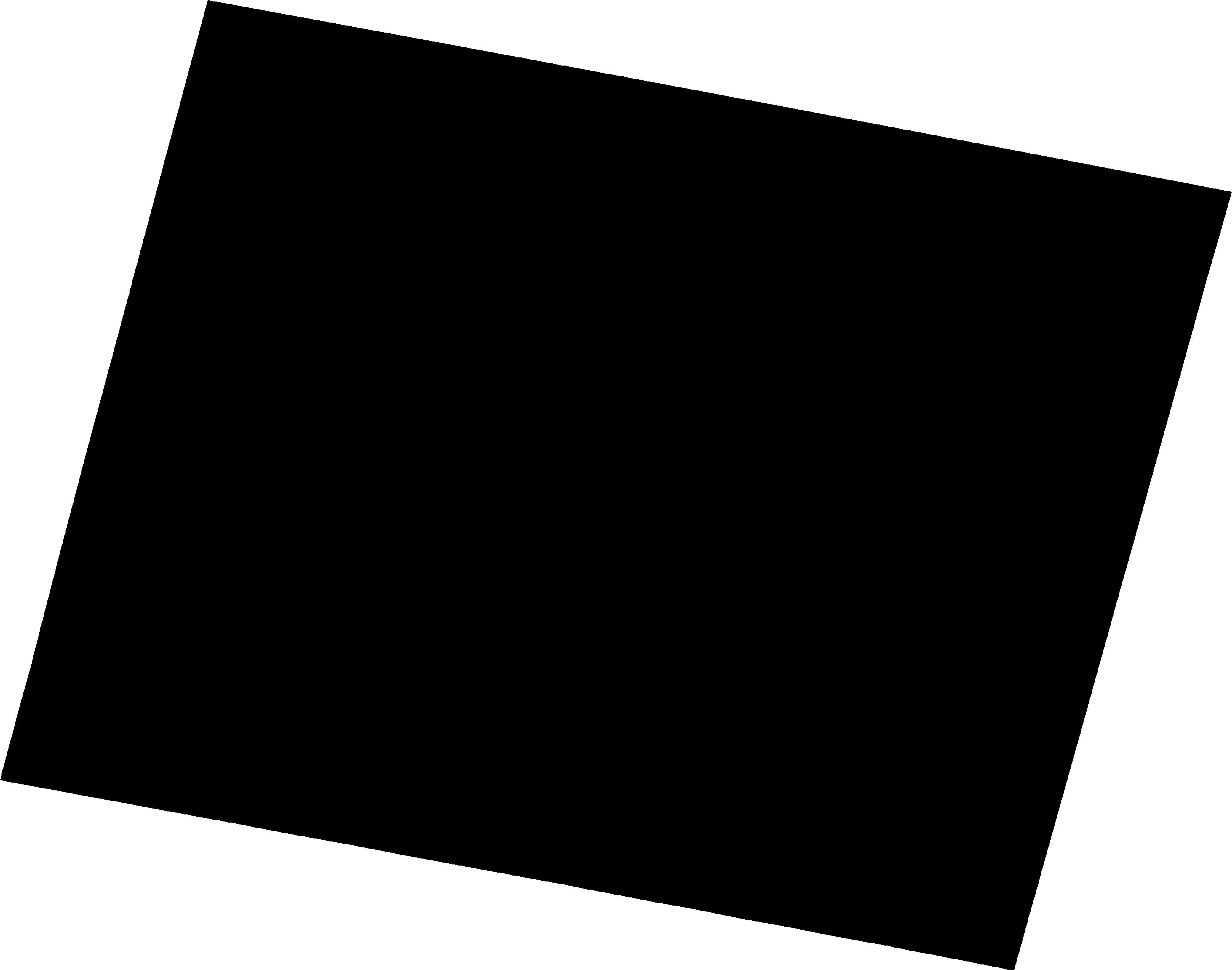}
	\caption{HM-RRN NC}
\end{subfigure}
\begin{subfigure}[t]{.15\textwidth}
	\centering
\includegraphics[width=\textwidth]{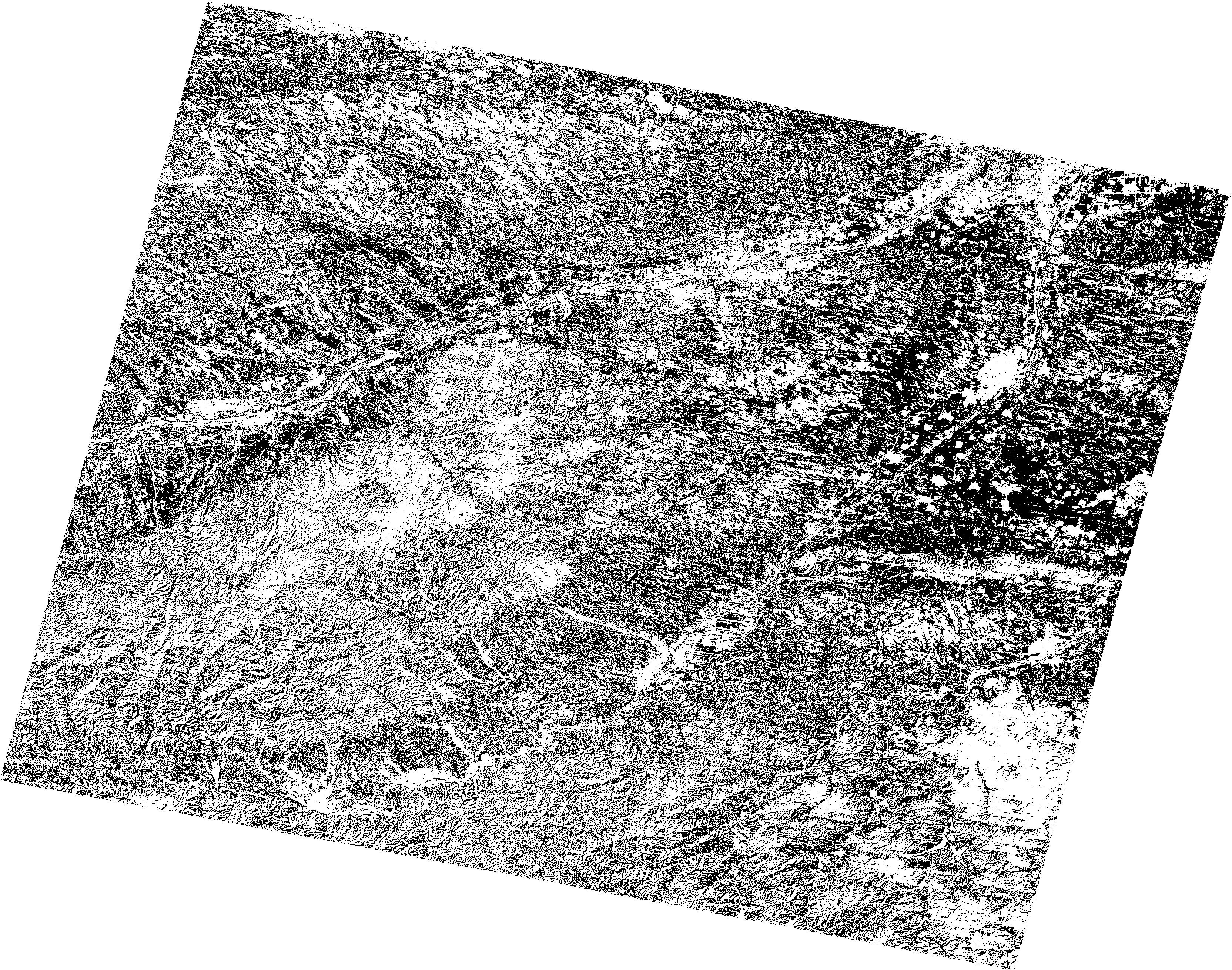}
	\caption{L-RRN-MoG NC}
\end{subfigure}
\begin{subfigure}[t]{.15\textwidth}
	\centering
\includegraphics[width=\textwidth]{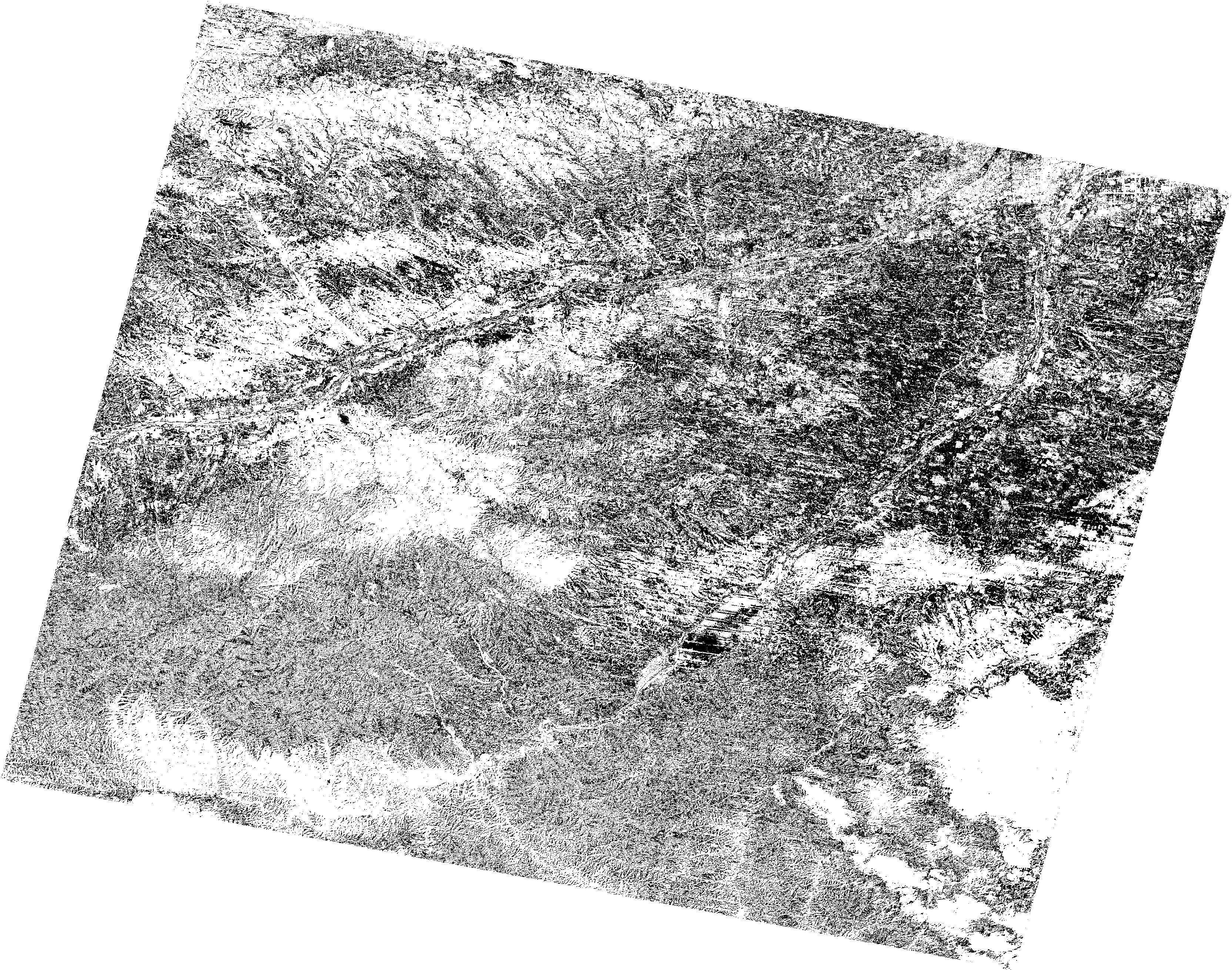}
	\caption{HM-RRN-MoL NC}
\end{subfigure}
\begin{subfigure}[t]{.15\textwidth}
	\centering
\includegraphics[width=\textwidth]{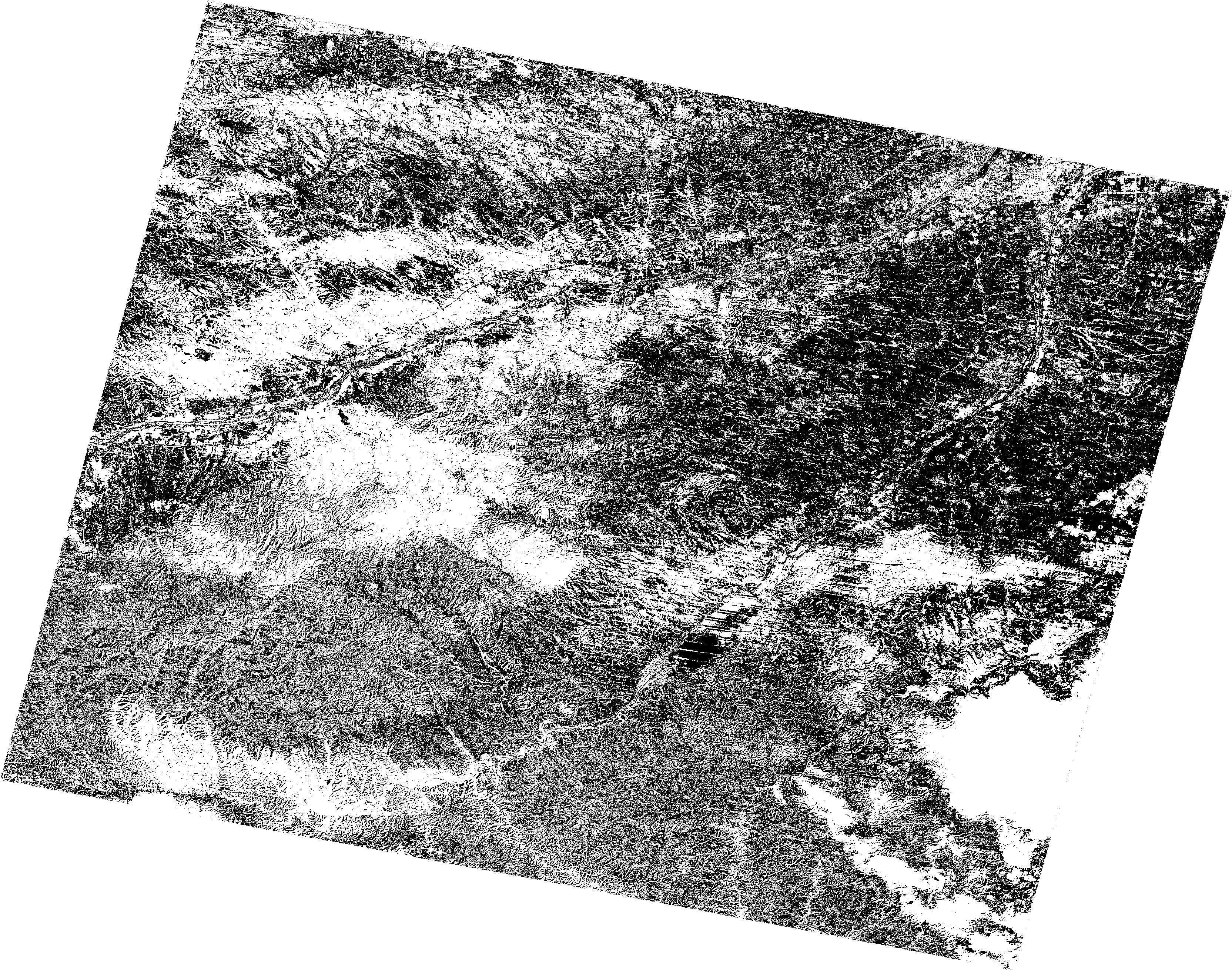}
	\caption{HM-RRN-MoG NC}
\end{subfigure}
\begin{subfigure}[t]{.15\textwidth}
	\centering
\includegraphics[width=\textwidth]{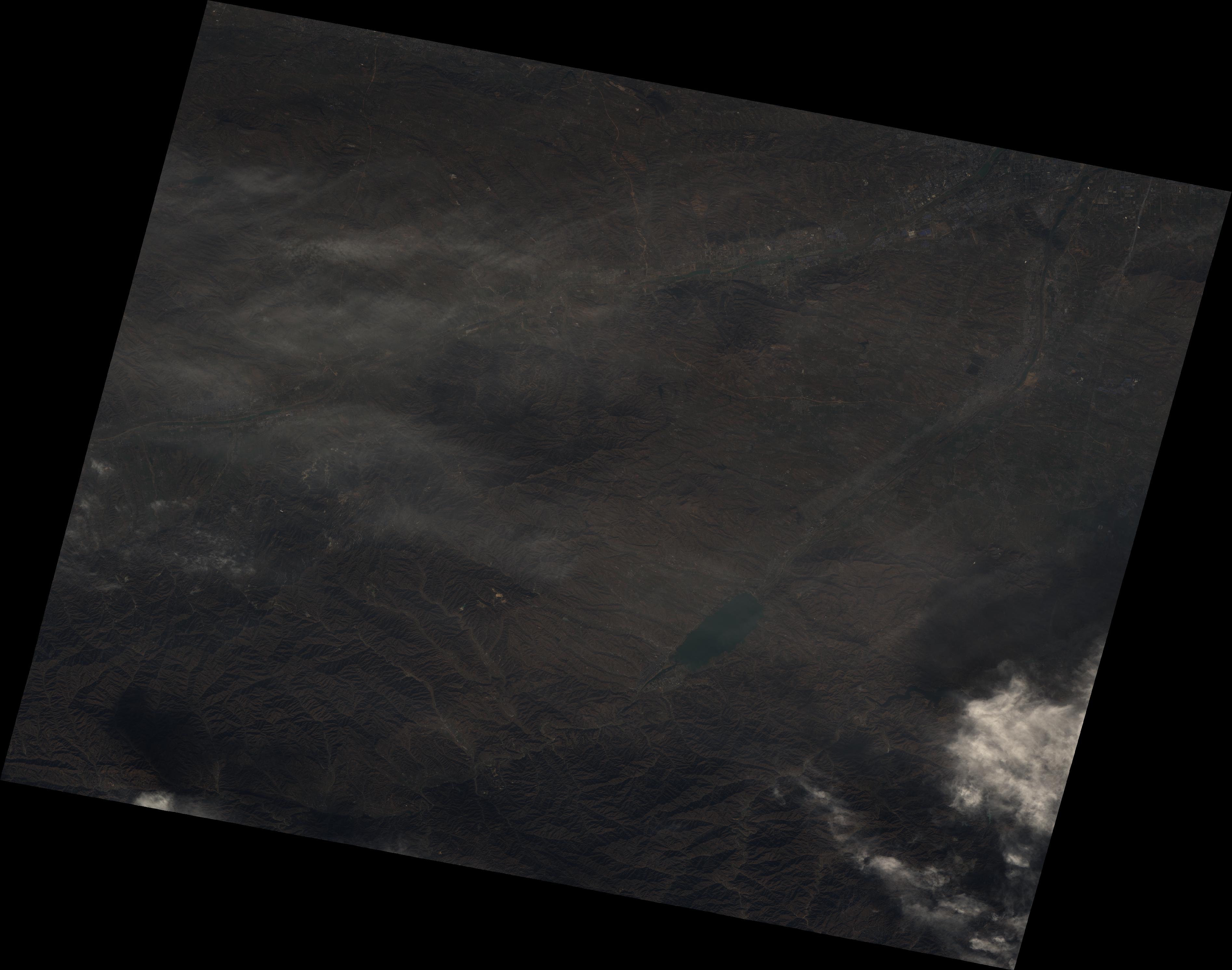}
	\caption{Source image D}
\end{subfigure}
\begin{subfigure}[t]{.15\textwidth}
	\centering
\includegraphics[width=\textwidth]{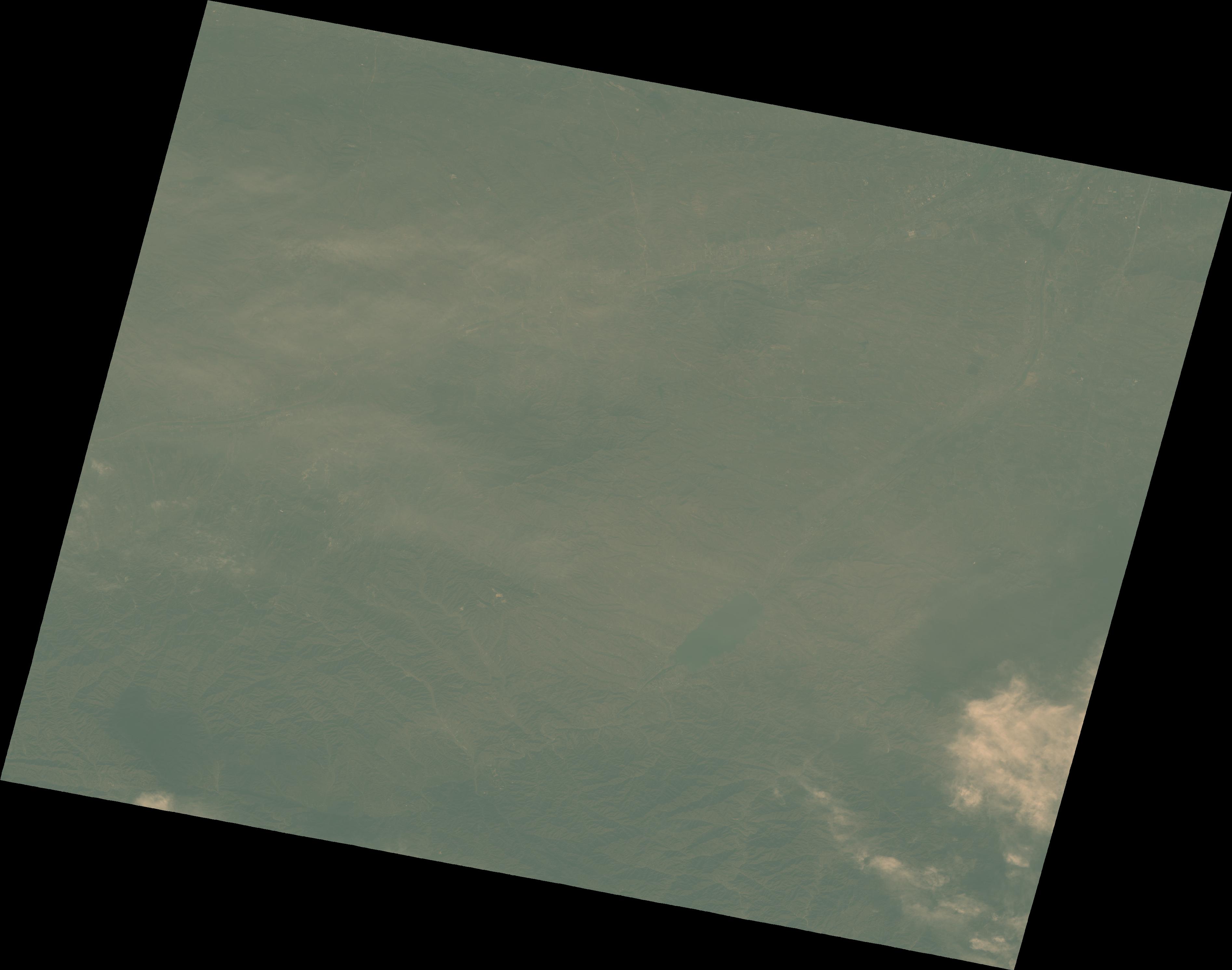}
	\caption{L-RRN D}
\end{subfigure}
\begin{subfigure}[t]{.15\textwidth}
	\centering
\includegraphics[width=\textwidth]{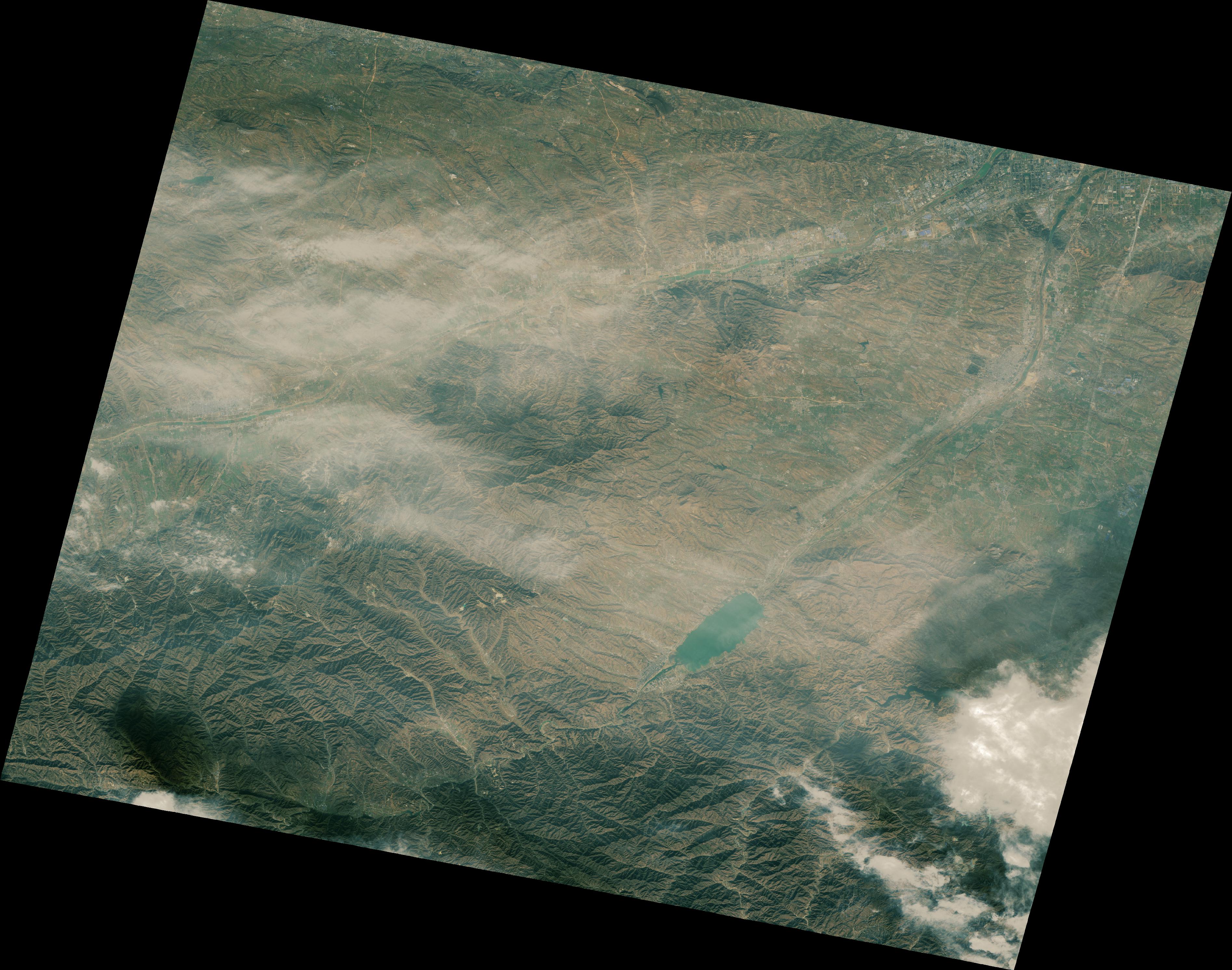}
	\caption{HM-RRN D}
\end{subfigure}
\begin{subfigure}[t]{.15\textwidth}
	\centering
\includegraphics[width=\textwidth]{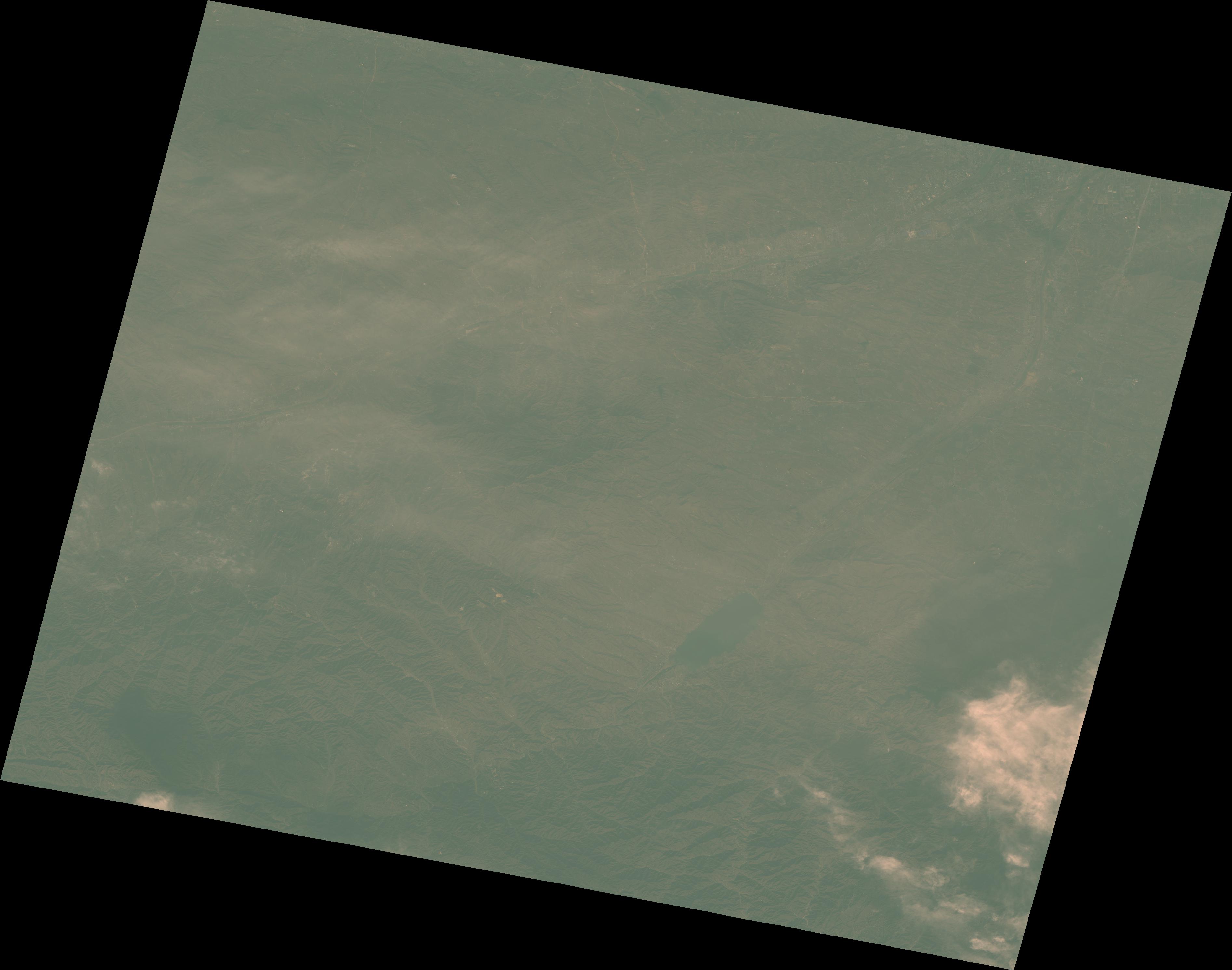}
	\caption{L-RRN-MoG D}
\end{subfigure}
\begin{subfigure}[t]{.15\textwidth}
	\centering
\includegraphics[width=\textwidth]{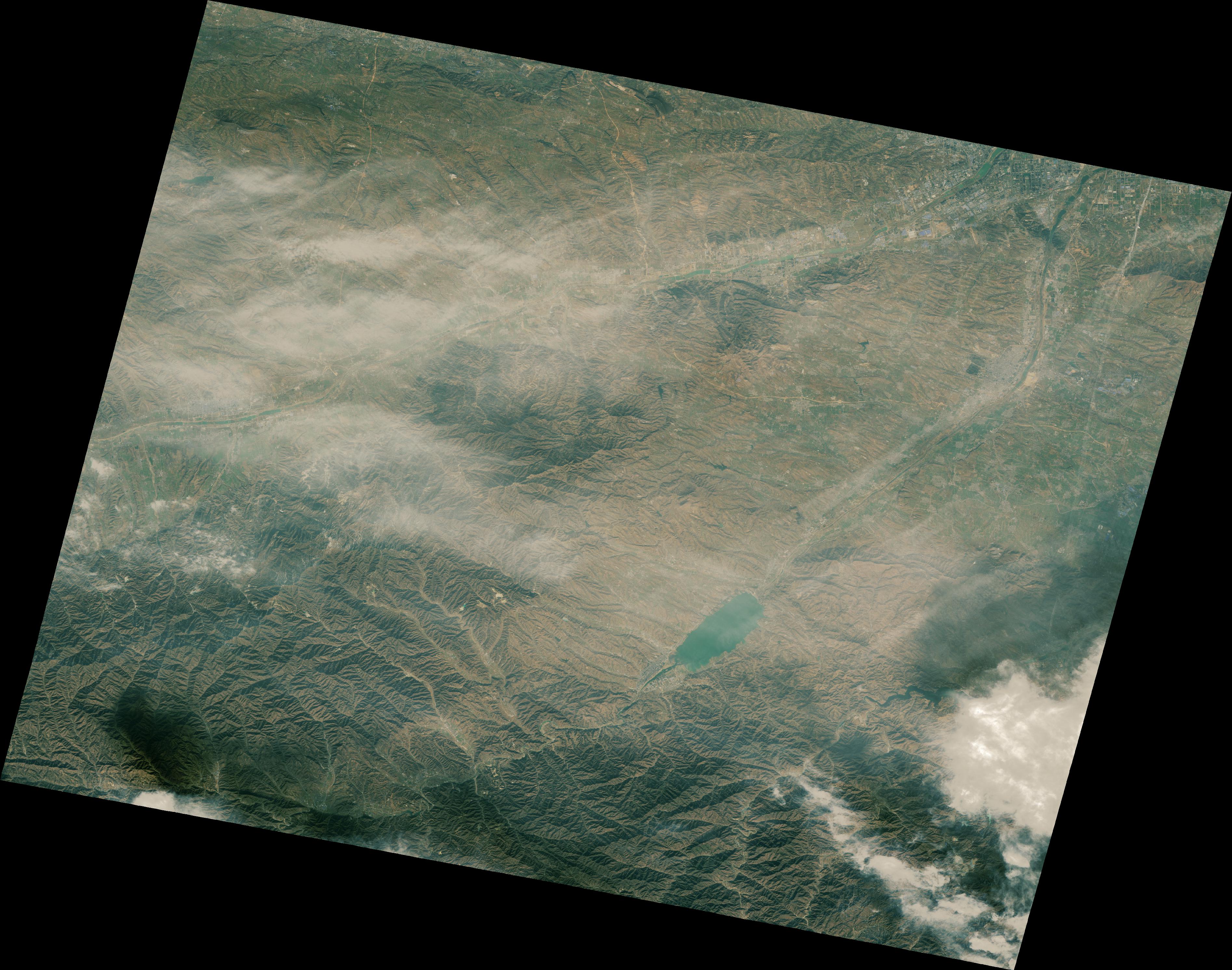}
	\caption{HM-RRN-MoL D}
\end{subfigure}
\begin{subfigure}[t]{.15\textwidth}
	\centering
\includegraphics[width=\textwidth]{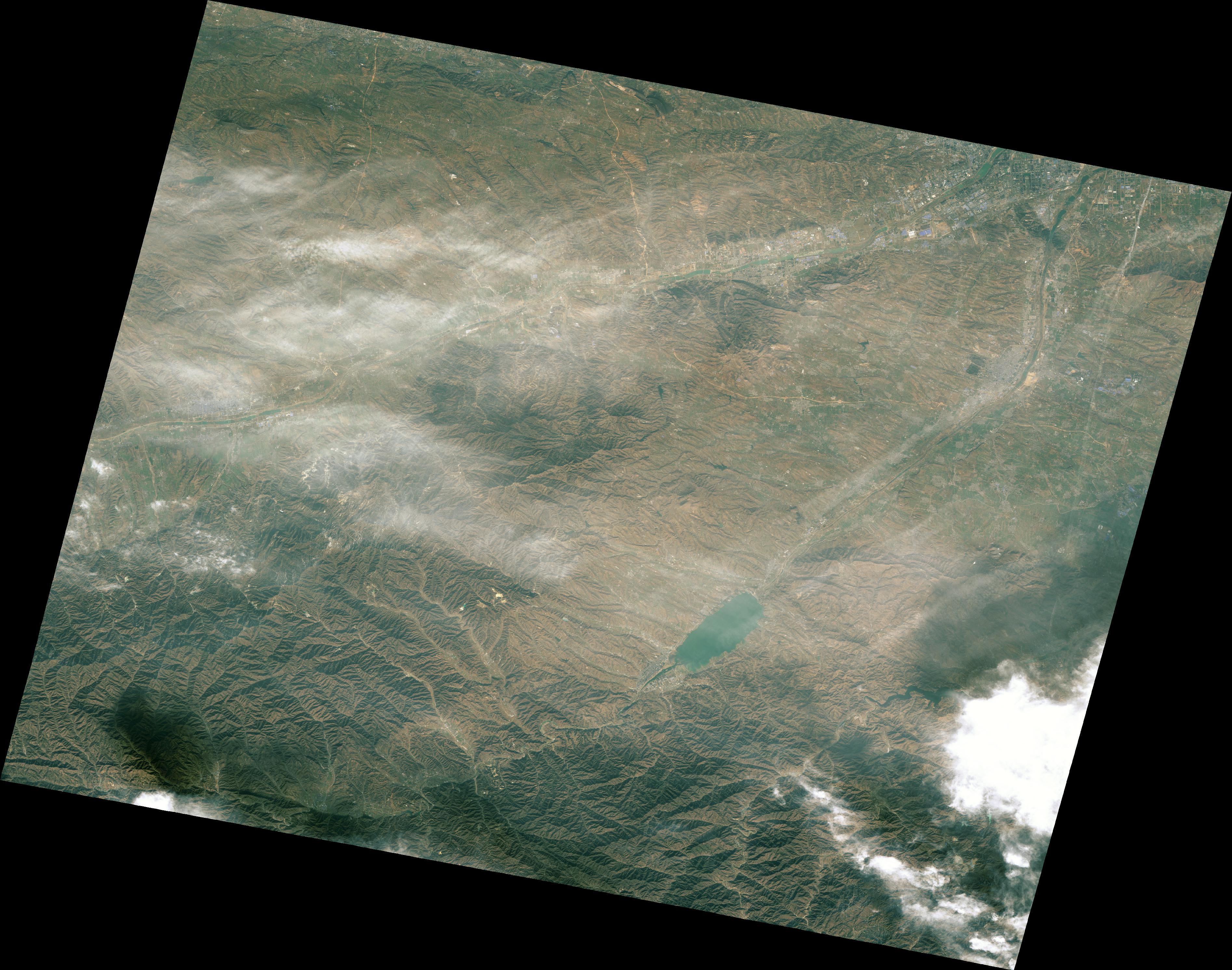}
	\caption{\textbf{HM-RRN-MoG D}}
\end{subfigure}
\caption{Comparison figure 5 different methods part III. Several fogs are in the source image, and the radiometric conditions are different between the source and target images. The target image is greenish, while the source image is darkish. After implementing the RRN methods, the normalized source image becomes yellowish. The black pixels of subfigure(b-f) represent the no-change set assumed/derived by the different methods.   The image normalized by HM-RRN-MoG visually possesses the best normalization results and most accurate no-change set.}
\label{fig:Comparison figure 5 different methods part III}
\end{figure}

The qualitative results are shown in figure~\ref{fig:Comparison figure 5 different methods part III} and the quantitative results are shown in table~\ref{No-change set relative radiometric normalization methods comparison III.}.

From figure~\ref{fig:Comparison figure 5 different methods part III}, we can find that HM-RRN-MoG method achieves the best performance: the clouds and fogs are white only in the subfigure (l), and the no-change set in the subfigure (f)  is visually the most accurate. The linear methods all make the overall color deviation of the RRN images and make the images overcast. The histogram methods alleviate this situation. However, the clouds and fogs are in dust color in the RRN images. Quantitatively, the linear-RRN-MoG method possesses the best log-likelihood according to table~\ref{No-change set relative radiometric normalization methods comparison III.} while HM-RRN-MoG method possesses the second high log-likelihood. Considering the HM-RRN-MoG has the best qualitative performance, we regard the HM-RRN-MoG as the best model.
\begin{table}[]
\caption{No-change set relative radiometric normalization methods comparison III.}
\label{No-change set relative radiometric normalization methods comparison III.}
\small
\begin{tabular}{llllll}
\hline
Method          & MLL & NCR & Time& $\Sigma$                                                                                                                   \\ \hline
Linear-RRN & -30.46             & 1.00                   & 0.033     & {[}1830.81, 1044.06,  879.82{]}                                                                                                                            \\ \hline
HM-RRN     & -15.49              & 1.00                   & 0.040                                                                                        & {[}39.10 30.13 28.17{]}                                       \\ \hline
\multirow{2}*{HM-RRN-MoL}      & \multirow{2}*{-15.35 }             & \multirow{2}*{0.75}                & \multirow{2}*{0.33}      &                                                                                             {[}14.81,  9.75,  9.73{]}\\
&&&&{[}50.90, 40.00, 37.48{]} \\ \hline
\multirow{2}*{\textbf{Linear-RRN-MoG}}  & \multirow{2}*{\textbf{-14.53}}              & \multirow{2}*{0.41}                & \multirow{2}*{0.53}      & {[}348.10, 121.28, 107.73{]}\\
& & & &{[}2880.99 1696.78 1423.61{]}                                                                              \\ \hline
\multirow{2}*{HM-RRN-MoG}      & \multirow{2}*{-15.26}              & \multirow{2}*{0.46}                & \multirow{2}*{0.75}      & {[}438.58, 206.30, 211.16{]}\\ &&&&{[}4897.28, 3299.45, 3325.82{]}                                                                                              \\ \hline
\end{tabular}
\small{where MLL is mean log likelihood and NCR is no-change set ratio, time is counted in minute,
$\Sigma=\begin{bmatrix} \sigma_{1}^2,\sigma_{2}^2,\sigma_{3}^2  \end{bmatrix}$ for Linear-RRN,
$\Sigma=\begin{bmatrix} \beta_{1},\beta_{2},\beta_{3}  \end{bmatrix}$ for HM-RRN
$\Sigma=\begin{bmatrix} \sigma_{11}^2,\sigma_{21}^2,\sigma_{31}^2  \\ \sigma_{12}^2,\sigma_{22}^2,\sigma_{32}^2 \end{bmatrix}$ for linear-RRN-MoG and HM-RRN-MoG,
$\Sigma= \begin{bmatrix} \beta_{11},\beta_{21},\beta_{31}  \\ \beta_{12},\beta_{22},\beta_{32} \end{bmatrix}$ for HM-RRN-MoL.}
\end{table}
\FloatBarrier
\subsection{Vegetation and water change detection with relative radiometric normalization}

\begin{figure}[ht]
\centering
\captionsetup[subfigure]{font=scriptsize,labelfont=scriptsize}
\begin{subfigure}[t]{.15\textwidth}
	\centering
	\includegraphics[width=\textwidth,clip,trim=0 32 0 0]{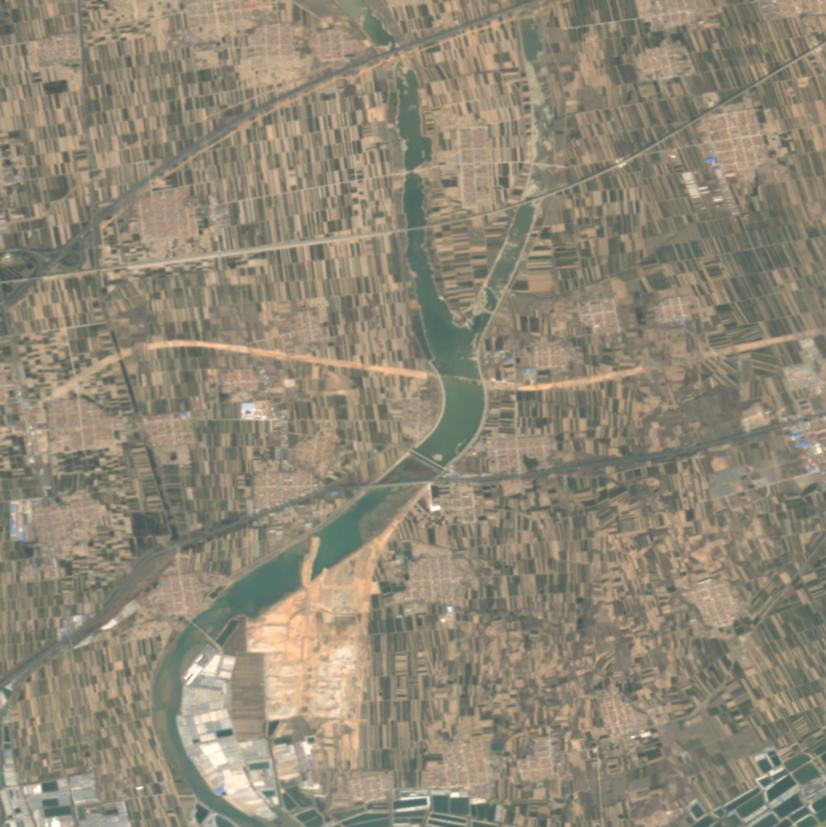}
	\caption{Target image E}
\end{subfigure}
\begin{subfigure}[t]{.15\textwidth}
	\centering
\includegraphics[width=\textwidth,clip,trim=0 32 0 0]{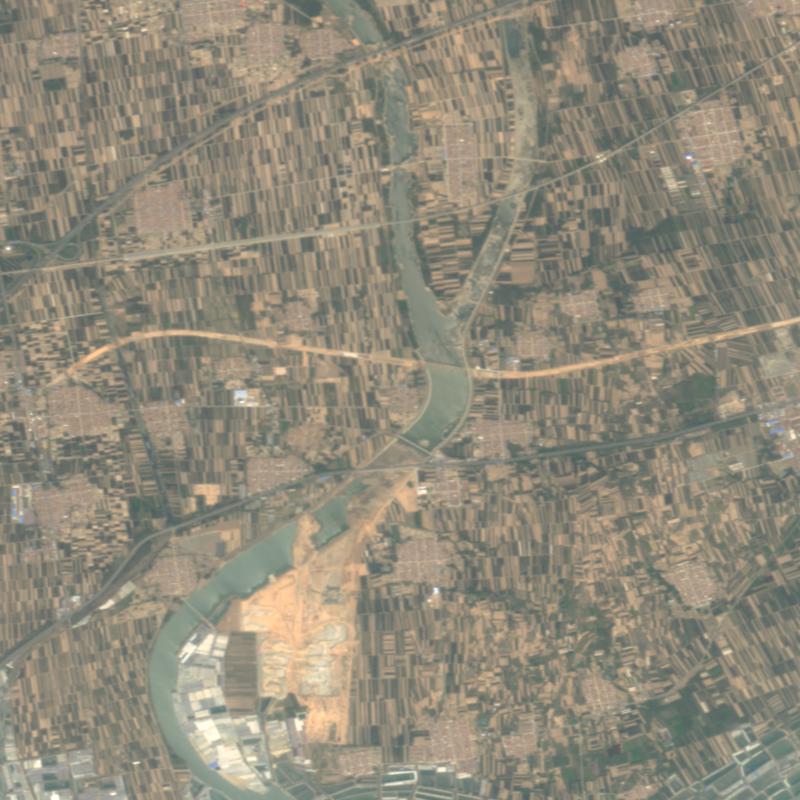}
	\caption{HM-RRN-MoG F}
\end{subfigure}
\begin{subfigure}[t]{.15\textwidth}
	\centering
\includegraphics[width=\textwidth,clip,trim=0 32 0 0]{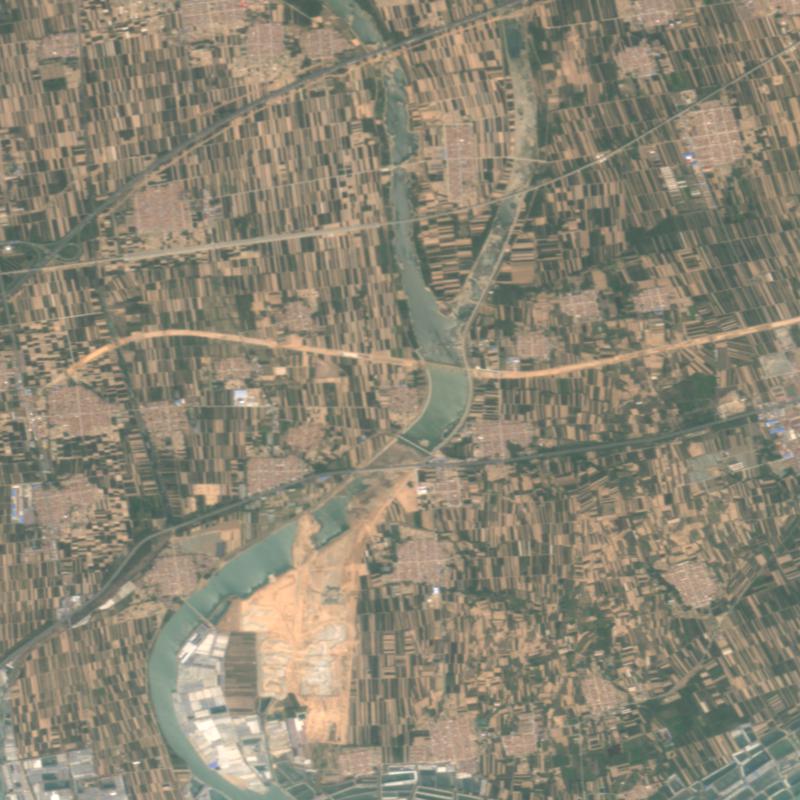}
	\caption{source image F}
\end{subfigure}
\begin{subfigure}[t]{.15\textwidth}
	\centering
	\includegraphics[width=\textwidth]{figure/GF1C_PMS_552.jpg}
	\caption{Target image B}
\end{subfigure}
\begin{subfigure}[t]{.15\textwidth}
	\centering
\includegraphics[width=\textwidth]{figure/GF1C_PMS_127_rrn_mog_hm.jpg}
	\caption{HM-RRN-MoG A}
\end{subfigure}
\begin{subfigure}[t]{.15\textwidth}
	\centering
\includegraphics[width=\textwidth]{figure/GF1C_PMS_127.jpg}
	\caption{source image A}
\end{subfigure}
\begin{subfigure}[t]{.15\textwidth}
	\centering
	\includegraphics[width=\textwidth,clip,trim=0 32 0 0]{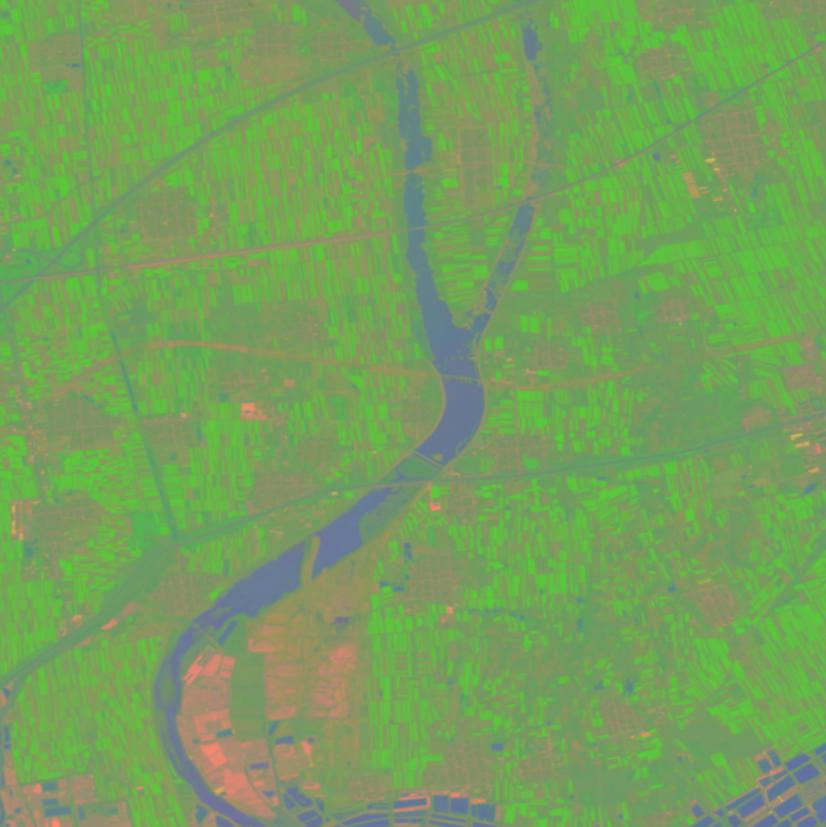}
	\caption{BVW target image E}
\end{subfigure}
\begin{subfigure}[t]{.15\textwidth}
	\centering
\includegraphics[width=\textwidth,clip,trim=0 32 0 0]{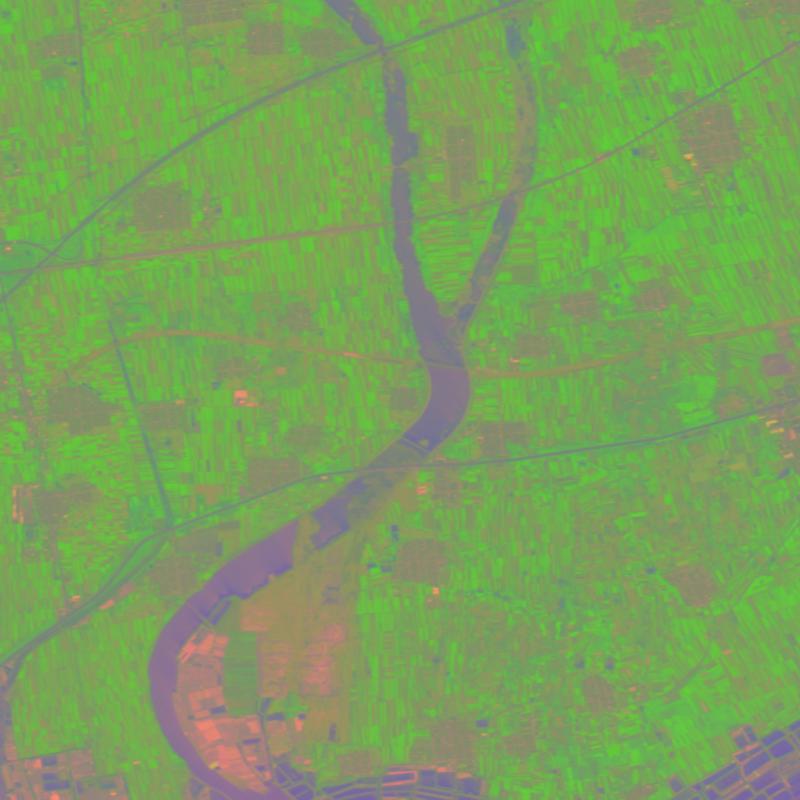}
	\caption{BVW HM-RRN-MoG F}
\end{subfigure}
\begin{subfigure}[t]{.15\textwidth}
	\centering
\includegraphics[width=\textwidth,clip,trim=0 32 0 0]{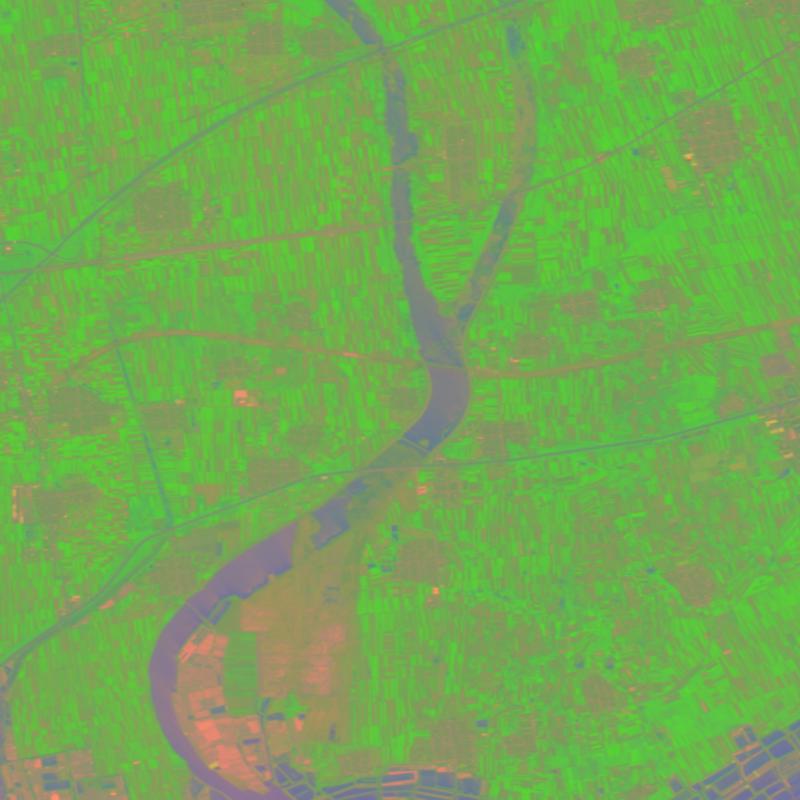}
	\caption{BVW source image F}
\end{subfigure}
\begin{subfigure}[t]{.15\textwidth}
	\centering
	\includegraphics[width=\textwidth]{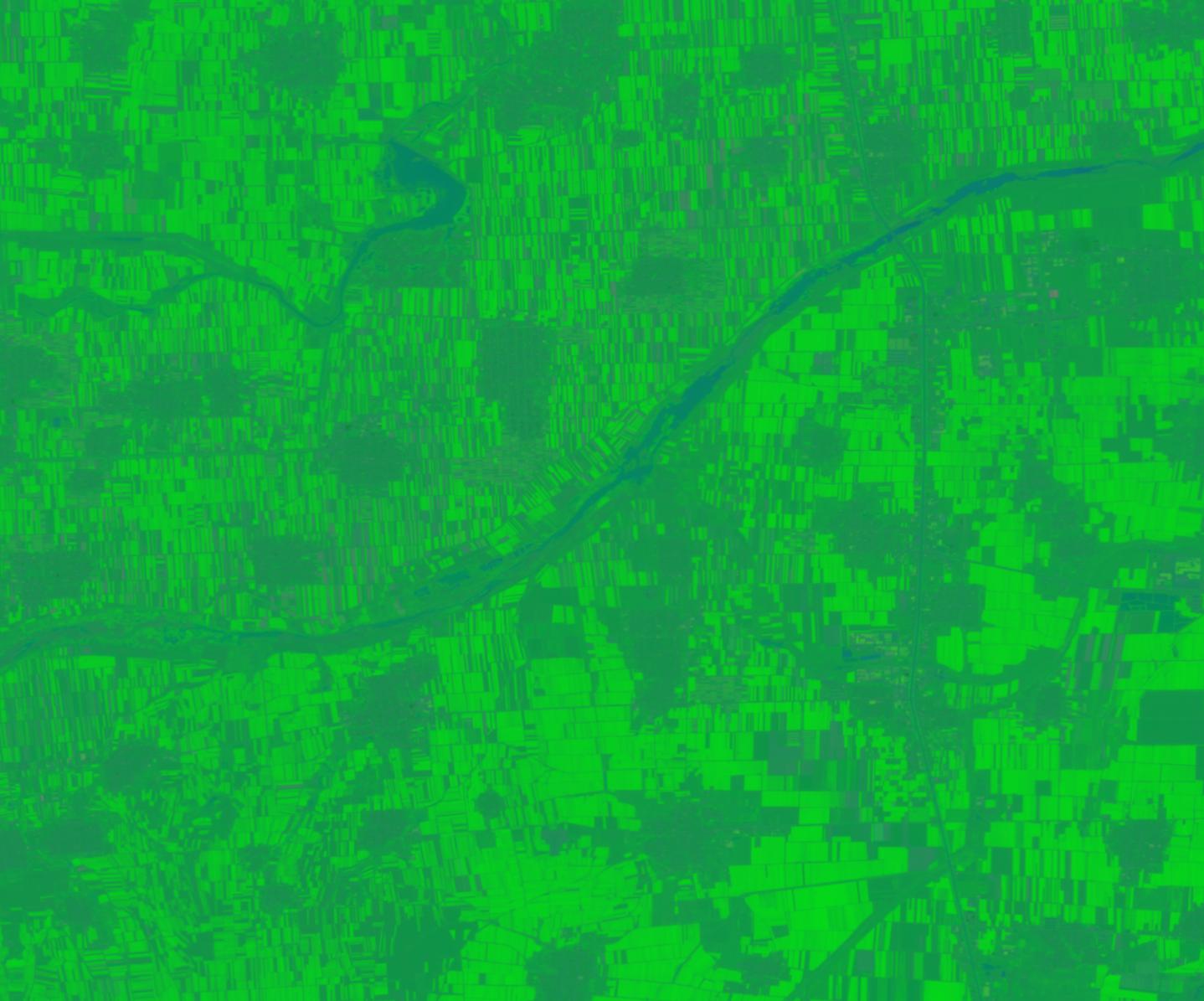}
	\caption{BVW target image B}
\end{subfigure}
\begin{subfigure}[t]{.15\textwidth}
	\centering
\includegraphics[width=\textwidth]{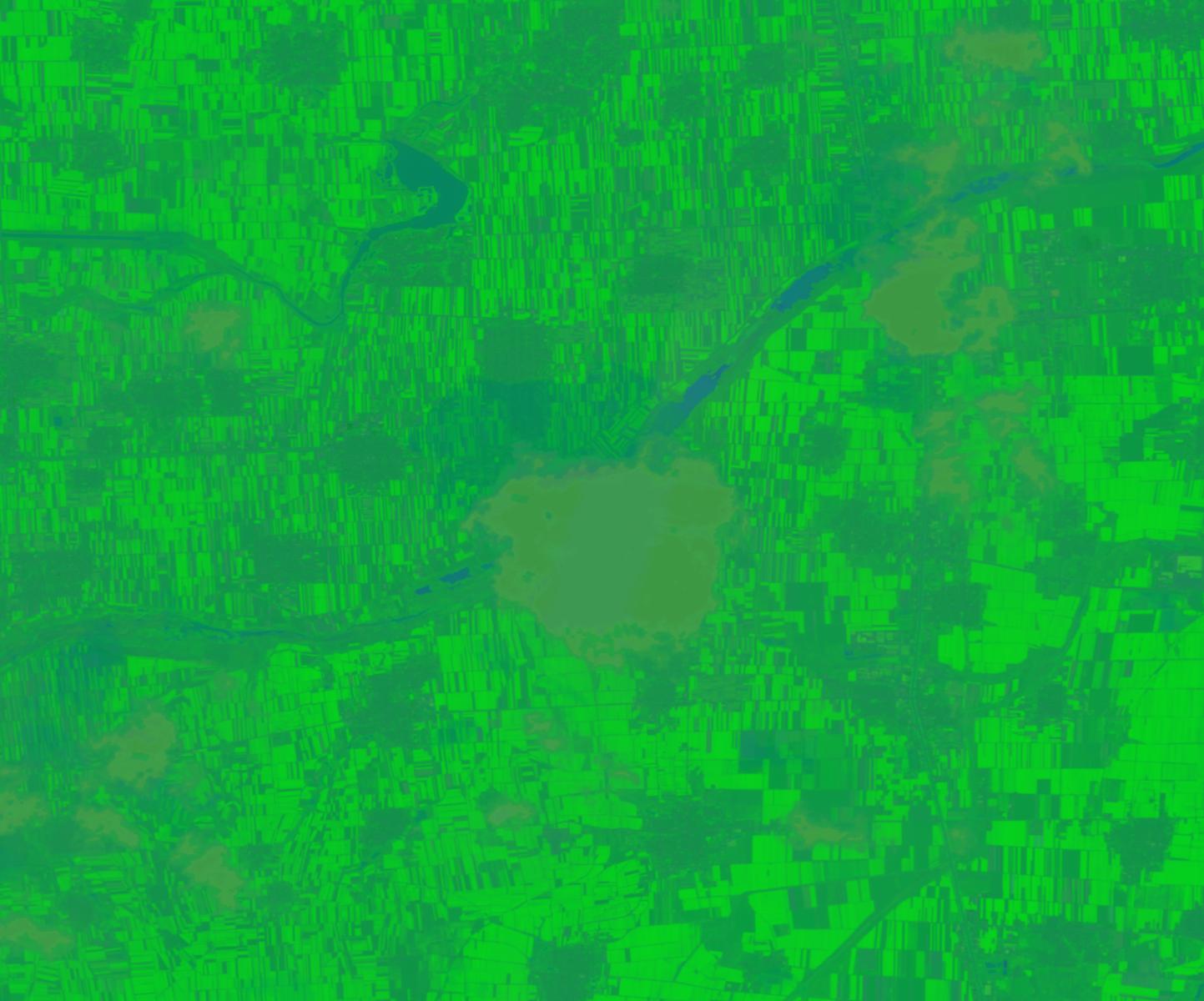}
	\caption{BVW HM-RRN-MoG A}
\end{subfigure}
\begin{subfigure}[t]{.15\textwidth}
	\centering
\includegraphics[width=\textwidth]{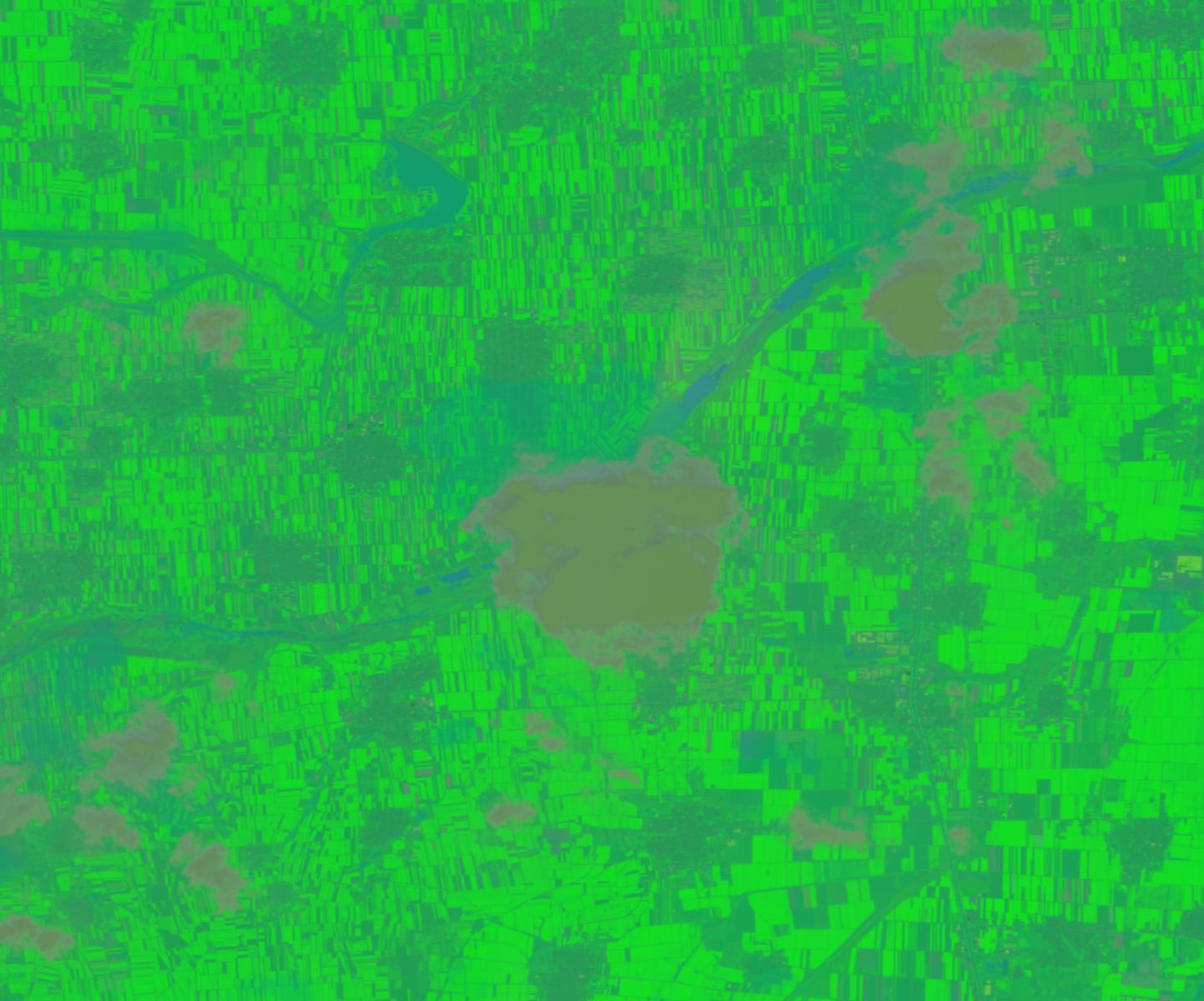}
	\caption{BVW source image A}
\end{subfigure}
\begin{subfigure}[t]{.47\textwidth}
	\centering
\includegraphics[width=\textwidth,clip,trim=0 32 0 0]{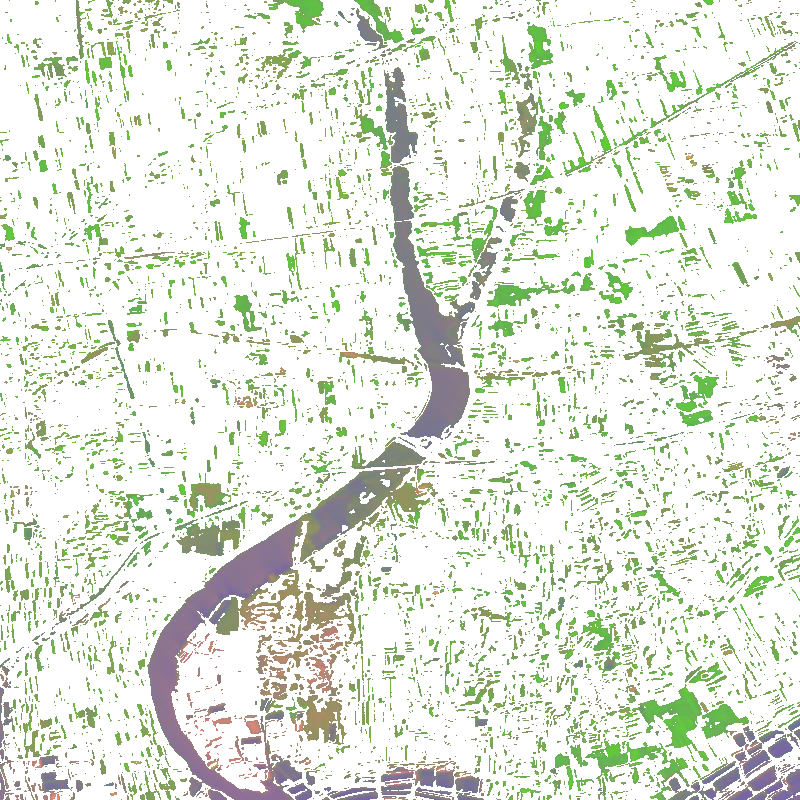}
	\caption{BVW HM-RRN-MoG F change set}
\end{subfigure}
\begin{subfigure}[t]{.47\textwidth}
	\centering
\includegraphics[width=\textwidth]{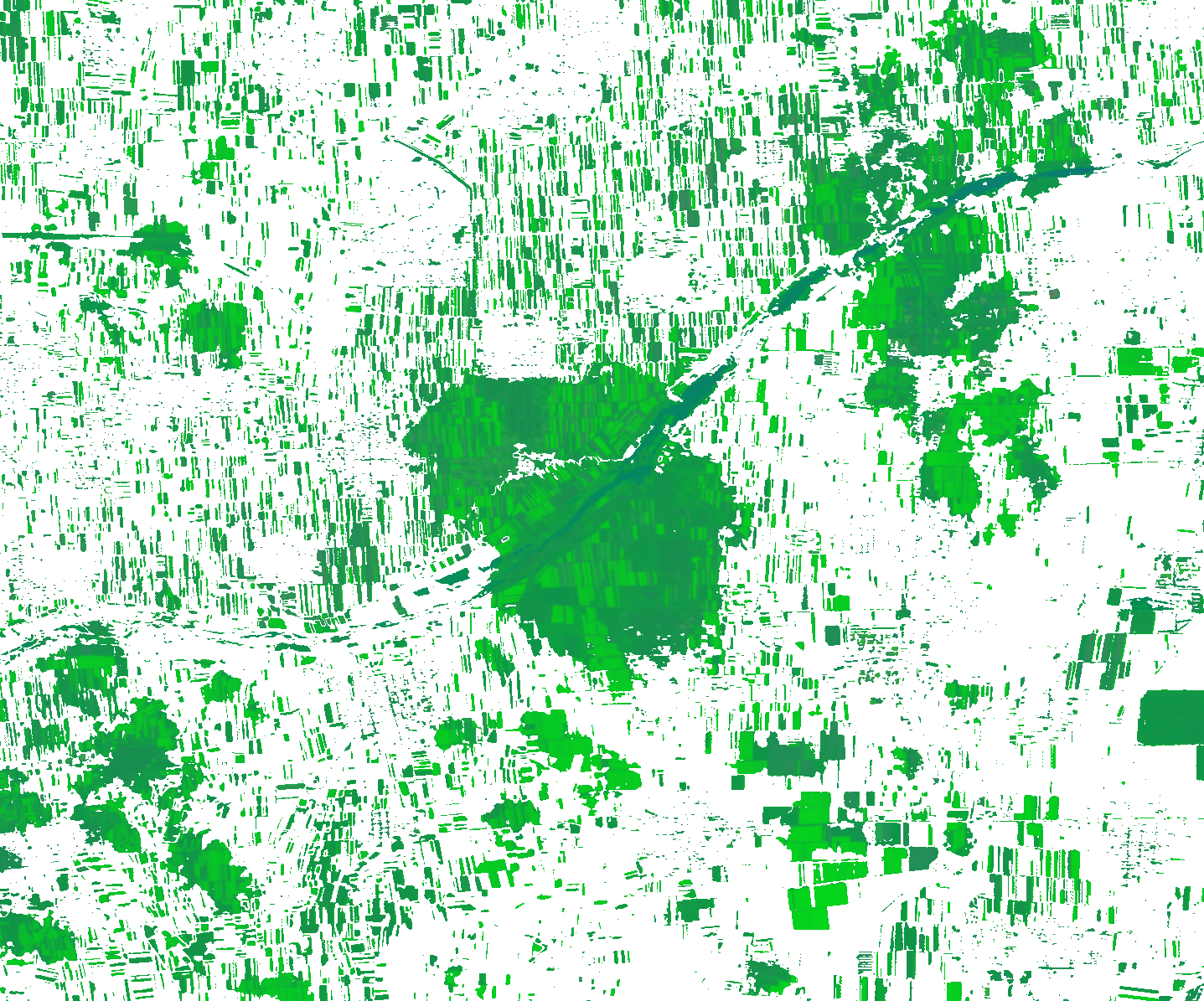}
	\caption{BVW target image B change set}
\end{subfigure}
\caption{Illustration of HM-RRN-MoG reducing reflectance/digital number/index inconsistences and visualization of NDVI and NDWI index on the change set. BVW is the blue-NDVI-NDWI rendered image using the blue reflectance/digital number as the red band, the NDVI index linearly mapped to 0-1 as the green band, the NDWI index linearly mapped to 0-1. The BVW change set is the remained BVW pixels of the change set. The bright green pixels are with high NDVI and bright blue pixels are with high NDWI.}
\label{fig:Illustration of HM-RRN-MoG reducing reflectance/digital number/index inconsistences and visualization of NDVI and NDWI index on the change set.}
\end{figure}

The comparison results are shown in table~\ref{table:HM-RRN-MoG reducing reflectance/(radiance), NDVI, NDWI inconsistence on no-change set.}. We select a small part of images for visualization. The visualized comparison, such as the effect of HM-RRN-MoG on blue/NDVI/NDWI, and the blue/NDVI/NDWI rendered change set are shown in the figure~\ref{fig:Illustration of HM-RRN-MoG reducing reflectance/digital number/index inconsistences and visualization of NDVI and NDWI index on the change set.}.

According to the table~\ref{table:HM-RRN-MoG reducing reflectance/(radiance), NDVI, NDWI inconsistence on no-change set.}, we could find that after implementing the HM-RRN-MoG  method, the RMSEs of reflectance/digital number, NDVI and NDWI are significantly reduced on the no-change set derived by the HM-RRN-MoG method. These characteristics are shown both in the non radiometric calibrated and non atmospheric corrected images and the radiometric calibrated and coarse atmospheric images. It shows the expansibility of the HM-RRN-MoG method and its potent for land cover change detection task regarding its ability to reducing the radiometric inconsistencies.

\begin{table}[]
\caption{Comparison of RMSE regarding image pairs' reflectance/(radiance), NDVI, NDWI inconsistence on no-change set.}
\label{table:HM-RRN-MoG reducing reflectance/(radiance), NDVI, NDWI inconsistence on no-change set.}
\small
\begin{tabular}{lllllll}
\hline
Image pair         & B & G& R& N & NDVI & NDWI \\ \hline
F \   E          & 0.021& 0.018  & 0.024 & 0.070& 0.102 & 0.094   \\ \hline
HM-RRN-MoG F  \ E         & \textbf{0.011}& \textbf{0.012}  & \textbf{0.017} & \textbf{0.035}& \textbf{0.051} & \textbf{0.043 }   \\ \hline
A    B        & 353.88& 168.25  & 707.61 & 903.87& 0.129 & 0.099  \\ \hline
HM-RRN-MoG A \  B        & \textbf{82.06}& \textbf{138.05}  & \textbf{216.53} & \textbf{400.24}& \textbf{0.080} & \textbf{0.061} \\ \hline
\end{tabular}
\small{where B is blue reflectance for image E and image F is digital number for image A and image B, G is green reflectance for image E and image F is digital number for image A and image B, R is red reflectance for image E and image F is digital number for image A and image B, N is near infrared reflectance for image E and image F is digital number for image A and image B, $RMSE=\sqrt{\frac{\sum_{p\in U_{no-change\ set}}{(\alpha_{p1}-\alpha_{p2})^2}}{\vert U_{no-change\ set} \vert}}$}, $U_{no-change\ set}$ is the no-change set, $\alpha_{p1}$ is the feature value located in $p$ position in the first image $\alpha_{p2}$ is the feature value located in $p$ position in the second image.
\end{table}
According to the figure~\ref{fig:Illustration of HM-RRN-MoG reducing reflectance/digital number/index inconsistences and visualization of NDVI and NDWI index on the change set.}, from the BVW rendered images, we can visually tell that the HM-RRN-MoG method reduces the NDVI and NDWI inconsistencies induced by the radiometric inconsistencies. From the blue-NDVI-NDWI (BVW) rendered change set, we can tell the water(NDWI) changes through the blue-green and blue-purple pixels, the vegetation(NDVI)  changes through the bright green pixels, and the man-made urban through orange, khaki and dark green pixels.
\FloatBarrier
\subsection{Change detection with no-change set}

\begin{figure}[ht]
\centering
\captionsetup[subfigure]{font=scriptsize,labelfont=scriptsize}
\begin{subfigure}[t]{.25\textwidth}
	\centering
	\includegraphics[width=\textwidth]{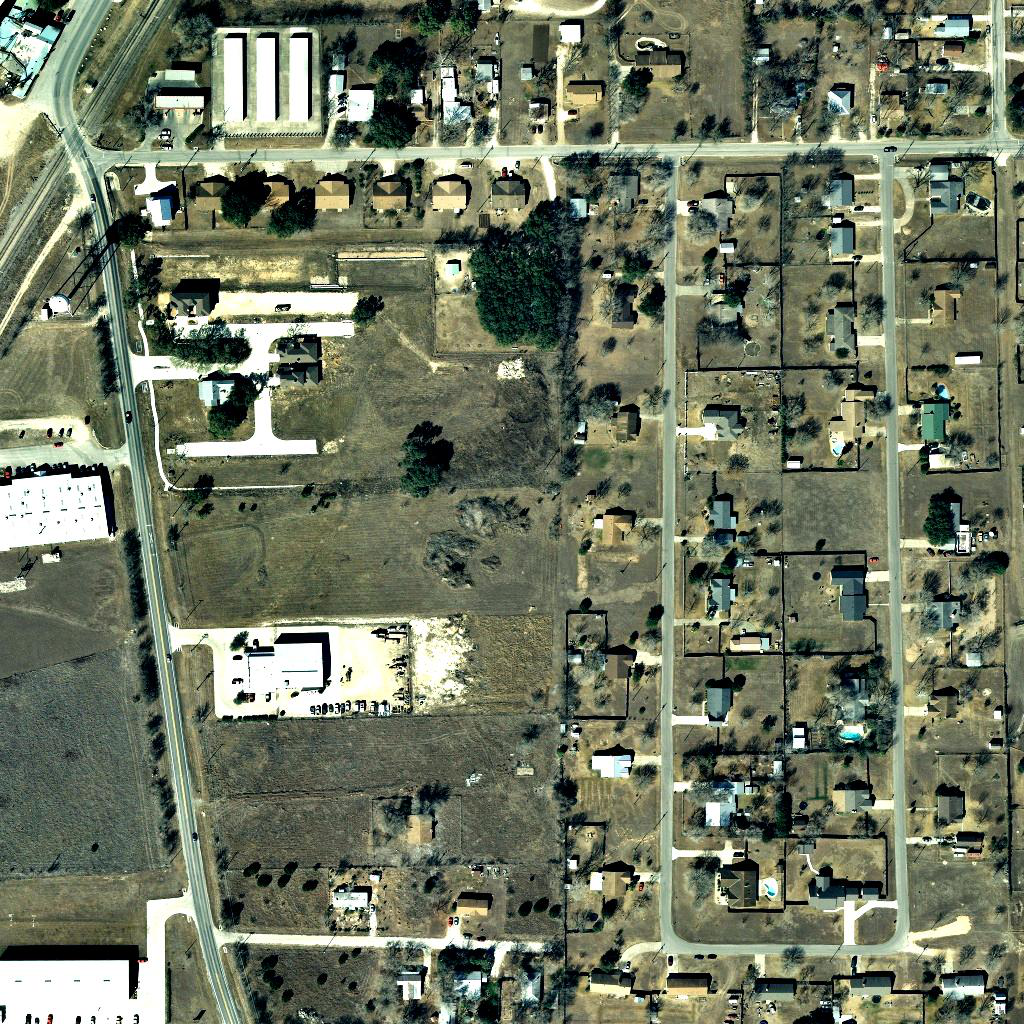}
	\caption{Source image G}
\end{subfigure}
\begin{subfigure}[t]{.25\textwidth}
	\centering
	\includegraphics[width=\textwidth]{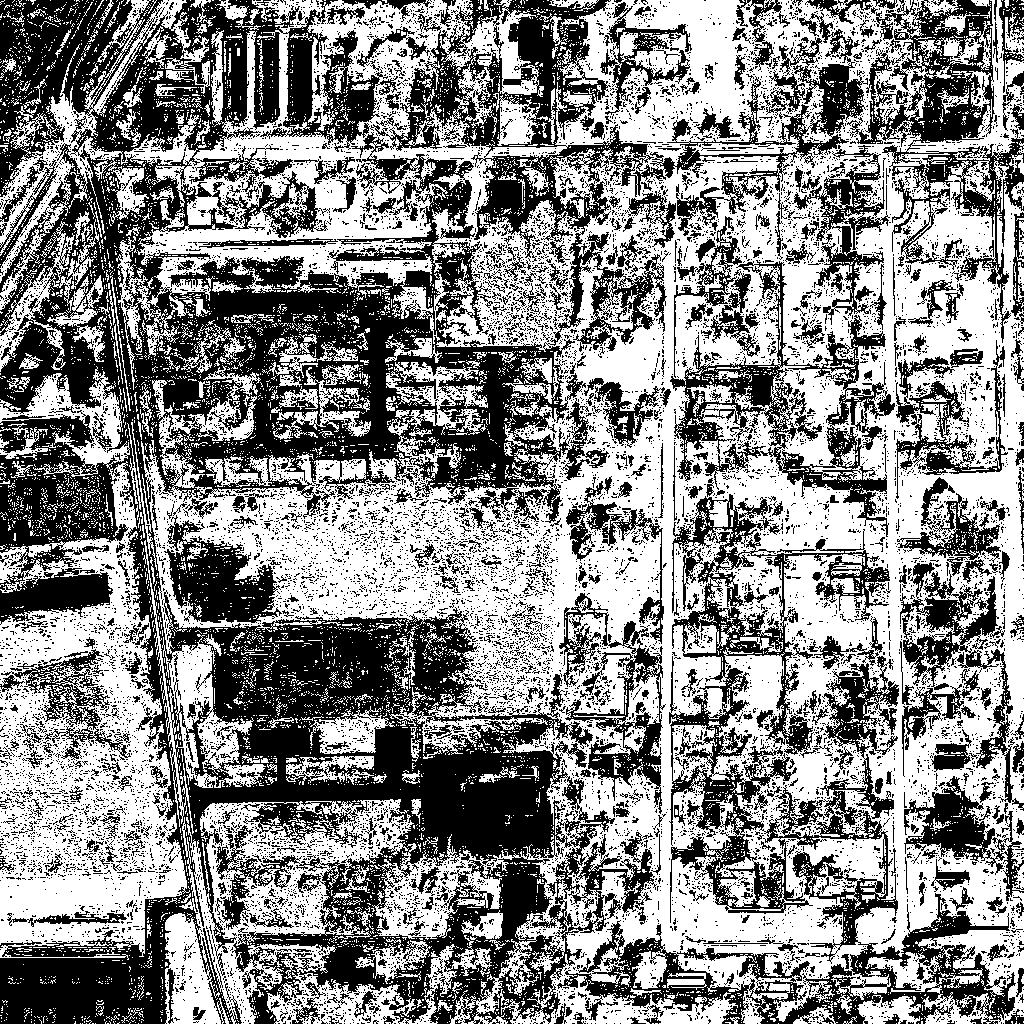}
	\caption{HM-RRN-MoG no-change set}
\end{subfigure}
\begin{subfigure}[t]{.25\textwidth}
	\centering
\includegraphics[width=\textwidth]{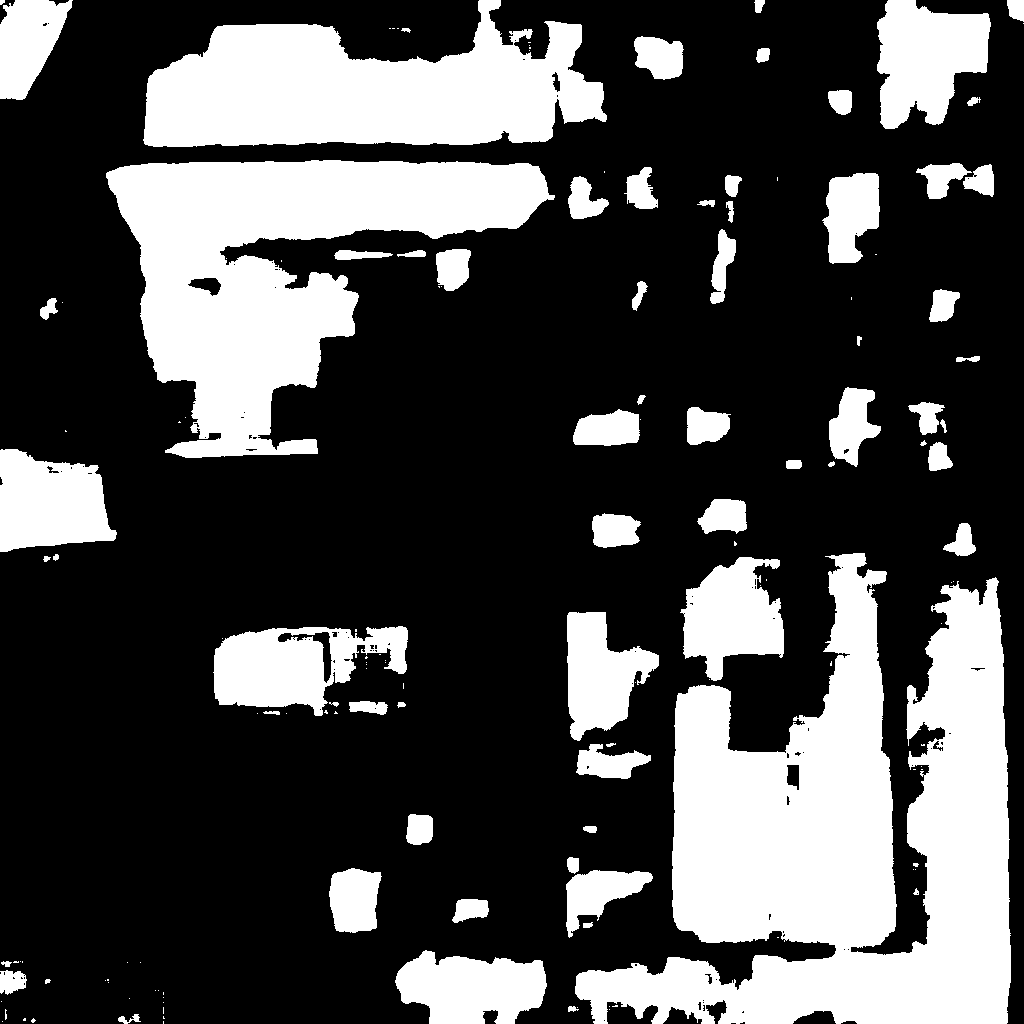}
	\caption{Source image G building area prediction}
\end{subfigure}
\begin{subfigure}[t]{.25\textwidth}
	\centering
\includegraphics[width=\textwidth]{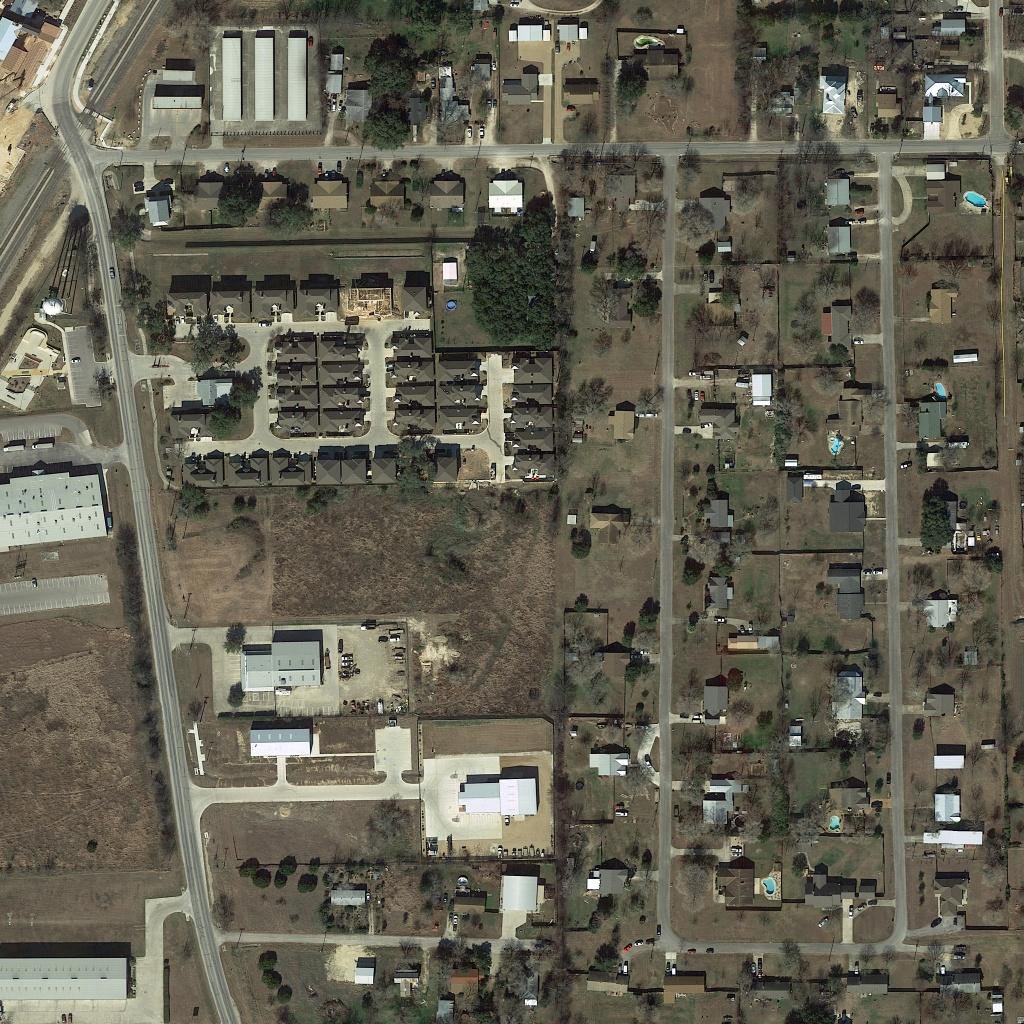}
	\caption{Target image H}
\end{subfigure}
\begin{subfigure}[t]{.25\textwidth}
	\centering
\includegraphics[width=\textwidth]{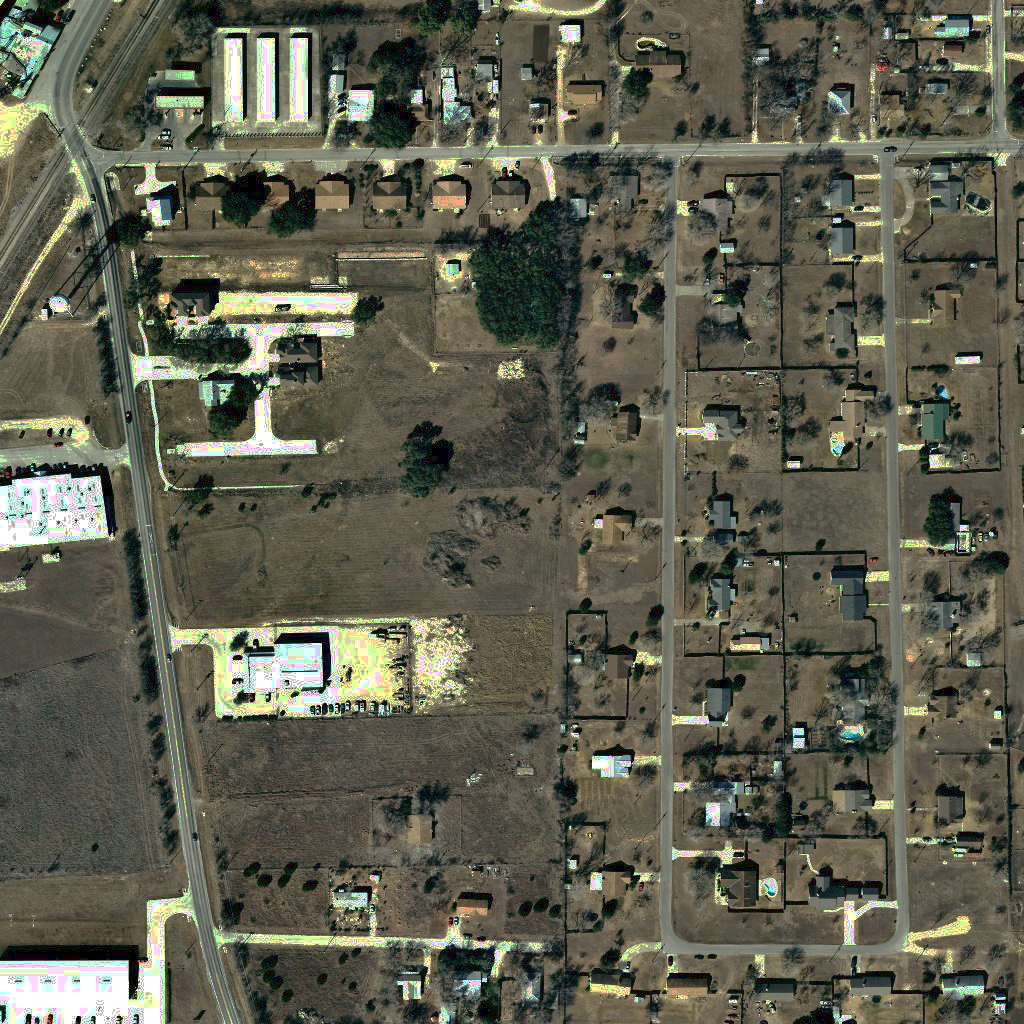}
	\caption{HM-RRN-MoG H}
\end{subfigure}
\begin{subfigure}[t]{.25\textwidth}
	\centering
\includegraphics[width=\textwidth]{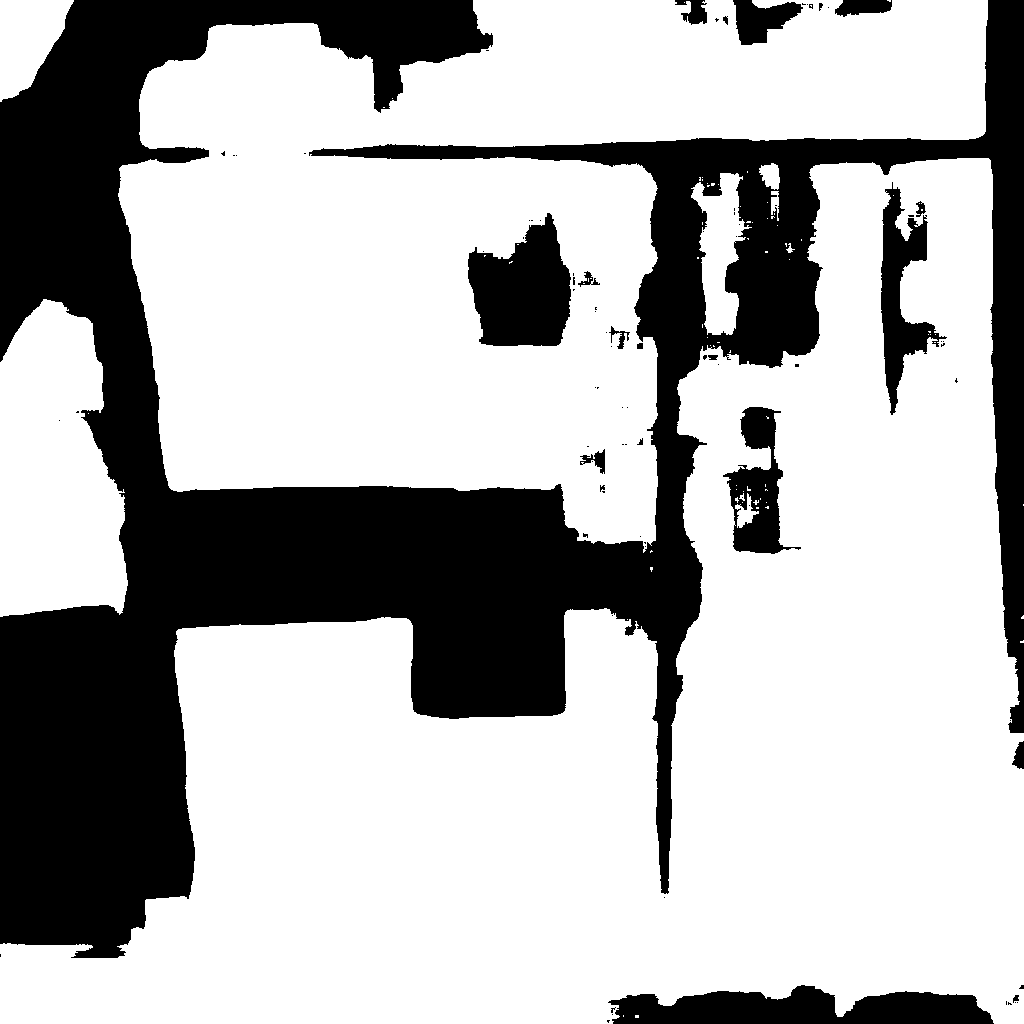}
	\caption{Target image H building area prediction}
\end{subfigure}
\begin{subfigure}[t]{.25\textwidth}
	\centering
\includegraphics[width=\textwidth]{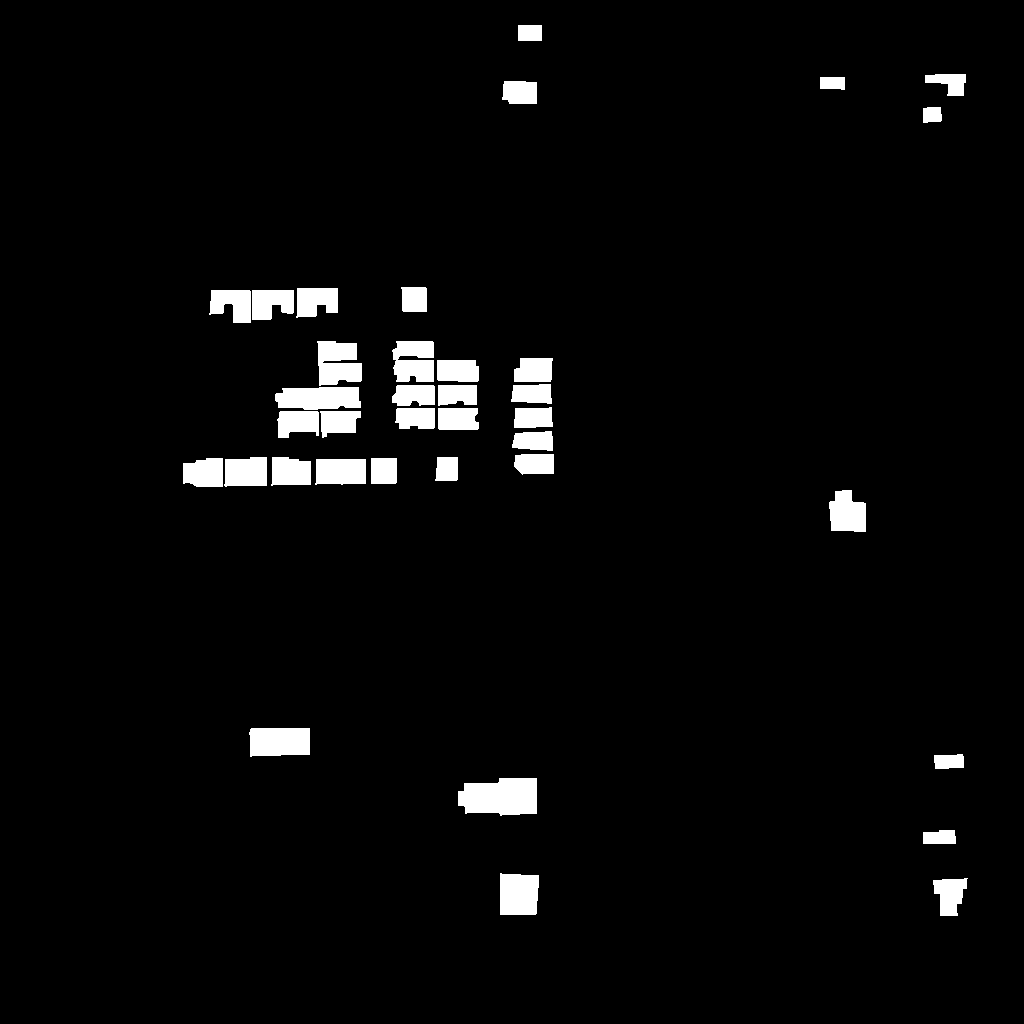}
	\caption{Change label}
\end{subfigure}
\begin{subfigure}[t]{.25\textwidth}
	\centering
\includegraphics[width=\textwidth]{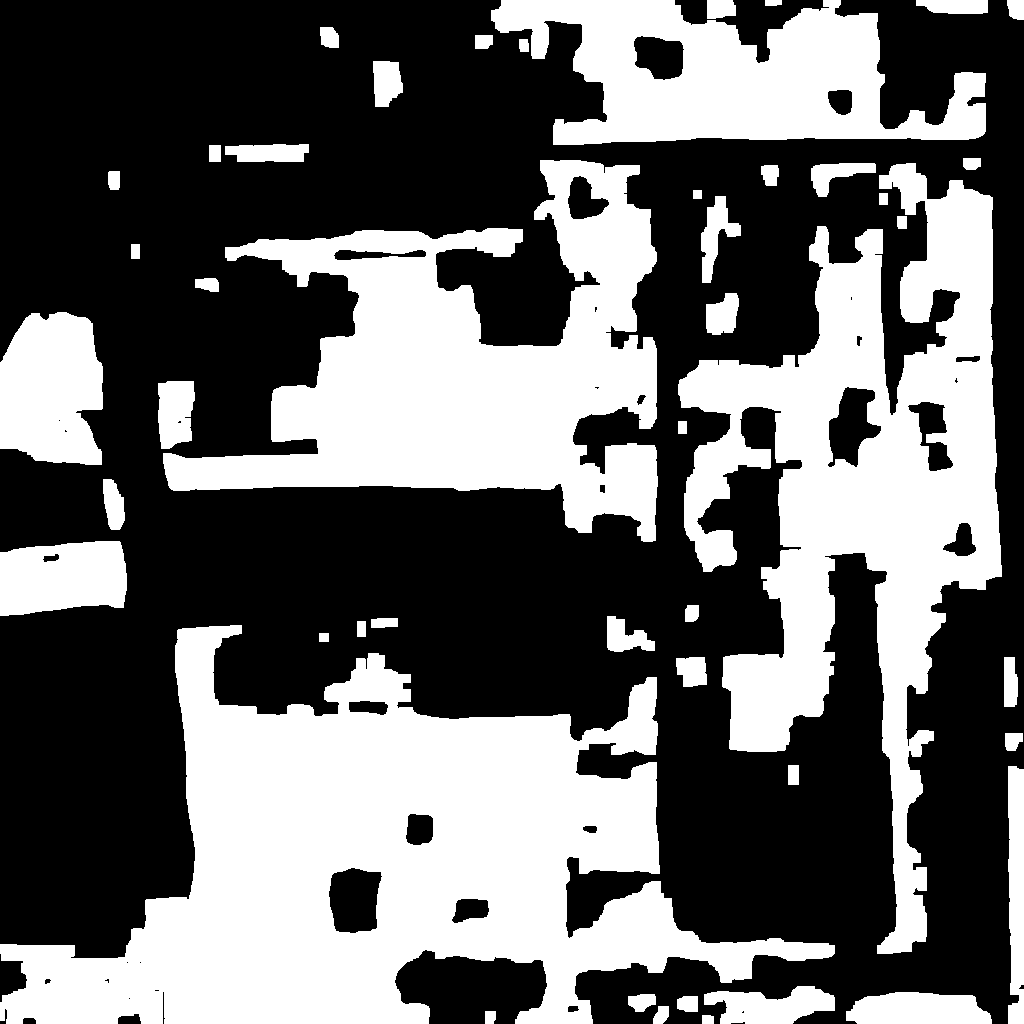}
	\caption{Change prediction}
\end{subfigure}
\begin{subfigure}[t]{.25\textwidth}
	\centering
\includegraphics[width=\textwidth]{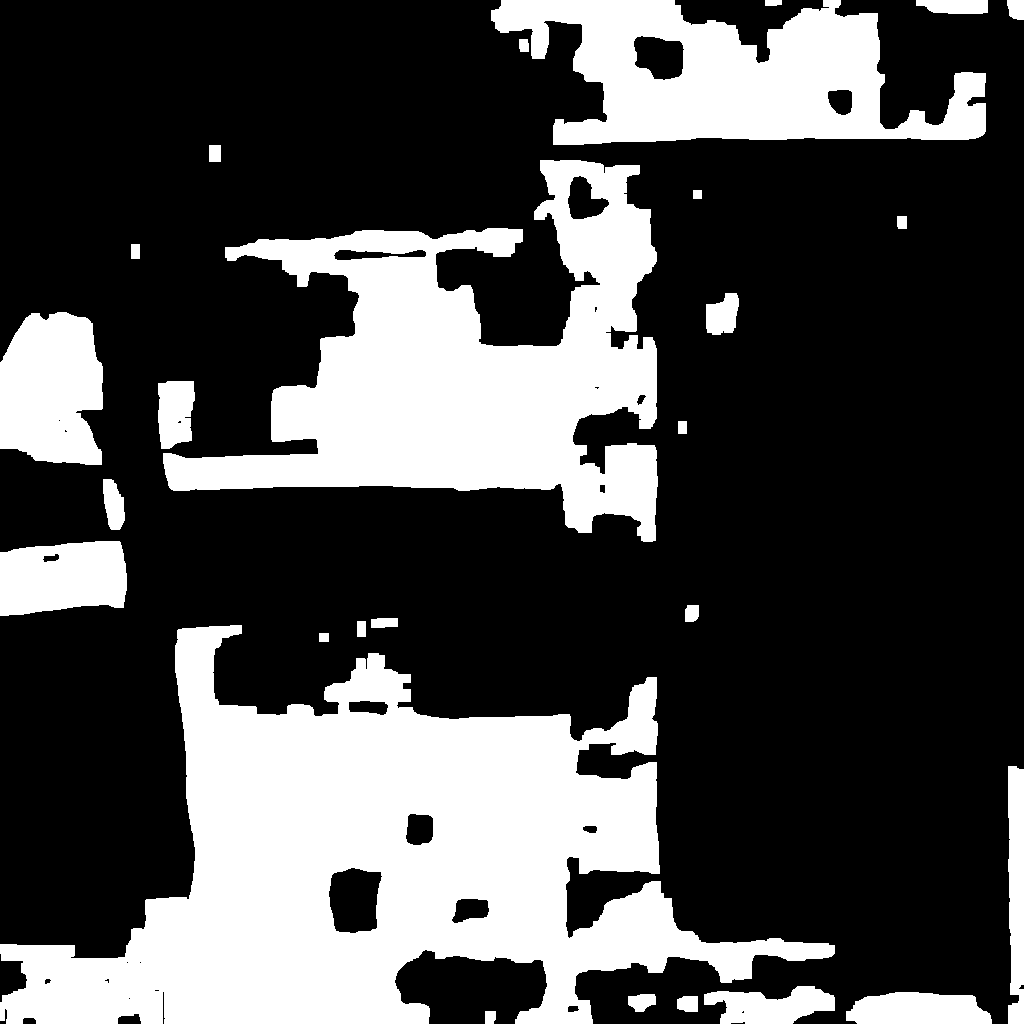}
	\caption{Change prediction with no-change set}
\end{subfigure}
\caption{Illustration of no-change set derived by the HM-RRN-MoG method supporting building change detection task. The white pixels in subfigures c and f are the predicted building area. The white pixels in subfigure g represent the ground truth of building change. The white pixels in subfigures h and i represent the predicted building change where i using the voting with no-change set method.}
\label{fig:Illustration of no-change set derived by the HM-RRN-MoG method supporting building change detection task}
\end{figure}

\begin{table}[ht]
\caption{Change detection evaluation with different methods\label{Change detection evaluation with different methods}}
\small
\centering
\begin{tabular}{cccccc}
\hline
Method          & Accuracy & Recall & Precision & F1-score                                                                                                                   \\ \hline
Direct set difference & 0.835     & \textbf{0.843}     & 0.215    &0.342 \\ \hline
Set difference excluding no-change set  & \textbf{0.901}  & 0.413 & \textbf{0.234}  & 0.298 \\ \hline
Voting with no-change set & 0.840 & 0.841 & 0.220  & 0.349 \\ \hline
Set difference with voted no-change set & 0.843 & 0.834 & 0.222  & \textbf{0.351} \\ \hline
\end{tabular}
\end{table}

Direct difference method computes the symmetrically difference of the predictions of target image and source image(i.e. $T_{buliding}^{pred}\triangle S_{buliding}^{pred}$). Difference excluding no-change set method compute the set difference of the direct results and no-change set(i.e. $T_{buliding}^{pred}\triangle S_{buliding}^{pred} - U_{no \ change \ set}$). Difference voting with no-change set method computes the ratio of no change points in each connected component and deletes connected components with a ratio bigger than $80\%$ and is achieved by the region grow algorithm. The set difference with voted no-change set method refines the no-change set by keeping each point in which there are more than $70\%$ no-change points in the (31,31) window centered.

According to table~\ref{Change detection evaluation with different methods}, we can find that the direct difference method possessed the highest recall. Due to the no-change set is of radiometric level rather than object level, some buildings changed to the hardened surface without significant change in the radiometric. When utilizing these kinds of no-change points, the truth changes are negatively influenced. For instance, the set difference excluding no-change set method significantly drops the recall although it has the best accuracy and precision (However, most of no change points indeed indicate the invariance of the object with proper integration.). We find the set difference with voted no-change set best integrate the no-change set, and this method achieves the highest F1 score. The visualization of no-change set supporting the building change detection task is shown in figure~\ref{fig:Illustration of no-change set derived by the HM-RRN-MoG method supporting building change detection task}

\FloatBarrier
\section{Conclusion}
In this paper, we present an auto robust relative radiometric normalization method via latent change noise modeling. Our model is theoretically grounded regarding the probabilistic theory and mathematics deduction. Our model possesses the ability to robustly against clouds/fogs/changes. It is evaluable with help of the log-likelihood and no-change set root-mean-square-error as well as relatively accurate. It can effectively reduce the reflectance/digital number, NDVI and NDWI inconsistencies between the source and target image on the no-change set, and its change set may be good guidance for vegetation and water change detection. As for the no-change set generated by our method, we find it can further boost the building change detection tasks when using the proper fusion approach.

Currently, we find that our method may be influenced by the geometric precision of the source and target image. In the future, on the one hand, we would study how to jointly achieve geometric correction and relative radiometric normalization no-change set derivation. On the other hand, we would try to unsupervisedly learn the representation of the oracle target image and implement absolute radiometric normalization through a deep generative model to eliminate the dependence of geometric correction.

\section*{Acknowledgement(s)}
We would like thank Jian Liao' supports on the geometric correction algorithm in regard to the preprocessing of the experiment images. We would like thank Fan Wang's supports on  engineered algorithms of the RPC orthorectification, radiometric calibration,
and atmospheric correction.

\section*{References}

\bibliography{mybibfile}

\begin{thebibliography}{30}
\providecommand{\natexlab}[1]{#1}
\providecommand{\url}[1]{\texttt{#1}}
\providecommand{\href}[2]{#2}
\providecommand{\path}[1]{#1}
\providecommand{\DOIprefix}{doi:}
\providecommand{\ArXivprefix}{arXiv:}
\providecommand{\URLprefix}{URL: }
\providecommand{\Pubmedprefix}{pmid:}
\providecommand{\doi}[1]{\href{http://dx.doi.org/#1}{\path{#1}}}
\providecommand{\Pubmed}[1]{\href{pmid:#1}{\path{#1}}}
\providecommand{\BIBand}{and}
\providecommand{\bibinfo}[2]{#2}
\ifx\xfnm\undefined \def\xfnm[#1]{\unskip,\space#1}\fi
\makeatletter\def\@biblabel#1{#1.}\makeatother
\bibitem[{Liu et~al.(2021)Liu, Lian, Zhan, Liu, Tian and
  Duan}]{liu2021automatically}
\bibinfo{author}{Liu\xfnm[ S.]}, \bibinfo{author}{Lian\xfnm[ J.]},
  \bibinfo{author}{Zhan\xfnm[ X.]}, \bibinfo{author}{Liu\xfnm[ C.]},
  \bibinfo{author}{Tian\xfnm[ Y.]}, \bibinfo{author}{Duan\xfnm[ H.]}.
\newblock \bibinfo{title}{Automatically eliminating seam lines with poisson
  editing in complex relative radiometric normalization mosaicking scenarios}.
\newblock \emph{\bibinfo{journal}{arXiv preprint arXiv:210607441}}
  \bibinfo{year}{2021};.
\bibitem[{Vermote et~al.(1997)Vermote, Tanr{\'e}, Deuze, Herman and
  Morcette}]{vermote1997second}
\bibinfo{author}{Vermote\xfnm[ E.F.]}, \bibinfo{author}{Tanr{\'e}\xfnm[ D.]},
  \bibinfo{author}{Deuze\xfnm[ J.L.]}, \bibinfo{author}{Herman\xfnm[ M.]},
  \bibinfo{author}{Morcette\xfnm[ J.J.]}.
\newblock \bibinfo{title}{Second simulation of the satellite signal in the
  solar spectrum, 6s: An overview}.
\newblock \emph{\bibinfo{journal}{IEEE transactions on geoscience and remote
  sensing}}
  \bibinfo{year}{1997};\bibinfo{volume}{35}(\bibinfo{number}{3}):\bibinfo{pages}{675--686}.
\bibitem[{El~Hajj et~al.(2008)El~Hajj, B{\'e}gu{\'e}, Lafrance, Hagolle, Dedieu
  and Rumeau}]{el2008relative}
\bibinfo{author}{El~Hajj\xfnm[ M.]}, \bibinfo{author}{B{\'e}gu{\'e}\xfnm[ A.]},
  \bibinfo{author}{Lafrance\xfnm[ B.]}, \bibinfo{author}{Hagolle\xfnm[ O.]},
  \bibinfo{author}{Dedieu\xfnm[ G.]}, \bibinfo{author}{Rumeau\xfnm[ M.]}.
\newblock \bibinfo{title}{Relative radiometric normalization and atmospheric
  correction of a spot 5 time series}.
\newblock \emph{\bibinfo{journal}{Sensors}}
  \bibinfo{year}{2008};\bibinfo{volume}{8}(\bibinfo{number}{4}):\bibinfo{pages}{2774--2791}.
\bibitem[{Du et~al.(2002)Du, Teillet and Cihlar}]{du2002radiometric}
\bibinfo{author}{Du\xfnm[ Y.]}, \bibinfo{author}{Teillet\xfnm[ P.M.]},
  \bibinfo{author}{Cihlar\xfnm[ J.]}.
\newblock \bibinfo{title}{Radiometric normalization of multitemporal
  high-resolution satellite images with quality control for land cover change
  detection}.
\newblock \emph{\bibinfo{journal}{Remote sensing of Environment}}
  \bibinfo{year}{2002};\bibinfo{volume}{82}(\bibinfo{number}{1}):\bibinfo{pages}{123--134}.
\bibitem[{Jensen(1983)}]{jensen1983urban}
\bibinfo{author}{Jensen\xfnm[ J.R.]}.
\newblock \bibinfo{title}{Urban/suburban land use analysis}.
\newblock \emph{\bibinfo{journal}{Manual of Remote Sensing, second edition}}
  \bibinfo{year}{1983};:\bibinfo{pages}{1571--1666}.
\bibitem[{Singh(1989)}]{singh1989review}
\bibinfo{author}{Singh\xfnm[ A.]}.
\newblock \bibinfo{title}{Review article digital change detection techniques
  using remotely-sensed data}.
\newblock \emph{\bibinfo{journal}{International journal of remote sensing}}
  \bibinfo{year}{1989};\bibinfo{volume}{10}(\bibinfo{number}{6}):\bibinfo{pages}{989--1003}.
\bibitem[{Olsson(1993)}]{olsson1993regression}
\bibinfo{author}{Olsson\xfnm[ H.]}.
\newblock \bibinfo{title}{Regression functions for multitemporal relative
  calibration of thematic mapper data over boreal forest}.
\newblock \emph{\bibinfo{journal}{Remote Sensing of Environment}}
  \bibinfo{year}{1993};\bibinfo{volume}{46}(\bibinfo{number}{1}):\bibinfo{pages}{89--102}.
\bibitem[{Zhang et~al.(2008)Zhang, Yang, Lin and Liao}]{zhang2008automatic}
\bibinfo{author}{Zhang\xfnm[ L.]}, \bibinfo{author}{Yang\xfnm[ L.]},
  \bibinfo{author}{Lin\xfnm[ H.]}, \bibinfo{author}{Liao\xfnm[ M.]}.
\newblock \bibinfo{title}{Automatic relative radiometric normalization using
  iteratively weighted least square regression}.
\newblock \emph{\bibinfo{journal}{International Journal of Remote Sensing}}
  \bibinfo{year}{2008};\bibinfo{volume}{29}(\bibinfo{number}{2}):\bibinfo{pages}{459--470}.
\bibitem[{Richards and Richards(1999)}]{richards1999remote}
\bibinfo{author}{Richards\xfnm[ J.A.]}, \bibinfo{author}{Richards\xfnm[ J.]}.
\newblock \bibinfo{title}{Remote sensing digital image analysis};
  vol.~\bibinfo{volume}{3}.
\newblock \bibinfo{publisher}{Springer}; \bibinfo{year}{1999}.
\bibitem[{Schott et~al.(1988)Schott, Salvaggio and
  Volchok}]{schott1988radiometric}
\bibinfo{author}{Schott\xfnm[ J.R.]}, \bibinfo{author}{Salvaggio\xfnm[ C.]},
  \bibinfo{author}{Volchok\xfnm[ W.J.]}.
\newblock \bibinfo{title}{Radiometric scene normalization using pseudoinvariant
  features}.
\newblock \emph{\bibinfo{journal}{Remote sensing of Environment}}
  \bibinfo{year}{1988};\bibinfo{volume}{26}(\bibinfo{number}{1}):\bibinfo{pages}{1--16}.
\bibitem[{Hall et~al.(1991)Hall, Strebel, Nickeson and
  Goetz}]{hall1991radiometric}
\bibinfo{author}{Hall\xfnm[ F.G.]}, \bibinfo{author}{Strebel\xfnm[ D.E.]},
  \bibinfo{author}{Nickeson\xfnm[ J.E.]}, \bibinfo{author}{Goetz\xfnm[ S.J.]}.
\newblock \bibinfo{title}{Radiometric rectification: toward a common
  radiometric response among multidate, multisensor images}.
\newblock \emph{\bibinfo{journal}{Remote sensing of environment}}
  \bibinfo{year}{1991};\bibinfo{volume}{35}(\bibinfo{number}{1}):\bibinfo{pages}{11--27}.
\bibitem[{Elvidge et~al.(1995)Elvidge, Yuan, Weerackoon and
  Lunetta}]{elvidge1995relative}
\bibinfo{author}{Elvidge\xfnm[ C.D.]}, \bibinfo{author}{Yuan\xfnm[ D.]},
  \bibinfo{author}{Weerackoon\xfnm[ R.D.]}, \bibinfo{author}{Lunetta\xfnm[
  R.S.]}.
\newblock \bibinfo{title}{Relative radiometric normalization of landsat
  multispectral scanner (mss) data using a automatic scattergram-controlled
  regression}.
\newblock \emph{\bibinfo{journal}{Photogrammetric Engineering and Remote
  Sensing}}
  \bibinfo{year}{1995};\bibinfo{volume}{61}(\bibinfo{number}{10}):\bibinfo{pages}{1255--1260}.
\bibitem[{Yang et~al.(2000)Yang, Lo et~al.}]{yang2000relative}
\bibinfo{author}{Yang\xfnm[ X.]}, \bibinfo{author}{Lo\xfnm[ C.]}, et~al.
\newblock \bibinfo{title}{Relative radiometric normalization performance for
  change detection from multi-date satellite images}.
\newblock \emph{\bibinfo{journal}{Photogrammetric engineering and remote
  sensing}}
  \bibinfo{year}{2000};\bibinfo{volume}{66}(\bibinfo{number}{8}):\bibinfo{pages}{967--980}.
\bibitem[{Biegel and Schott(1985)}]{biegel1985radiometric}
\bibinfo{author}{Biegel\xfnm[ J.D.]}, \bibinfo{author}{Schott\xfnm[ J.R.]}.
\newblock \bibinfo{title}{Radiometric calibration and image processing of
  landsat tm data to improve assessment of thermal signatures}.
\newblock In: \emph{\bibinfo{booktitle}{Infrared technology X}}; vol.
  \bibinfo{volume}{510}. \bibinfo{organization}{International Society for
  Optics and Photonics}; \bibinfo{year}{1985}:\unskip
  \bibinfo{pages}{193--200}.
\bibitem[{Sun et~al.(2012)Sun, Fang, Liu, Wang and Tong}]{sun2012automatic}
\bibinfo{author}{Sun\xfnm[ T.]}, \bibinfo{author}{Fang\xfnm[ J.Y.]},
  \bibinfo{author}{Liu\xfnm[ X.]}, \bibinfo{author}{Wang\xfnm[ J.N.]},
  \bibinfo{author}{Tong\xfnm[ Q.X.]}.
\newblock \bibinfo{title}{Automatic relative radiometric normalization method
  based on sift feature matching}.
\newblock \emph{\bibinfo{journal}{Journal of Infrared and Millimeter Waves}}
  \bibinfo{year}{2012};\bibinfo{volume}{31}(\bibinfo{number}{4}):\bibinfo{pages}{355--359}.
\bibitem[{Yong et~al.(2017)Yong, Meng, Zuo and Zhang}]{yong2017robust}
\bibinfo{author}{Yong\xfnm[ H.]}, \bibinfo{author}{Meng\xfnm[ D.]},
  \bibinfo{author}{Zuo\xfnm[ W.]}, \bibinfo{author}{Zhang\xfnm[ L.]}.
\newblock \bibinfo{title}{Robust online matrix factorization for dynamic
  background subtraction}.
\newblock \emph{\bibinfo{journal}{IEEE transactions on pattern analysis and
  machine intelligence}}
  \bibinfo{year}{2017};\bibinfo{volume}{40}(\bibinfo{number}{7}):\bibinfo{pages}{1726--1740}.
\bibitem[{Meng and De~La~Torre(2013)}]{meng2013robust}
\bibinfo{author}{Meng\xfnm[ D.]}, \bibinfo{author}{De~La~Torre\xfnm[ F.]}.
\newblock \bibinfo{title}{Robust matrix factorization with unknown noise}.
\newblock In: \emph{\bibinfo{booktitle}{Proceedings of the IEEE International
  Conference on Computer Vision}}. \bibinfo{year}{2013}:\unskip
  \bibinfo{pages}{1337--1344}.
\bibitem[{Chen et~al.(2017)Chen, Cao, Zhao, Meng and Xu}]{chen2017denoising}
\bibinfo{author}{Chen\xfnm[ Y.]}, \bibinfo{author}{Cao\xfnm[ X.]},
  \bibinfo{author}{Zhao\xfnm[ Q.]}, \bibinfo{author}{Meng\xfnm[ D.]},
  \bibinfo{author}{Xu\xfnm[ Z.]}.
\newblock \bibinfo{title}{Denoising hyperspectral image with non-iid noise
  structure}.
\newblock \emph{\bibinfo{journal}{IEEE transactions on cybernetics}}
  \bibinfo{year}{2017};\bibinfo{volume}{48}(\bibinfo{number}{3}):\bibinfo{pages}{1054--1066}.
\bibitem[{Cao et~al.(2016)Cao, Zhao, Meng, Chen and Xu}]{cao2016robust}
\bibinfo{author}{Cao\xfnm[ X.]}, \bibinfo{author}{Zhao\xfnm[ Q.]},
  \bibinfo{author}{Meng\xfnm[ D.]}, \bibinfo{author}{Chen\xfnm[ Y.]},
  \bibinfo{author}{Xu\xfnm[ Z.]}.
\newblock \bibinfo{title}{Robust low-rank matrix factorization under general
  mixture noise distributions}.
\newblock \emph{\bibinfo{journal}{IEEE Transactions on Image Processing}}
  \bibinfo{year}{2016};\bibinfo{volume}{25}(\bibinfo{number}{10}):\bibinfo{pages}{4677--4690}.
\bibitem[{Cao et~al.(2015)Cao, Chen, Zhao, Meng, Wang, Wang and
  Xu}]{cao2015low}
\bibinfo{author}{Cao\xfnm[ X.]}, \bibinfo{author}{Chen\xfnm[ Y.]},
  \bibinfo{author}{Zhao\xfnm[ Q.]}, \bibinfo{author}{Meng\xfnm[ D.]},
  \bibinfo{author}{Wang\xfnm[ Y.]}, \bibinfo{author}{Wang\xfnm[ D.]},
  \bibinfo{author}{Xu\xfnm[ Z.]}.
\newblock \bibinfo{title}{Low-rank matrix factorization under general mixture
  noise distributions}.
\newblock In: \emph{\bibinfo{booktitle}{Proceedings of the IEEE international
  conference on computer vision}}. \bibinfo{year}{2015}:\unskip
  \bibinfo{pages}{1493--1501}.
\bibitem[{Zhao et~al.(2014)Zhao, Meng, Xu, Zuo and Zhang}]{zhao2014robust}
\bibinfo{author}{Zhao\xfnm[ Q.]}, \bibinfo{author}{Meng\xfnm[ D.]},
  \bibinfo{author}{Xu\xfnm[ Z.]}, \bibinfo{author}{Zuo\xfnm[ W.]},
  \bibinfo{author}{Zhang\xfnm[ L.]}.
\newblock \bibinfo{title}{Robust principal component analysis with complex
  noise}.
\newblock In: \emph{\bibinfo{booktitle}{International conference on machine
  learning}}. \bibinfo{organization}{PMLR}; \bibinfo{year}{2014}:\unskip
  \bibinfo{pages}{55--63}.
\bibitem[{Yue et~al.(2021)Yue, Zhao, Xie, Zhang and Meng}]{yue2021unsupervised}
\bibinfo{author}{Yue\xfnm[ Z.]}, \bibinfo{author}{Zhao\xfnm[ Q.]},
  \bibinfo{author}{Xie\xfnm[ J.]}, \bibinfo{author}{Zhang\xfnm[ L.]},
  \bibinfo{author}{Meng\xfnm[ D.]}.
\newblock \bibinfo{title}{Unsupervised single image super-resolution under
  complex noise}.
\newblock \emph{\bibinfo{journal}{arXiv preprint arXiv:210700986}}
  \bibinfo{year}{2021};.
\bibitem[{Rui et~al.(2021)Rui, Cao, Xie, Yue, Zhao and Meng}]{rui2021learning}
\bibinfo{author}{Rui\xfnm[ X.]}, \bibinfo{author}{Cao\xfnm[ X.]},
  \bibinfo{author}{Xie\xfnm[ Q.]}, \bibinfo{author}{Yue\xfnm[ Z.]},
  \bibinfo{author}{Zhao\xfnm[ Q.]}, \bibinfo{author}{Meng\xfnm[ D.]}.
\newblock \bibinfo{title}{Learning an explicit weighting scheme for adapting
  complex hsi noise}.
\newblock In: \emph{\bibinfo{booktitle}{Proceedings of the IEEE/CVF Conference
  on Computer Vision and Pattern Recognition}}. \bibinfo{year}{2021}:\unskip
  \bibinfo{pages}{6739--6748}.
\bibitem[{McFeeters(1996)}]{mcfeeters1996use}
\bibinfo{author}{McFeeters\xfnm[ S.K.]}.
\newblock \bibinfo{title}{The use of the normalized difference water index
  (ndwi) in the delineation of open water features}.
\newblock \emph{\bibinfo{journal}{International journal of remote sensing}}
  \bibinfo{year}{1996};\bibinfo{volume}{17}(\bibinfo{number}{7}):\bibinfo{pages}{1425--1432}.
\bibitem[{Koller and Friedman(2009)}]{koller2009probabilistic}
\bibinfo{author}{Koller\xfnm[ D.]}, \bibinfo{author}{Friedman\xfnm[ N.]}.
\newblock \bibinfo{title}{Probabilistic graphical models: principles and
  techniques}.
\newblock \bibinfo{publisher}{MIT press}; \bibinfo{year}{2009}.
\bibitem[{Dempster et~al.(1977)Dempster, Laird and Rubin}]{dempster1977maximum}
\bibinfo{author}{Dempster\xfnm[ A.P.]}, \bibinfo{author}{Laird\xfnm[ N.M.]},
  \bibinfo{author}{Rubin\xfnm[ D.B.]}.
\newblock \bibinfo{title}{Maximum likelihood from incomplete data via the em
  algorithm}.
\newblock \emph{\bibinfo{journal}{Journal of the Royal Statistical Society:
  Series B (Methodological)}}
  \bibinfo{year}{1977};\bibinfo{volume}{39}(\bibinfo{number}{1}):\bibinfo{pages}{1--22}.
\bibitem[{Wu(1983)}]{wu1983convergence}
\bibinfo{author}{Wu\xfnm[ C.J.]}.
\newblock \bibinfo{title}{On the convergence properties of the em algorithm}.
\newblock \emph{\bibinfo{journal}{The Annals of statistics}}
  \bibinfo{year}{1983};:\bibinfo{pages}{95--103}.
\bibitem[{Werman et~al.(1985)Werman, Peleg and Rosenfeld}]{werman1985distance}
\bibinfo{author}{Werman\xfnm[ M.]}, \bibinfo{author}{Peleg\xfnm[ S.]},
  \bibinfo{author}{Rosenfeld\xfnm[ A.]}.
\newblock \bibinfo{title}{A distance metric for multidimensional histograms}.
\newblock \emph{\bibinfo{journal}{Computer Vision, Graphics, and Image
  Processing}}
  \bibinfo{year}{1985};\bibinfo{volume}{32}(\bibinfo{number}{3}):\bibinfo{pages}{328--336}.
\bibitem[{Chen and Shi(2020)}]{chen2020spatial}
\bibinfo{author}{Chen\xfnm[ H.]}, \bibinfo{author}{Shi\xfnm[ Z.]}.
\newblock \bibinfo{title}{A spatial-temporal attention-based method and a new
  dataset for remote sensing image change detection}.
\newblock \emph{\bibinfo{journal}{Remote Sensing}}
  \bibinfo{year}{2020};\bibinfo{volume}{12}(\bibinfo{number}{10}):\bibinfo{pages}{1662}.
\bibitem[{Zhou et~al.(2018)Zhou, Zhang and Wu}]{zhou2018d}
\bibinfo{author}{Zhou\xfnm[ L.]}, \bibinfo{author}{Zhang\xfnm[ C.]},
  \bibinfo{author}{Wu\xfnm[ M.]}.
\newblock \bibinfo{title}{D-linknet: Linknet with pretrained encoder and
  dilated convolution for high resolution satellite imagery road extraction}.
\newblock In: \emph{\bibinfo{booktitle}{Proceedings of the IEEE Conference on
  Computer Vision and Pattern Recognition Workshops}}.
  \bibinfo{year}{2018}:\unskip \bibinfo{pages}{182--186}.

\end{thebibliography}

\end{document}